\documentclass[hidelinks]{article}

\usepackage[final]{neurips_2021}
\usepackage{natbib}
\usepackage[utf8]{inputenc} %
\usepackage[T1]{fontenc}    %
\usepackage{hyperref}       %
\usepackage{url}            %
\usepackage{booktabs}       %
\usepackage{amsfonts}       %
\usepackage{nicefrac}       %
\usepackage{microtype}      %
\usepackage{xcolor}         %

\usepackage{amsfonts,amsmath,amssymb,amsthm,graphicx,float, %
  enumerate,color, url, hyperref, multicol,bm,algorithm,algorithmic}
\usepackage[normalem]{ulem}

\newcommand{\htheta}{{\hat{\boldsymbol{\theta}}}}
\newcommand{\Nor}{\mathbb{N}}

\newcommand{\0}{\mathbf{0}}%
\newcommand{\btheta}{\boldsymbol{\theta}}%
\newcommand{\bbeta}{\boldsymbol{\beta}}%

\newcommand{\onen}{\frac{1}{n}}%
\newcommand{\oneN}{\frac{1}{N}}%
\newcommand{\M}{\mathbf{M}}%
\newcommand{\sumN}{\sum_{i=1}^{N}}%
\newcommand{\tp}{^{\mathrm{T}}}%
\newcommand{\x}{\mathbf{x}}%
\newcommand{\y}{\mathbf{y}}%
\newcommand{\ud}{\mathrm{d}}%
\newcommand{\tr}{\mathrm{tr}}%
\newcommand{\Exp}{\mathbb{E}}%
\newcommand{\Var}{\mathbb{V}}%
\newcommand{\os}{{\mathrm{os}}}%
\newcommand{\Dd}{\mathcal{D}_{\delta}}%
\newcommand{\X}{\mathbf{X}}%
\newcommand{\z}{\mathbf{z}}%
\newcommand{\I}{\mathbf{I}}%

\newcommand{\U}{\mathbf{U}}
\newcommand{\V}{\mathbf{V}}
\renewcommand{\u}{\mathbf{u}}
\renewcommand{\v}{\mathbf{v}}
\newcommand{\bLambda}{\boldsymbol{\Lambda}}
\newcommand{\op}{o_P(1)}

\newtheorem{theorem}{Theorem} 
\theoremstyle{definition}
\newtheorem{assumption}{Assumption}
\newtheorem{remark}{Remark}

\newcommand{\bvtheta}{{\bm{\vartheta}}}
\newcommand{\plt}{\mathrm{plt}}

\newcommand{\ttheta}{\tilde{\bm{\theta}}}

\newcommand{\lik}{\mathrm{lik}}%

\newcommand{\f}{{\mathrm{f}}}

\allowdisplaybreaks

\title{Nonuniform Negative Sampling and Log Odds Correction with Rare Events Data}

\author{%
  HaiYing Wang\\ %
  Department of Statistics\\
  University of Connecticut\\
  \texttt{haiying.wang@uconn.edu} \\
  \And
  Aonan Zhang \\
  ByteDance Inc.\\
  \texttt{aonan.zhang@bytedance.com} \\
  \AND
  Chong Wang \\
  ByteDance Inc.\\
  \texttt{chong.wang@bytedance.com} \\
}

\begin{document}

\maketitle

\begin{abstract}
We investigate the issue of parameter estimation with nonuniform negative sampling for imbalanced data. We first prove that, with imbalanced data, the available information about unknown parameters is only tied to the relatively small number of positive instances, which justifies the usage of negative sampling. However, if the negative instances are subsampled to the same level of the positive cases, there is information loss. To maintain more information, we derive the asymptotic distribution of a general inverse probability weighted (IPW) estimator and obtain the optimal sampling probability that minimizes its variance. To further improve the estimation efficiency over the IPW method, we propose a likelihood-based estimator by correcting log odds for the sampled data and prove that the improved estimator has the smallest asymptotic variance among a large class of estimators. It is also more robust to pilot misspecification. We validate our approach on simulated data as well as a real click-through rate dataset with more than 0.3 trillion instances, collected over a period of a month. Both theoretical and empirical results demonstrate the effectiveness of our method.
\end{abstract}

\section{Introduction}
Imbalanced data are ubiquitous in scientific fields and applications with binary response outputs, where the number of instances in the positive class is much smaller than that in the negative class. For example, in online search or recommendation systems where billions of impressions appear each day, non-click impressions usually dominate.
Using these non-click impressions as negative instances is prevalent and proved to be useful especially in modern machine learning systems~\citep{mcmahan2013ad,he2014practical,chen2016deep,guo2017deepfm,huang2020embedding}.
A common approach to address imbalanced data is to balance it by subsampling the negative class \citep{drummond2003c4,Zhou2009Exploratory} and/or up-sampling the positive class \citep{chawla2002smote,Han2005Borderline,mathew2017classification,douzas2017self}, and a great deal of attentions have been drawn to this problem, see \cite{Japkowicz2000,king2001logistic,chawla2004editorial,estabrooks2004multiple,owen2007infinitely,sun2007cost,chawla2009data,rahman2013addressing,fithian2014local,lemaitre2017imbalanced,Wang2020RareICML}, and the references therein.

In this paper, we focus on subsampling the negative class, i.e., negative sampling, while keeping all rare positive instances.
  This is beneficial for modern online learning systems, where negative sampling significantly reduces data volume in distributed storage while saving training time for multiple downstream models. %
 Rare events data and negative sampling are studied in \cite{Wang2020RareICML}, but it focused on linear logistic regression and only considered uniform sampling. We approach this problem under general binary response models from the perspective of optimal subsampling, which aims to minimize the asymptotic variance of the IPW estimator \citep{WangZhuMa2018,ting2018optimal, wang2021optimal,yu2020quasi}.  This topic was not well investigated neither for imbalanced data nor for negative sampling. In addition, existing optimal probability formulation minimizes the conditional asymptotic variance and the variations due to the data randomness are ignored. We fill these gaps by deriving the asymptotic distributions for both full data and subsample estimators under general binary response models, and we find the unconditional optimal probability for negative sampling. %
  In addition, since IPW may inflate the variance %
  \citep{hesterberg1995weighted, owen2000safe}, we develop a more efficient likelihood-based estimator through nonuniform log odds correction to avoid IPW. 
Our main contributions are summarized as follows.
\begin{itemize} %
\item Under a general binary response model with rare events data, we find that the difference between the full data estimator and the true parameter converges to a normal distribution at a rate that is tied to the number of rare positive instances only.
  This indicates the possibility to throw away the majority of negative instances without information loss, i.e, it justifies the usage of negative sampling. 
\item We show that there is no asymptotic information loss with aggressive negative sampling if the negative instances kept dominates the positive instances. However there is information loss expressed as a variance inflation if the negative instances are brought down to the same level of the positive instances. For this case, we obtain optimal subsampling probabilities by minimizing the unconditional asymptotic variance of a general IPW estimator.
\item We develop a likelihood estimator through nonuniform log odds correction for sampled data, and prove that it has the smallest asymptotic variance %
    among a large class of estimators. %
\item We apply the proposed method to a real online streaming click-through rate (CTR) dataset with more than 0.3 trillion instances and demonstrate its effectiveness. 
\end{itemize}

The rest of paper is organized as follows. Section~\ref{sec:problem} defines the problem this paper focuses on and shows the full data results.  Section~\ref{sec:weight-estim-optim} presents the optimal probability for the IPW estimator. Section~\ref{sec:more-effic-estim} proposes the likelihood-based estimator and %
establishes its asymptotic optimality. Section~\ref{sec:practical} discusses practical implementation %
and the theoretical properties of the resulting estimators. %
Section~\ref{sec:numer-exper} presents numerical experiments with simulated data, and applies the proposed method to a real online streaming CTR dataset with more than 0.3 trillion instances.
Section~\ref{sec:conclusions} concludes the paper and points out limitations of our investigation. Proofs and additional experiments with misspecifications are in the supplementary materials.

\section{Problem setup and full data results}\label{sec:problem}
Let $\{(\x_i,y_i)\}_{i=1}^N$ be training data of size $N$ that satisfies the binary response model,
\begin{equation}\label{eq:1}
  \Pr(y=1\mid\x)=p(\x;\btheta),
\end{equation}
where $y\in\{0,1\}$ is the binary class label ($y=1$ for the positive and $y=0$ for the negative), $\x$ is the vector of features, and $\btheta$ is the unknown $d$-dimensional parameter. %

Let $N_1$ be the number of positive instances ($N_1=\sumN y_i$) and $N_0$ be the number of negative instances ($N_0=N-N_1$). We consider the scenario of massive imbalanced data, i.e., $N_1\ll N_0$. %
For asymptotic investigations when $N_1$ is much smaller than $N_0$, it is more appropriate to assume that $N_1$ increases in a slower rate compared with $N_0$, i.e., $N_1/N_0\rightarrow0$ in probability as $N\rightarrow\infty$ \citep{Wang2020RareICML}. 
This requires $\Pr(y=1)=\Exp\{p(\x;\btheta)\}\rightarrow0$ as $N\rightarrow\infty$ on the model side. %
We assume that the parameter $\btheta$ contains two components $\btheta=(\alpha,\bbeta\tp)\tp$ and the log odds can be written as 
\begin{equation*}
  g(\x;\btheta):=
  \log\Big\{\frac{p(\x;\btheta)}{1-p(\x;\btheta)}\Big\}
  =\alpha+f(\x;\bbeta),
\end{equation*}
where the true parameter, denoted as $\btheta^*=(\alpha^{*},{\bbeta^{*}}\tp)\tp$, satisfies that $\alpha^{*}\rightarrow-\infty$ as $N\rightarrow\infty$ and $\bbeta^{*}$ is fixed. Here $f(\x;\bbeta)$ is a smooth function of $\bbeta$, such as a linear function or a nonlinear neural network model. If it is linear, %
 the model reduces to the logistic regression model. 
A diverging $\alpha^*$ and a fixed $\bbeta^{*}$ indicate that both the marginal and conditional probabilities for a positive instance are small. Heuristically, this implies that a change in the feature value does make a rare event become a large probability event. We can also allow $\bbeta^{*}$ to change with $N$, but as long as $\bbeta^{*}$ has a finite limit, the situation is essentially the same as a fixed $\bbeta^{*}$. So here we assume $\bbeta^{*}$ to be fixed to simplify the presentation.

Throughout this paper, we use $\dot{g}(\x;\btheta)%
$  and $\ddot{g}(\x;\btheta)%
$ to denote the gradient and Hessian of $g(\x;\btheta)$ with respect to (w.r.t.) $\btheta$, respectively. Let $\|\v\|%
$ be the Frobenius norm and $\v^{\otimes2}=\v\v\tp$ for a vector or a matrix $\v$. %
We denote $\pi(\x,y)$ as the sampling probability for an instance $(\x,y)$. For negative cases, we sometimes use a shorthand notation $\pi(\x)$ to denote its sampling probability.
We use $\op$ to represent a random quantity that converges to $0$ in probability as $N\rightarrow\infty$. %

\subsection{It is the number of positive instances that matters}\label{sec:it-number-positive}
We show that for rare events data, the estimation error rate for $\btheta$ is related to $N_1$ instead of $N$. Using the full data, a commonly used estimator of $\btheta$ is the maximum likelihood estimator (MLE)
\begin{equation*}%
  \htheta_{\f}
  :=\arg\max_{\btheta}\sumN
  \big[y_ig(\x_i;\btheta)-\log\big\{1+e^{g(\x_i;\btheta)}\big\}\big].
\end{equation*}
To investigate the theoretical properties of the estimator $\htheta_{\f}$, we make the following assumptions. 
\begin{assumption}\label{asmp:1}
  The first, second, and third derivatives of $f(\x;\bbeta)$ and $e^{f(\x;\bbeta)}f(\x;\bbeta)$ w.r.t. any components of $\bbeta$
  are bounded by a square intergrable random variable $B(\x)$.
\end{assumption}
\begin{assumption}\label{asmp:2}
  The matrix $\Exp\{\dot{g}^{\otimes2}(\x;\btheta^{*})\}$ is finite and positive definite.
\end{assumption}
Assumption~\ref{asmp:1} imposes constraints on the smoothness of $f(\x;\btheta)$ w.r.t. $\btheta$ and on the tightness of the distribution of $\x$. %
Assumption~\ref{asmp:2} ensures that all components of $\btheta$ are estimable. 

\begin{theorem}\label{thm:1}
  Let $\V_{\f}=\Exp\{e^{f(\x;\bbeta^{*})}\}\M_{\f}^{-1}$ where
$\M_{\f}=\Exp\{e^{f(\x;\bbeta^{*})}\dot{g}^{\otimes2}(\x;\btheta^{*})\}$.
  Under Assumptions~\ref{asmp:1} and \ref{asmp:2}, %
  as $N\rightarrow\infty$,
  \begin{equation*}%
    \sqrt{N_1}(\htheta_{\f}-\btheta^{*}) \longrightarrow \Nor(\0,\ \V_{\f}),
    \quad\text{ in distribution.}
  \end{equation*}
\end{theorem}
This result shows that even if all the $N$ instances are used for model training, the resulting estimator $\htheta_{\f}$ converges to the true parameter $\btheta^{*}$ at the rate of $N_1^{-1/2}$, which is much slower than the usual rate of $N^{-1/2}$ for regular cases. This indicates that with rare events data, the available information about unknown parameters is at the scale of $N_1$.
Although $N\rightarrow\infty$ much faster than $N_1\rightarrow\infty$ (in terms of $N_1/N\rightarrow0$), $N$ does not explicitly show in the asymptotic normality result of Theorem~\ref{thm:1}.
 For the specific case of linear logistic regression with $g(\x;\btheta)=\alpha+\x\tp\bbeta$, our Theorem~\ref{thm:1} reduces to Theorem 1 of \cite{Wang2020RareICML}. 

\subsection{Negative sampling}
Since the available information ties to the number of positive instances instead of the full data size, one can keep all the positive instances and significantly subsmaple the negative instances to reduce the computational cost. For better estimation efficiency, we consider nonuniform sampling. Let $\varphi(\x)>0$ be an integrable function and $\rho$ be the sampling rate on the negative class. %
Without loss of generality, assume $\Exp\{\varphi(\x)\}=1$ so that $\pi(\x_i):=\rho\varphi(\x_i)$ is the sampling probability for the $i$-th data point if $y_i=1$. %
We present the nonuniform negative sampling procedure in Algorithm~\ref{alg1}.  %

\begin{algorithm}[H]
  \caption{Negative sampling}\label{alg1}
For $i=1, ..., N$: 
\-\hspace{0.05cm} if $y_i=1$,\; include $\{\x_i,y_i, \pi(\x_i,y_i)=1\}$ in the sample;\\
\-\hspace{2.55cm} if $y_i=0$,\;
    calculate $\varphi(\x_i)$ and generate $u_i\sim \mathbb{U}(0,1)$;\\
\-\hspace{4.1cm} if $u_i\le\rho\varphi(\x_i)$, include $\{\x_i,y_i, \pi(\x_i,y_i)=\rho\varphi(\x_i)\}$ in the sample.
\end{algorithm}

\section{Weighted estimation and its optimal negative sampling probability}
\label{sec:weight-estim-optim}
Let $\delta_i=1$ if the $i$-th data point is selected and $\delta_i=0$ otherwise. %
Given a subsample taken according to $\pi(\x_i,y_i)=y_i+(1-y_i)\rho\varphi(\x_i)$, $i=1, ..., N$, the IPW estimator of $\btheta$ is %
\begin{equation}\label{eq:4}
  \htheta_w %
  =\arg\max_{\btheta}\sumN\delta_i\pi^{-1}(\x_i,y_i)
  \big[y_ig(\x_i;\btheta)-\log\big\{1+e^{g(\x_i;\btheta)}\big\}\big].
\end{equation}
We need an assumption on the subsampling rate to investigate the subsample estimators.
\begin{assumption}\label{asmp:3}
  The subsampling rate $\rho$ satisfies that 
  $c_N:=e^{\alpha^*}/\rho\rightarrow c$ for a constant $c$ such that $0\le c<\infty$. %
\end{assumption}
 Note that $c_N$ cannot be zero but its limit $c$ can be exactly zero. 
It can be shown that $c_N\Exp\{e^{f(\x;\bbeta^{*})}\} = {N_1}{(N_0\rho)^{-1}}\{1+\op\}$,  %
so $c\Exp\{e^{f(\x;\bbeta^{*})}\}$ is the asymptotic positive/negative ratio for the sample. %
The theorem below shows the asymptotic distribution of $\htheta_w$.
\begin{theorem}\label{thm:2}
  Under Assumptions \ref{asmp:1}-\ref{asmp:3},
  if $\Exp[\{\varphi(\x)+\varphi^{-1}(\x)\}B^2(\x)]<\infty$ then
  as $N\rightarrow\infty$, 
  \begin{equation*}
    \sqrt{N_1}(\htheta_w-\btheta^{*}) \longrightarrow \Nor(\0,\ \V_w),
    \quad\text{ in distribution,}
  \end{equation*}
  where $\V_w=\V_{\f}+\V_{\mathrm{sub}}$ and $\V_{\mathrm{sub}}=c\Exp\{e^{f(\x;\bbeta^{*})}\}\M_{\f}^{-1}
  \Exp\{\varphi^{-1}(\x)e^{2f(\x;\bbeta^{*})}\dot{g}^{\otimes2}(\x;\btheta^{*})\}
  \M_{\f}^{-1}$. 
\end{theorem}
  The subsampled estimator $\htheta_w$ have the same convergence rate as the full data estimator. However, the asymptotic variance $\V_w$ may be inflated by the term $\V_{\mathrm{sub}}$ from subsampling the negative cases. Here $\V_{\mathrm{sub}}$ is zero if $c=0$. 
  If we keep much more negative instances than the positive instances in the subsample ($c=0$), then there is no  asymptotic information loss %
  due to subsampling ($\V_w=\V_{\f}$). In this scenario, the number of negative instances can still be aggressively reduced ($\rho\rightarrow0$) so that the computational efficiency is significantly improved, and the sampling function $\varphi(\x)$ does not play an significant role. If we have to reduce the negative instances to the same level of the positive instances ($0<c<\infty$), then the variance inflation $\V_{\mathrm{sub}}$ is not negligible. In this scenario, %
  a well designed sampling function $\varphi(\x)$ it is more relevant and critical.

  Theorem~\ref{thm:2} shows that the distribution of the estimation error $err:=\sqrt{N_1}(\htheta_w-\btheta^{*})$ is approximated by a normal distribution $\Nor(\0, \V_w)$. Thus for any $\epsilon>0$ the probability of excess error $\Pr(err>\epsilon)$ is %
  approximated by  $\Pr\{\|\Nor(\0,\V_w)\|>\epsilon\}=\Pr\big(\sum_{j=1}^d\lambda_jz_j^2>\epsilon\big)$, where $\lambda_j$'s are eigenvalues of $\V_w$ and  $z_j$'s are independent standard normal random variables. Therefore, a smaller trace of $\V_w$ means a smaller probability of excess error of the same level since $\tr(\V_w)=\sum_{j=1}^d\lambda_j$.

The following theorem gives the optimal negative sampling function for the IPW estimator.
\begin{theorem}\label{thm:3}
  For a given sampling rate $\rho$, the asymptotically optimal $\varphi(\x)$ %
  that minimizes $\tr(\V_w)$ is
\begin{equation}\label{eq:11}
  \varphi_{\os}(\x)=
  \frac{\min\{t(\x;\btheta^{*}),T\}}{\Exp[\min\{t(\x;\btheta^{*}),T\}]},
\end{equation}
where $t(\x;\btheta)=p(\x;\btheta)\|\M_{\f}^{-1}\dot{g}(\x;\btheta)\|$
and $T$ is the maximum number so that $\rho[\min\{t(\x;\btheta^{*}),T\}]\le\Exp[\min\{t(\x;\btheta^{*}),T\}]$ with probability one.
\end{theorem}

\begin{remark}\label{remark:2}
  If $\rho\rightarrow0$ then $\rho t(\x;\btheta^{*})\le\Exp\{t(\x;\btheta^{*})\}$ almost surely, so $T$ can be dropped (i.e., $T=\infty$).   
  If $\lim_{N\rightarrow\infty}\rho>0$ then $c=0$, %
  so the variance inflation due to subsampling is negligible. Thus, the truncation term $T$ can be ignored in practice with imbalanced data; it only plays an role for not very imbalanced data when the sampling ratio is not very small.
\end{remark}

  In $\varphi_{\os}(\x)$, $t(\x;\btheta^{*})$ consists of two components, $p(\x;\btheta^{*})$ and $\|\M_{\f}^{-1}\dot{g}(\x;\btheta^{*})\|$. The term $p(\x;\btheta^{*})$ is from the binary response model structure, and it gives a higher preference for a data point with larger $p(\x;\btheta^{*})$ in the negative class.
  The term $\|\M_{\f}^{-1}\dot{g}(\x;\btheta^{*})\|$ corresponds to the A optimality criterion \citep{pukelsheim2006optimal}.
  For computational benefit in optimal sampling, the A optimality is often replaced by the L optimality \citep[e.g.,][]{WangZhuMa2018,ting2018optimal}.
Under the L optimality, $\|\M_{\f}^{-1}\dot{g}(\x;\btheta^{*})\|$ is replaced by $\|\dot{g}(\x;\btheta^{*})\|$, and this minimizes the asymptotic mean squared error (MSE) of $\M_{\f}\htheta_w$. 
  In case the gradient is difficult to obtain, one could ignore the gradient term and simply use $p(\x;\btheta^{*})$ to replace $t(\x;\btheta^{*})$. Although this does not give an optimal sampling probability, it outperforms the uniform subsampling because it takes into account the information of the model structure like the local case-control (LCC) sampling. 

\section{More efficient estimation based on the likelihood with log odds correction}
\label{sec:more-effic-estim}
\subsection{Nonuniform log odds correction}
The optimal function $\varphi_{\os}(\x)$ assigns larger probabilities to more informative instances. However, the IPW estimator in (\ref{eq:4}) assigns smaller weights to more informative instances in estimation, so the resulting estimator can be improved to have higher estimation efficiency. 
A naive unweighted estimator is biased, and the bias correction approach in \cite{fithian2014local,wang2019more} does not work for the sampling procedure in Algorithm~\ref{alg1} even when the underlying model is the logistic regression. We adopted the idea of \cite{han2020local} and seek to define a more efficient estimator through finding the corrected likelihood of the sampled data based on any negative sampling probability $\pi(\x)$.
For data included in the subsample (where $\delta=1$), the conditional probability of $y=1$ is
\begin{equation}\label{eq:2}
  p_{\pi}(\x;\btheta)
    :=\Pr(y=1\mid\x,\delta=1)
    =\frac{1}{1+e^{-g(\x;\btheta)-l}},
  \end{equation}
  where $l=-\log\{\pi(\x)\}$. Please see the detailed derivation of (\ref{eq:2}) in the supplement.
To avoid the IPW, we propose the estimator based on the log odds corrected likelihood in (\ref{eq:2}), namely, 
\begin{equation}\label{eq:5}
  \htheta_{\lik}=\arg\max_{\btheta}\ell_{\lik}(\btheta),
  \quad\text{where}\quad
  \ell_{\lik}(\btheta)
  =\sumN\delta_i\big[y_ig(\x_i;\btheta)
  -\log\big\{1+e^{g(\x_i;\btheta)+l_i}\big\}\big].
\end{equation}
With $\htheta_w$ in (\ref{eq:4}), $\pi(\x_i)$ is in the inverse. If an instance with much smaller $\pi(\x_i)$ is selected in the sample, it dominates the objective function, making the resulting estimator unstable. With $\htheta_{\lik}$, this problem is ameliorated because $\pi(\x_i)$ is in the logarithm in the log-likelihood of (\ref{eq:5}) and $\log(v)$ goes to infinity much slower than $v^{-1}$ as $v\downarrow0$.

\subsection{Theoretical analysis of the likelihood-based estimator}

The following shows the asymptotic distribution of $\htheta_{\lik}$.
\begin{theorem}\label{thm:4}
  Under Assumptions \ref{asmp:1}-\ref{asmp:3},
  if $\Exp\{e^{f(\x;\bbeta^{*})}\varphi^{-1}(\x)B(\x)\}<\infty$,
  then
as $N\rightarrow\infty$, 
\begin{equation*}
  \sqrt{N_1}(\htheta_{\lik}-\btheta) \longrightarrow \Nor(\bm{0},\V_{\lik})
  \quad\text{ in distribution,}
\end{equation*}
where $\V_{\lik}=\Exp\{e^{f(\x;\bbeta^{*})}\}\bLambda_{\lik}^{-1}$ and
$\bLambda_{\lik}=\Exp\Big[\frac{e^{f(\x;\bbeta^{*})}\dot{g}^{\otimes2}(\x;\btheta^{*})}
{1+c\varphi^{-1}(\x)e^{f(\x;\bbeta^{*})}}\Big]$.
\end{theorem}
Theorem~\ref{thm:3} shows that the proposed estimator $\htheta_{\lik}$ is asymptotic normal with variance $\V_{\lik}$. We compare $\V_{\lik}$ with $\V_w$ to show that $\htheta_{\lik}$ has a higher estimation efficiency than $\htheta_w$.
\begin{theorem}\label{thm:5}
If $\V_w$ and $\V_{\lik}$ are finite and positive definite, then
$\V_{\lik}\le\V_w$, i.e., $\V_w-\V_{\lik}$ is non-negative definite. The equality holds when $c=0$ and in this case $\V_{\lik}=\V_w=\V_{\f}$.
\end{theorem}

Since the estimator $\htheta_{\lik}$ in Eg.~(\ref{eq:5}) is based on the conditional log-likelihood of the sampled data, it actually has the highest estimation efficiency among a class of asymptotically unbiased estimators. %
 
\begin{theorem}\label{thm:6}
  Let $\X=(\x_1,...,\x_N)\tp$ be the feature matrix and $\Dd$ denote the sampled data. Let 
  $\htheta_U$ be a subsample estimator with the following asymptotic representation:
  \begin{equation}\label{eq:24}
    \htheta_U=\U(\btheta^{*};\Dd)+N_1^{-1/2}\op,
  \end{equation}
  where the variance $\Var\{\U(\btheta^{*};\Dd)\}$ exists, and $\U(\btheta^{*};\Dd)$ satisfies that
  $\Exp\{\U(\btheta^{*};\Dd)\mid\X\}=\btheta^{*}$  and
  $\Exp\{\dot\U(\btheta^{*};\Dd)\mid\X\}=\0$ with  $\dot\U(\btheta^{*};\Dd)=\partial\U(\btheta^{*};\Dd)/\partial{\btheta^{*}}\tp$. If $N_1\Var\{\U(\btheta^{*};\Dd)\}\rightarrow\V_U$ in probability, then $\V_U\ge\V_{\lik}$.
\end{theorem}

\begin{remark}
  Theorem~\ref{thm:6} tells us that $\htheta_{\lik}$ is statistically the most efficient among a class of estimators, which includes both $\htheta_w$ and $\htheta_{\lik}$ as special cases. 
  The condition $\Exp\{\U(\btheta^{*};\Dd)\mid\X\}=\btheta^{*}$ ensures that the estimator is asymptotically unbiased. The condition $\Exp\{\dot\U(\btheta^{*};\Dd)\mid\X\}=\0$ can be intuitively interpreted as the requirement that the derivative of the $\op$ term in (\ref{eq:24}) is also negligible. 
  Clearly, this is satisfied by all unbiased estimators for which the $\op$ term in (\ref{eq:24}) is $\0$. 
\end{remark}

\section{Practical considerations}
\label{sec:practical}
In this section, we first show how we design a more practical estimator based on the previous results and then present the theoretical analysis behind these improvements.
\subsection{Making estimators more practical}
Like the LCC sampling \citep{fithian2014local} and the A-optimal sampling \citep{wang2019more}, the $\varphi_{\os}(\x)$ in (\ref{eq:11}) depends on $\btheta$, and thus a pilot value of $\btheta$, say $\ttheta$, is required in practice. Here, $\ttheta$ can be constructed from a pilot sample, say $(\tilde{\x}_i,\tilde{y}_i)$, $i=1, ..., \tilde{n}$, taken by uniform sampling from both classes with equal expected sample sizes. 
With $\ttheta$ obtained, calculate 
$\tilde{t}_i=p(\tilde\x_i;\tilde\btheta)\|\tilde\M_{\f}^{-1}\dot{g}(\tilde\x_i;\tilde\btheta)\|$
for $i=1, ..., \tilde{n}$, where $\tilde\M_{\f}$ is the Hessian matrix of the pilot sample objective function. 
In case that the Hessian matrix and gradients are hard to record, the gradient term can be dropped, i.e., use $\tilde{t}_i=p(\tilde\x;\ttheta)$. As mentioned in Remark~\ref{remark:2}, $T$ can be ignored for very imbalanced data. 
The expectation in the denominator of (\ref{eq:11}) can be approximated by mimicking the method of moment estimator, namely by
\begin{equation}\label{eq:9}\textstyle
  \tilde\omega= 2N_1(\tilde{n}N)^{-1}\sum_{\tilde{y}_i=1}\tilde{t}_i
  +2N_0(\tilde{n}N)^{-1}\sum_{\tilde{y}_i=0}\tilde{t}_i.
\end{equation}
In practice, it is common to set a lower threshold, say $\varrho$, on sampling probabilities, to ensure that all instances have positive probabilities to be selected. This is more critical for the IPW estimator while the likelihood estimator is not very sensitive to very small sampling probabilities. 

Denote the pilot value as $\tilde\bvtheta=(\ttheta, \tilde\omega)$. 
The following probabilities can be practically implemented
\begin{equation}\label{eq:8}
  \pi_{\varrho}^{\os}(\x;\tilde\bvtheta)
  =\min[\max\{\rho\tilde\varphi_{\os}(\x),\varrho\},1],
  \quad\text{ where }\quad
  \tilde\varphi_{\os}(\x)=\tilde\omega^{-1}t(\x;\ttheta),
\end{equation}
and we use this in our experiments in Section~\ref{sec:numer-exper}. The $\tilde\omega$ in (\ref{eq:9}) is essentially a normalizing term so that the expected subsample size is $\rho$. For some practical systems such as online models, %
$\tilde\omega$ is often treated as a tuning parameter. We will illustrated this in our real data example in Section~\ref{sec:exper-real-data}.

\subsection{Theoretical analysis of practical estimators}
Denote $\htheta_w^{\tilde\bvtheta}$ and $\htheta_{\lik}^{\tilde\bvtheta}$ the IPW estimator and the likelihood estimator, respectively, with practically estimated optimal probability in (\ref{eq:8}). We have the following results.

\begin{theorem}\label{thm:7}
  Assume that the pilot estimator $\tilde\bvtheta$ is independent of the data, %
  $\tilde\varphi_{\os}(\x)\rightarrow\varphi_{\plt}(\x)$ in probability, and %
  $\rho^{-1}\varrho\rightarrow c_l>0$. Under Assumptions \ref{asmp:1}-\ref{asmp:3}, as $N\rightarrow\infty$,
the $\htheta_w^{\tilde\bvtheta}$ satisfies that, 
\begin{equation*}
  \sqrt{N_1}(\htheta_w^{\tilde\bvtheta}-\btheta)
  \rightarrow \Nor(\bm{0},\V_w^{\plt}) \quad\text{in distribution,}
\end{equation*}
 where $\V_w^{\plt}$ has the same expression of $\V_w$ except that $\varphi(\x)$ is replaced by $\max\{\varphi_{\plt}(\x),c_l\}$. If $\Pr\{\varphi_{\plt}(\x)\ge c_l\}=1$ and the pilot estimator is consistent, then $\V_w^{\plt}$ achieves the optimal variance. 
\end{theorem}

\begin{theorem}\label{thm:8}
  Assume that the pilot estimator $\tilde\bvtheta$ is independent of the data, %
  $\tilde\varphi_{\os}(\x)\rightarrow\varphi_{\plt}(\x)$ in probability, and %
  $\varrho=o(\rho)$.
    Under Assumptions \ref{asmp:1}-\ref{asmp:3}, as $N\rightarrow\infty$,
$\htheta_{\lik}^{\tilde\bvtheta}$ satisfies that,
\begin{equation*}
  \sqrt{N_1}(\htheta_{\lik}^{\tilde\bvtheta}-\btheta)
  \rightarrow \Nor\big(\bm{0},\V_{\lik}^{\plt}\big)
  \quad\text{in distribution,}
\end{equation*}
where
$\V_{\lik}^{\plt}$ has the same expression of $\V_{\lik}$ except that $\varphi(\x)$ is replaced by $\varphi_{\plt}(\x)$. If the pilot estimator is consistent, then $\V_{\lik}^{\plt}$ achieves the variance with the optimal probability. 
\end{theorem}

\begin{remark}
  Theorems~\ref{thm:7} and \ref{thm:8} do not require $\tilde\bvtheta$ to be consistent, i.e., the pilot estimator can be misspecified. However $\tilde\bvtheta$ has to be consistent in order to achieve the asymptotic variances with the optimal probability. Compared with the likelihood estimator $\htheta_{\lik}^{\tilde\bvtheta}$, the IPW estimator $\htheta_w^{\tilde\bvtheta}$ requires a stronger condition on $\varrho$ to have asymptotic normality, because $\pi_{\varrho}^{\os}(\x;\tilde\bvtheta)$ is in the denominator of the objective function for $\htheta_w^{\tilde\bvtheta}$ so the turbulence of $\tilde\bvtheta$ is amplified.  %
  The requirement $c_l>0$ means that $\varrho$ cannot be too small %
  in practice. The likelihood estimator does not have this constraint. 
\end{remark}

\section{Numerical experiments}
\label{sec:numer-exper}

We present simulated results and report the performance on a CTR dataset with more than 0.3 trillion instances. More experiments with model and pilot misspecifications are available in the supplement.

\subsection{Experiments on simulated data}
\label{sec:exper-simul-data}
We implemented the following negative sampling methods for logistic regression. 1) uniW: uniform sampling with the IPW estimator. 2) uniLik: uniform sampling with the likelihood estimator. 3) optW: optimal sampling with the IPW estimator. 4) optLik: optimal sampling with the likelihood estimator. We also implement the full data MLE (Full) and the LCC for comparisons.

In each repetition of the simulation, we generate full data of size $N=5\times10^5$ from the logistic regression with $g(\x;\btheta)=\alpha+\x\tp\bbeta$. We set the true $\bbeta^{*}$ to be a $6\times1$ vector of ones and set different values for $\alpha$ for different feature distributions so that the positive/negative ratio is close to 1:400.
We consider the following feature distributions: %
(a) Normal distribution which is symmetric with light tails; the true intercept is $\alpha=-7.65$. 
(b) Log-normal distribution which is asymmetric and positively skewed; the true intercept is $\alpha=-0.5$.   
(c) $t_3$ distribution which is symmetric with heavier tails; the true intercept is $\alpha=-7$.

We set the sampling rate as $\rho= 0.002, 0.004, 0.006, 0.01$, and $0.02$
for all sampling methods.  %
In each repetition, uniform samples of average size $100$ are selected from each class to calculate the pilot estimates, so the uncertainty due to the pilot estimates are taken into account.
We also consider pilot misspecification by adding a uniform random number from $\mathbb{U}(0,1.5)$ to the pilot so that the pilot is systematically different from the true parameter. 
We repeat the simulation for %
$R=1000$ times to calculate the MSE as $R^{-1}\sum_{r=1}^R\|\htheta^{(r)}-\btheta\|^2$, where $\btheta^{(r)}$ is the estimate at the $r$-th repetition.

\begin{figure}[htb]\centering
  \begin{minipage}{0.32\textwidth}\centering
    \includegraphics[width=\textwidth,page=1]{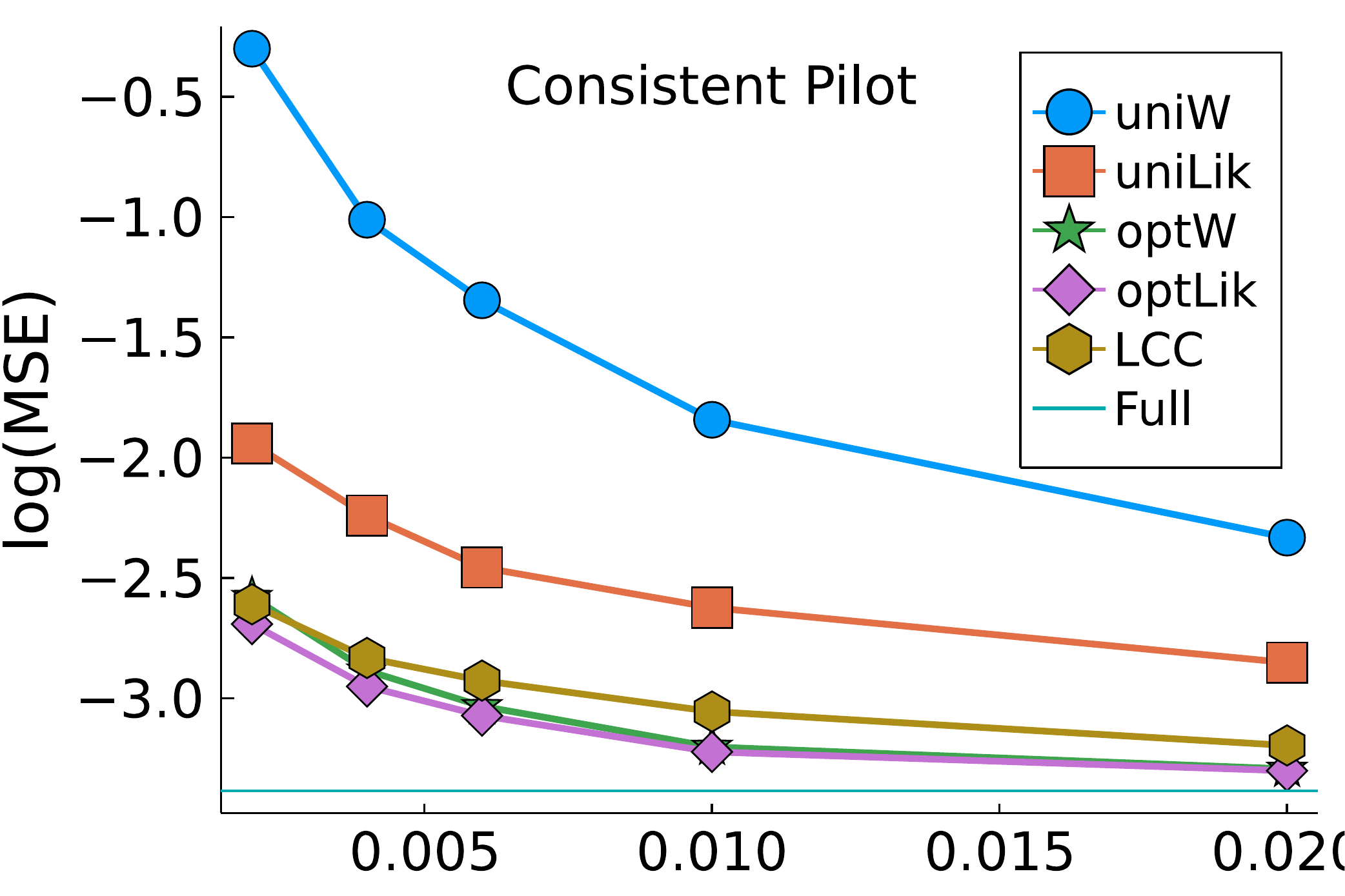}
    \includegraphics[width=\textwidth,page=1]{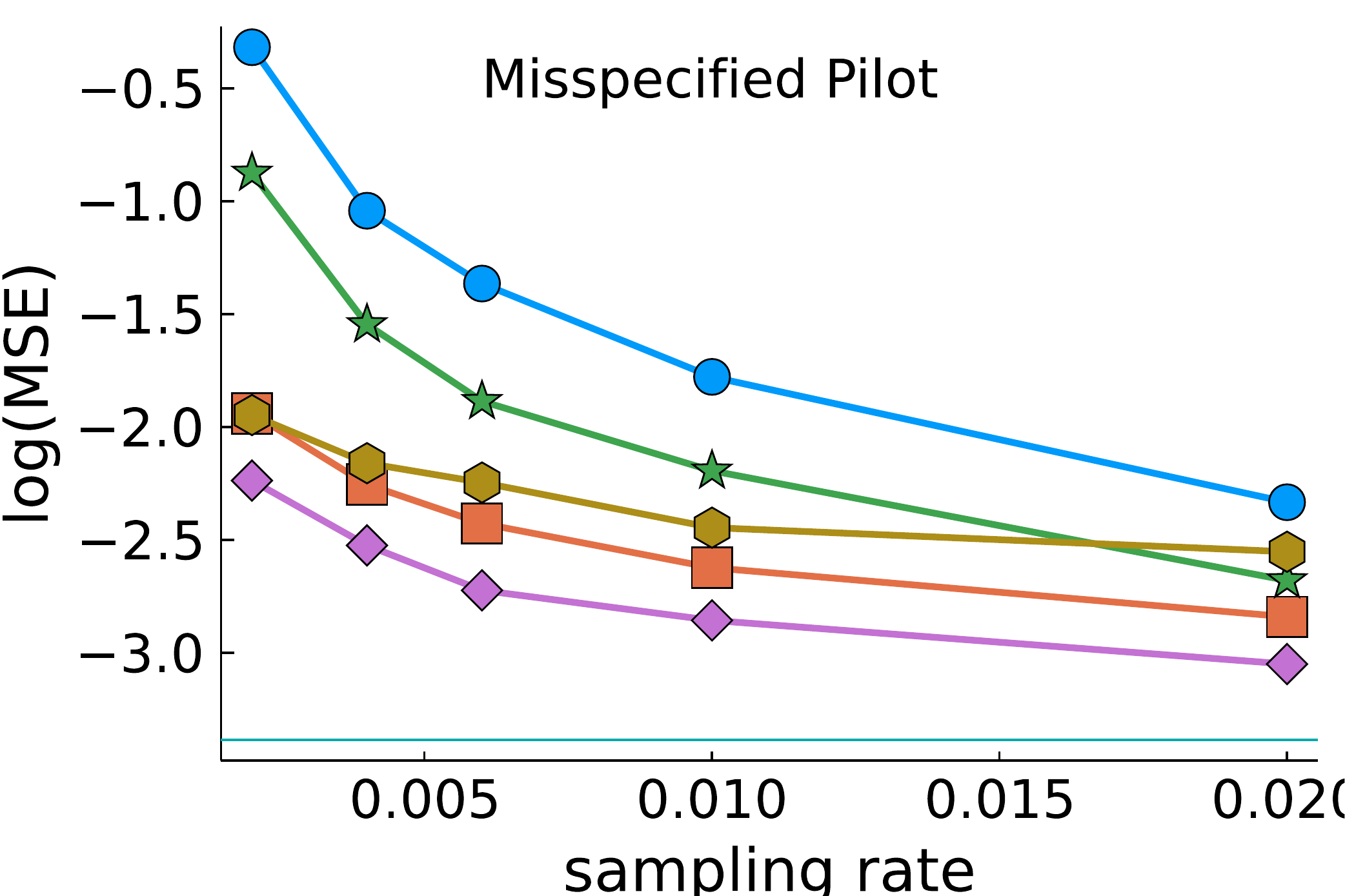}
    (a). {$\x$'s are normal}
  \end{minipage} %
  \begin{minipage}{0.32\textwidth}\centering
    \includegraphics[width=\textwidth,page=2]{figures/00mse.pdf}
    \includegraphics[width=\textwidth,page=2]{figures/00mseMisPilot.pdf}
    (b) {$\x$'s are lognormal}
  \end{minipage} %
  \begin{minipage}{0.32\textwidth}\centering
    \includegraphics[width=\textwidth,page=3]{figures/00mse.pdf}
    \includegraphics[width=\textwidth,page=3]{figures/00mseMisPilot.pdf}
    (c) {$\x$'s are $t_3$}
  \end{minipage}
  \caption{Log(MSEs) for different estimators (the smaller the better). The top row uses consistent pilot estimators; the bottom uses misspecified pilot estimators.}
  \label{fig:logistic1}
\end{figure}

Results are presented in Figure~\ref{fig:logistic1}. The MSEs of all estimators decrease as the sampling rate $\rho$ increases. The optLik outperforms other sampling methods in general, and its performance is close to the full data MLE when $\rho=0.02$, especially if the pilot is consistent.
The advantage of optLik over optW is more evident for smaller $\rho$.
If the pilot is misspecified, optLik is quite robust but optLik and LCC significantly deteriorate. Since uniLik and uniW do not require a pilot, they are not affected by pilot misspecification, but they have much larger MSE than optLik and optW when the pilot is consistent, especially uniW.
Different feature distributions also affect the performances of different sampling methods.

To verify Theorem~\ref{thm:1} numerically, we let the full data size $N$ increase at a faster rate than the average number of positive instances $N_1^a=\Exp(N_1)$ so that the probability $\Pr(y=1)$ decreases towards zero. As $N$ increases, we fixed the value of $\bbeta^*$ and set decreasing values to $\alpha^*$ to mimic our scaling regime. %
We considered two covariate distributions: 1) a multivariate normal distribution for which a logistic regression model is a correct model; and 2) a multivariate log-normal distribution for which a logistic regression model is misspecified. %
We simulated for 100 times to calculate the empirical variance $\hat{\V}_f$ of the full data MLE, and report the results in Table~\ref{tab:1}.  
It is seen that $\tr(\hat{\V}_f)$ is decreasing towards zero. 
According to Theorem~\ref{thm:1}, $\tr(\hat{\V}_f)$ should converge in a rate of $1/N_1$, so $N_1^a\tr(\hat{\V}_f)$ should be relatively stable as $N$ increases. This is indeed the case as seen in the third and sixth columns of Table~\ref{tab:1}. On the other hand, $N\tr(\hat{\V}_f)$ gets large dramatically as $N$ increases. The aforementioned observations confirm the theoretical result in Theorem~\ref{thm:1}. Furthermore, Table~\ref{tab:1} shows that the convergence rate in Theorem~\ref{thm:1} may also be true for some misspecified models. 
\begin{table}[htbp]%
  \caption{Empirical variances of the full data MLE for different full data size and average number of positive instances combinations.}\label{tab:1}
\centering
\begin{tabular}{lccrcccr}\hline
& \multicolumn{3}{c}{Correct model} & & \multicolumn{3}{c}{Mis-sprcified model} \\
  \cline{2-4} \cline{6-8}\\[-3.5mm]
  ($N$, $N_1^a$) & $\tr(\hat{\V}_f)$ & $N_1^a\tr(\hat{\V}_f)$ & $N\tr(\hat{\V}_f)$
  && $\tr(\hat{\V}_f)$ & $N_1^a\tr(\hat{\V}_f)$ & $N\tr(\hat{\V}_f)$\\\hline
($10^3$, { } $32$) & 0.169 & 5.41 & 169.17  && 0.969 & 30.99 & 968.70 \\
($10^4$, { } $64$) & 0.097 & 6.20 & 969.29  && 0.322 & 20.59 & 3217.12\\
($10^5$, $128$) & 0.045 & 5.76 & 4497.24 && 0.135 & 17.32 & 13527.60\\
($10^6$, $256$) & 0.018 & 4.62 & 18048.40&& 0.046 & 11.74 & 45847.40\\\hline
\end{tabular}
\end{table}

\subsection{Experiments on real data}
\label{sec:exper-real-data}

We conduct experiments on {an internal CTR dataset from our products} %
with more than 0.3 trillion instances. There are 299 input features, including user and ad profiles together with rich context features. We concatenate all features as a single high-dimensional vector $\x$, whose length is greater than $5,000$. We use a 3-layer fully connected deep neural network as a feature extractor that maps $\x$ into a 256-dimensional dense features $h(\x)$. The downstream model is a linear logistic regression model taking $h(\x)$ as input to predict the binary output whether the user clicks the ad or not.

The positive/negative ratio is around 1:80. 
Due to limited storage, we first uniformly drop 80\% negative instances and try various negative sampling algorithms on the rest 60 billion instances. On this pre-processed dataset, we then split out 1 percent of the data for testing, and do negative sampling on the rest data. The entire training procedure scans the subsampled data set from begin to end in one-pass according to the timestamp. The testing is done all the way through training. We calculate the area under the ROC curve (AUC) for  testing instances. The entire model is trained on a distributed machine learning platform using 1,200 CPU cores and can finish within two days.

We adopt the sampling probability $\pi_{\varrho}^{\os}(\x;\tilde\bvtheta)$
as demonstrated in~(\ref{eq:8}) and use the empirical sampling rate to calibrate the normalizing term $\tilde{\omega}$. As mentioned before, we remove %
the gradient term for implementation tractability so $t(\x;\ttheta)=p(\x;\ttheta)$. As the result will show, this approximate score produces consistent results with the simulation experiments. We compare four methods: 1) uniW and 2) uniLik are the same as in simulated experiments; and 3) optW and 4) optLik represent nonuniform sampling probability using $\pi_{\varrho}^{\os}(\x;\tilde\bvtheta)$
with the IPW estimator and with the likelihood estimator. We use ``opt" to refer to both optW and optLik when focusing on the sampling.

\begin{figure}[htb]
\centering
  \begin{minipage}{0.49\linewidth}\centering
    \includegraphics[width=\linewidth]{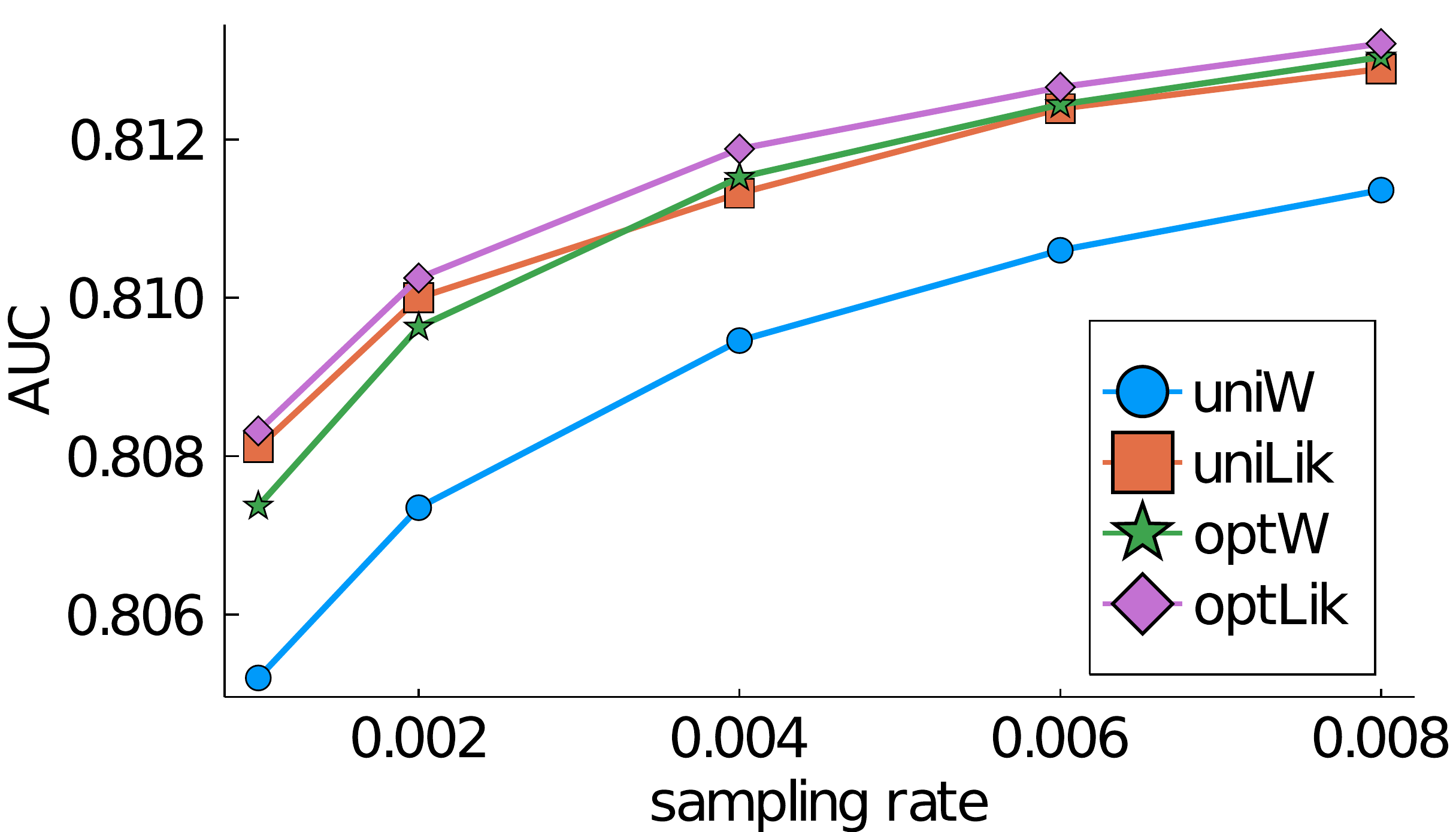}
    (a). {\small Extremely small sampling rates.}
  \end{minipage}
  \begin{minipage}{0.49\linewidth}\centering
    \includegraphics[width=\linewidth]{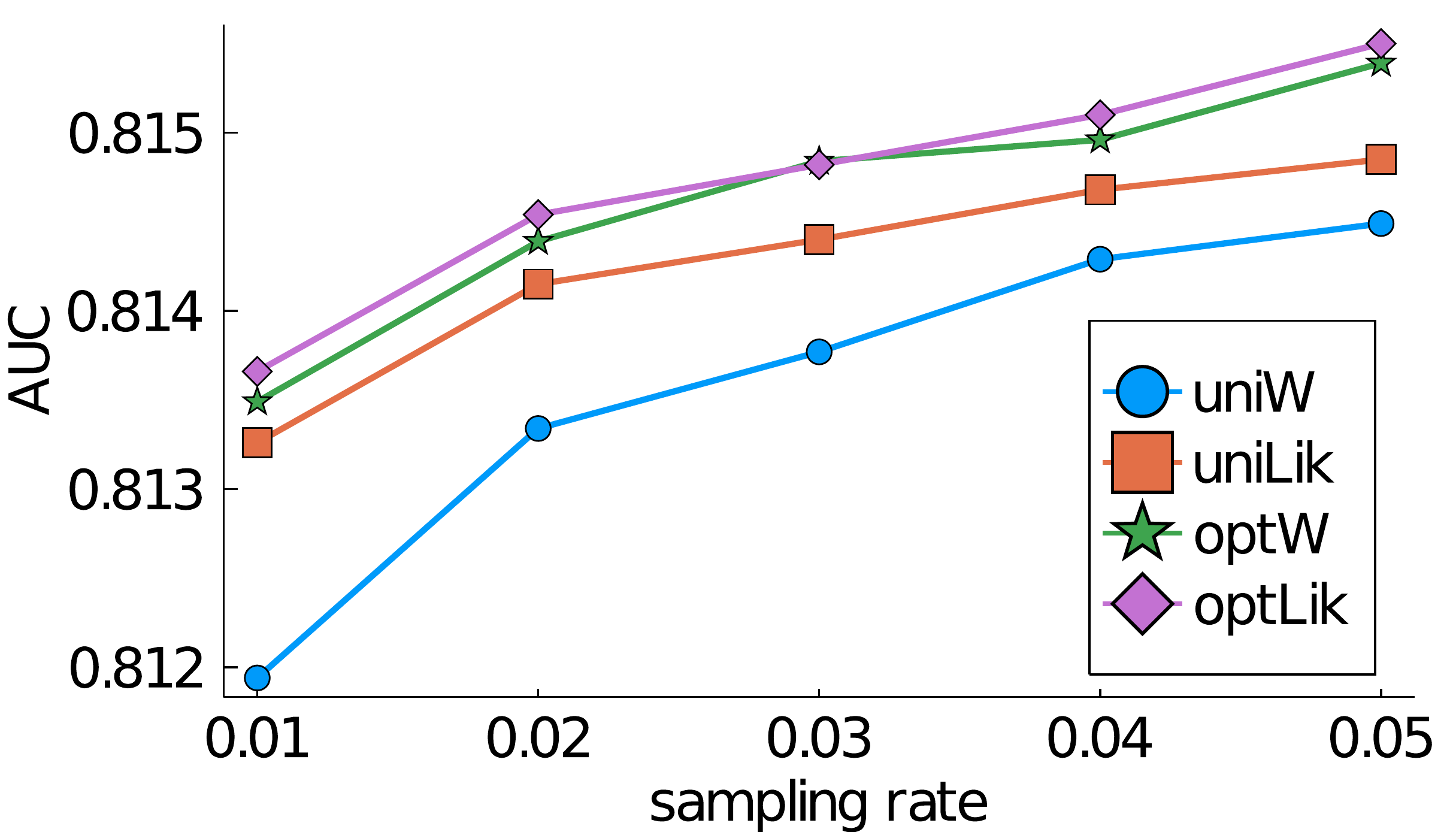}
    (b) {\small Moderate sampling rates.}
  \end{minipage}
  \caption{Empirical testing AUC of subsample estimators for different sample sizes (the larger the better).}
  \label{fig:lcc1}
\end{figure}

We study two sets of negative sample rates (w.r.t. the original 0.3 trillion data): (i) $[0.01, 0.05]$. In this case, negative instances still outnumber positive instances. (ii) $[0.001, 0.008]$. In this case, negative subsampling is too aggressive that positive instances dominates negative instances. We demonstrate the result in Figure~\ref{fig:lcc1}. When the sample rate is moderate (Figure~\ref{fig:lcc1}.(b)), opt is nearly optimal and can significantly outperform uniform subsampling. There is still a small gap between the IPW estimator and the likelihood estimator. When the sample rate is extremely small (Figure~\ref{fig:lcc1}.(a)), %
{opt becomes sub-optimal and its gap w.r.t. uniform sampling is closer. Still we see a clear difference between the IPW estimator and the likelihood estimator. This is due to a larger asymptotic variance for uniform sampling when the sampling rate goes to zero.} Note that on this huge data set, a small relative AUC gain (e.g., 0.0005) usually could bring about significant revenue gain in products.

\subsection{Sensitivity analysis}\label{sec:sensitivity-analysis}
\begin{figure*}[htp]
\centering%
    \includegraphics[width=\linewidth]{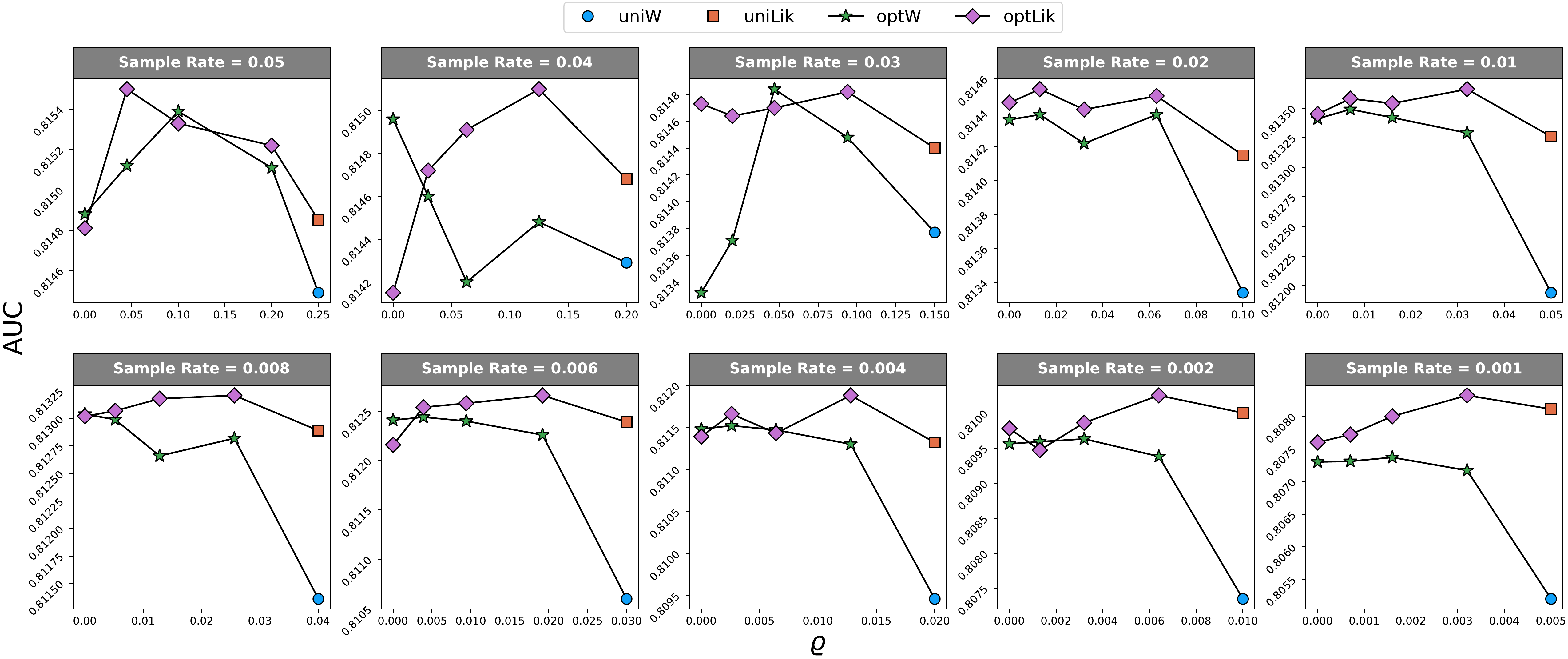}\vspace{-20pt}
  \caption{AUC for IPW and likelihood estimators when tuning the truncation lower bound $\varrho$.}
  \label{fig:sensitivity}
\end{figure*}

In practice, we find it necessary to use cross-validation to find the optimal $\varrho$ in (\ref{eq:8}) given a fixed sample rate. As demonstrated in Figure~\ref{fig:sensitivity}, for each sampling rate we tune $\varrho$ from a very small value (in this case, $10^{-6}$) all the way to its largest value. %
When $\varrho$ achieves its largest value, $\rho\tilde{\omega}^{-1}=0$ and $\pi_{\varrho}^{\os}(\x;\tilde\bvtheta)$ reduces to uniform negative sampling. We observe that $\pi_{\varrho}^{\os}(\x;\tilde\bvtheta)$ with a moderate $\varrho$ achieves the best results for various sampling rates. Note that $\varrho$ is the lower truncation level for negative sampling probabilities. Thus, a larger $\varrho$ means more ``simple" negative instances ($\rho\tilde{\omega}^{-1}p(\x_i;\ttheta)$ is smaller than $\varrho$) will have better chances to be selected. This coincides with recent empirical results for negative subsampling in large search systems~\citep{huang2020embedding}, where they mix ``simple" (smaller $p(\x_i;\ttheta)$) and ``hard" (larger $p(\x_i;\ttheta)$) negative instances. We also observe an empirical variance of around $0.0001$
for each setup, demonstrating that the relative
improvement is consistent.

\section{Conclusions, discussion, and limitations}
\label{sec:conclusions}
In this paper, we have derived asymptotic distributions for full data and subsample estimators with rare events data. We have also found the optimal sampling probability that minimizes the unconditional asymptotic variance of the IPW estimator, and proposed the likelihood estimator through nonuniform log odds correction to further improve the estimation efficiency. The optimal probability depends on unknown parameters so we have discussed practical implementations and investigated the convergences of the resultant estimators. Experiments on both simulated and real big online streaming data confirm our theoretical findings.
Heuristically, our nonuniform negative sampling method gives preference to data points that are harder to observe, so it would give preference to sub-populations that are difficult to observe. We do not think this would bring any negative societal impacts. We have also examined our real data application and did not notice discrimination against any sub-populations.

This paper has the following limitations: 1) We assume that the model is correctly specified. Theoretical properties with model misspecification is not considered, and this important question requires future studies. 2) Oversampling the rare positive instances is another common practice to deal with imbalanced data. Its application together with negative sampling is not considered. 3) The assumption of a fixed $\bbeta^{*}$ means that both the marginal and conditional probabilities for a positive instance to occur are small. If the proportion of the positive instances for a given feature in a dataset is large, this assumption may not be appropriate. Further investigation is need to see if our results is still applicable. 
In the following, we present three scenarios that the scaling regime may or may not fit in:

Scenario 1 (phone call while driving) Car crashes occur with small probabilities, and making phone calls while driving significantly increases the probability of a car crash. However, the probability of car crashes among people making phone calls while driving is still small, so for these types of features, our scaling regime is appropriate to model the rare events.

Scenario 2 (anti-causal learning) Anti-causal learning \cite{scholkopf2012causal,kilbertus2018generalization} assumes that label ($y$) causes observations ($x$). Thus $\Pr(x|y)$ represents the causal mechanism and is fixed. One standard example is that diseases ($y$) cause symptoms ($x$). Our scaling regime fits the framework of anti-causal learning. To see this, using Bayes Theorem we write the log odds as 
\begin{equation*}
 \alpha+f(\x;\bbeta)=\log\frac{\Pr(y=1)}{\Pr(y=0)} + \log\frac{\Pr(\x\mid y=1)}{\Pr(\x\mid y=0)}. 
\end{equation*}
In anti-causal learning, the marginal distribution of $y$ changes while the conditional distribution of $\x$ given $y$ is fixed. Thus only the scale factor $\alpha$ changes, and $f(\x;\bbeta)$ is fixed.

Scenario 3 (local transmission of the COVID-19) Our scaling regime has some limitations and does not apply to all types of rare events data. For example, although the COVID-19 infection rate for the whole population is low, the infection rate for people whose family members are in the same house with positive test results is high. This means the change of a family member's test result converts a small-probability-event to a large-probability-event, and our scaling regime would not be appropriate.

\begin{ack}

HaiYing Wang's research was partially supported by NSF grant CCF 2105571.

\end{ack}

\newpage
\begin{center}\Large\bf
Proofs and Additional Numerical Experiments
\end{center}
\appendix
\setcounter{equation}{0}
\renewcommand{\theequation}{S.\arabic{equation}}
\renewcommand{\thesection}{S.\arabic{section}}
\renewcommand{\thefigure}{S.\arabic{figure}}
\renewcommand{\thetable}{S.\arabic{table}}
\renewcommand{\thetheorem}{S.\arabic{theorem}}

Before presenting the proof, we point out that Assumption~\ref{asmp:2} implies that 
$\Exp\{v(\x)\dot{g}^{\otimes2}(\x;\btheta)\}$ is positive definite for a positive function $v(\x)>0$ almost surely. This is because Assumption~\ref{asmp:2} means that $\Pr\{l\tp\dot{g}(\x;\btheta)\neq0\}>0$ for any $l\neq\0$, and therefore $\Pr\{v^{1/2}(\x)l\tp\dot{g}(\x;\btheta)\neq0\}>0$ for any $l\neq\0$, implying that $\Exp\{v(\x)\dot{g}^{\otimes2}(\x;\btheta)\}$ is positive definite. We also point out that if a sequence is $\op$ conditionally then it is also $\op$ unconditionally and vice versa \citep{xiong2008some,cheng2010bootstrap}. 

To facilitate the presentation, define $a_N=\sqrt{Ne^{\alpha^{*}}}$. We notice that 
\begin{equation}\label{eq:26}
  N_1%
  =N\Exp\bigg\{\frac{e^{\alpha^{*}+f(\x;\bbeta^{*})}}
  {1+e^{\alpha^{*}+f(\x;\bbeta^{*})}}\bigg\}\{1+\op\}
  =a_N^2\Exp\{e^{f(\x;\bbeta^{*})}\}\{1+\op\},
\end{equation}
from the dominated convergence theorem. Thus the normalizing term $\sqrt{N_1}$ for the asymptotic normality in the paper can be replaced by $a_N\Exp^{1/2}\{e^{f(\x;\bbeta^{*})}\}$.

\section{Proof of Theorem~\ref{thm:1}}
  The estimator $\htheta_{\f}$ is the maximizer of 
\begin{equation*}
  \ell(\btheta)
  =\sumN\big[y_ig(\x_i;\btheta)
  -\log\{1+e^{g(\x_i;\btheta)}\}\big],
\end{equation*}
so $a_N(\hat\btheta-\btheta^{*})$ 
is the maximizer of
\begin{align*}
  \gamma(\u)=\ell(\btheta^{*}+a_N^{-1}\u)-\ell(\btheta^{*}).
\end{align*}
By Taylor's expansion,
\begin{align*}
  \gamma(\u)
  &=a_N^{-1}\u\tp\dot\ell(\btheta^{*})
    +0.5a_N^{-2}\sumN
    \phi(\x_i;\btheta^{*})
    \{\u\tp\dot{g}(\x_i;\btheta^{*})\}^2
    +\Delta_{\ell}+R,
\end{align*}
where
\begin{equation*}
  \dot\ell(\btheta)=\frac{\partial\ell(\btheta)}{\partial\btheta}
  =\sumN\big\{y_i-p(\x_i;\btheta)\big\}\dot{g}(\x_i;\btheta),
\end{equation*}
\begin{align*}
  \Delta_{\ell}=\frac{1}{2a_N^2}\sumN\big\{y_i-p(\x_i;\btheta^{*})\big\}
  \u\tp\ddot{g}(\x_i;\btheta^{*})\u,
\end{align*}
and $R$ is the remainder term. By direct calculation,
\begin{align*}
  R
  &=\frac{1}{6a_N^3}\sumN
    \phi(\x_i;\btheta^{*}+a_N^{-1}\Acute\u)
    \big\{1-2p(\x_i;\btheta^{*}+a_N^{-1}\Acute\u)\big\}
    \{\u\tp\dot{g}(\x_i;\btheta^{*}+a_N^{-1}\Acute\u)\}^3\\
  &+\frac{1}{6a_N^3}\sumN
    \phi(\x_i;\btheta^{*}+a_N^{-1}\Acute\u)
    \{\u\tp\dot{g}(\x_i;\btheta^{*}+a_N^{-1}\Acute\u)\}
    \{\u\tp\ddot{g}(\x_i;\btheta^{*}+a_N^{-1}\Acute\u)\u\}\\
  &+\frac{1}{6a_N^3}\sumN\bigg[\big\{y_i-p(\x_i;\btheta^{*}+a_N^{-1}\Acute\u)\big\}
    \sum_{j_1,j_2=1}^du_{j_1}u_{j_2}
    \dddot{g}_{j_1j_2}(\x_i;\btheta^{*}+a_N^{-1}\Acute\u)\u\bigg],
\end{align*}
where $\dddot{g}_{j_1j_2}(\x;\btheta)$ is the gradient of $\ddot{g}_{j_1j_2}(\x,\btheta)$ and $\Acute\u$ lies between $\0$ and $\u$.
We see that
\begin{align}
  |R|
  &\le\frac{\|\u\|^3}{6a_N^3}\sumN
    p(\x_i;\btheta^{*}+a_N^{-1}\Acute\u) C(\x_i,\btheta^{*}+a_N^{-1}\Acute\u)
  +\frac{d\|\u\|^3}{6a_N^3}\sumN y_i B(\x_i)\notag\\ %
  &\le\frac{\|\u\|^3e^{a_N^{-1}\|\u\|}}{6Na_N}
    \sumN \exp\{f(\x_i;\bbeta+a_N^{-1}\Acute\u_{(-1)})\}
    C(\x_i,\btheta^{*}+a_N^{-1}\Acute\u)
  +\frac{d\|\u\|^3}{6a_N^3}\sumN y_i B(\x_i)\notag\\ %
  &\le\frac{d^3\|\u\|^3e^{a_N^{-1}\|\u\|}}{6Na_N}
    \sumN B(\x_i) %
    +\frac{d\|\u\|^3}{6a_N^3}\sumN y_i B(\x_i)
  \equiv\Delta_{R1}+\Delta_{R2},\label{eq:12}
\end{align}
where $C(\x,\btheta)=\|\dot{g}(\x,\btheta)\|^3+\|\dot{g}(\x,\btheta)\|
\|\ddot{g}(\x,\btheta)\|+\sum_{j_1j_2}^d\|\dddot{g}_{j_1j_2}(\x,\btheta)\|$ and $\Acute\u_{(-1)}$ is $\Acute\u$ with the first element removed. From Assumption~\ref{asmp:1}, $\Exp(\Delta_{R1})\rightarrow0$ and
\begin{align*}
  \Exp(\Delta_{R2})
  \le\frac{d\|\u\|^3}{6Na_N}\sumN
  \Exp\big\{e^{f(\x_i;\bbeta^{*})} B(\x_i)\}
  =\frac{d\|\u\|^3}{6a_N}
  \Exp\big\{e^{f(\x;\bbeta^{*})} B(\x)\big\}\rightarrow0.
\end{align*}
Since $\Delta_{R1}$ and $\Delta_{R2}$ are both positive, Markov's inequality shows that they are both $\op$ so $R=\op$.
For $\Delta_{\ell}$, the mean $\Exp(\Delta_{\ell})=\0$ and the variance satisfies that
\begin{align*}
  \Var(\Delta_{\ell})
  \le\frac{\|\u\|^4}{4a_N^4}\sumN \Exp\{p(\x_i;\btheta^{*})
  \|\ddot{g}(\x_i;\btheta^{*})\|^2\}
  &\le\frac{\|\u\|^4}{4Na_N^2}\sumN \Exp[e^{f(\x_i;\bbeta^{*})}
  \|\ddot{g}(\x_i;\btheta^{*})\|^2]\\
  &=\frac{\|\u\|^4}{4a_N^2}\Exp[e^{f(\x;\bbeta^{*})}
  \|\ddot{g}(\x,\btheta^{*})\|^2]\rightarrow0,
\end{align*}
so $\Delta_{\ell}=\op$.

If we can show that
\begin{align}\label{eq:s3p}
  a_N^{-1}\dot\ell(\btheta^{*})
  \longrightarrow \Nor\big(\0,\ \M_{\f}\big),
\end{align}
in distribution, %
and
\begin{align}\label{eq:s4p}
  a_N^{-2}\sumN\phi(\x_i;\btheta^{*})
  \dot{g}^{\otimes2}(\x_i;\btheta^{*})
 \longrightarrow \M_{\f},
\end{align}
in probability, then from the Basic Corollary in page 2 of \cite{hjort2011asymptotics}, we know that $a_N(\hat\btheta-\btheta^{*})$, the maximizer of $\gamma(\u)$, satisfies that
\begin{equation}\label{eq:s1p}
  a_N(\hat\btheta-\btheta^{*})
  =\M_{\f}^{-1}\times a_N^{-1}\dot\ell(\btheta^{*})+\op.
\end{equation}
Slutsky's theorem together with (\ref{eq:s3p}) and (\ref{eq:s1p}) implies the result in Theorem~\ref{thm:1}. We prove \eqref{eq:s3p} and \eqref{eq:s4p} in the following. 

Note that %
\begin{equation*}
  \dot\ell(\btheta^{*})
  =\sumN\big\{y_i-p(\x_i;\btheta^{*})\big\}\dot{g}(\x_i;\btheta^{*})
\end{equation*}
is a summation of i.i.d. quantities. Since the distribution of $\{y-p(\x;\btheta^{*})\}\dot{g}(\x;\btheta^{*})$ depends on $N$ because  $\alpha^{*}\rightarrow-\infty$ as $N\rightarrow\infty$, we need to use the Lindeberg-Feller central limit theorem for triangular arrays \citep[see, Section $^*$2.8 of][]{Vaart:98}.

We examine the mean and variance of $a_N^{-1}\dot\ell(\btheta^{*})$. For the mean, from the fact that
\begin{align*}
  \Exp[\{y_i-p(\x_i;\btheta^{*})\}\dot{g}(\x_i;\btheta^{*})]
  =\Exp\Big[\Exp\big\{y_i-p(\x_i;\btheta^{*})\bigm|\x_i\big\}
  \dot{g}(\x_i;\btheta^{*})\x_i\Big]=\0,
\end{align*}
we know that $\Exp\{a_N^{-1}\dot\ell(\btheta^{*})\}=\0$.

For the variance,
\begin{align*}
  \Var\{a_N^{-1}\dot\ell(\btheta^{*})\}
  &=a_N^{-2}\sumN\Var[\{y-p(\x;\btheta^{*})\}\dot{g}(\x_i;\btheta^{*})]
  =a_N^{-2}N\Exp\{\phi(\btheta^{*})\dot{g}^{\otimes}(\x;\btheta^{*})\}\\
  &=a_N^{-2}N\Exp\bigg[\frac{e^{\alpha^{*}+f(\x;\bbeta^{*})}}
    {\{1+e^{\alpha^{*}+f(\x;\bbeta^{*})}\}^2}
    \dot{g}^{\otimes}(\x;\btheta^{*})\bigg]
  =\Exp\bigg[\frac{e^{f(\x;\bbeta^{*})}}
    {\{1+e^{\alpha^{*}+f(\x;\bbeta^{*})}\}^2}
    \dot{g}^{\otimes}(\x;\btheta^{*})\bigg].
\end{align*}
We have
\begin{align*}
  \frac{e^{f(\x;\bbeta^{*})}}
    {\{1+e^{\alpha^{*}+f(\x;\bbeta^{*})}\}^2}
    \dot{g}^{\otimes}(\x;\btheta^{*})
  \longrightarrow
  e^{f(\x;\bbeta^{*})}
  \dot{g}^{\otimes}(\x;\btheta^{*}) \quad\text{almost surely,}
\end{align*}
and
\begin{align*}
  \frac{e^{f(\x;\bbeta^{*})}}
    {\{1+e^{\alpha^{*}+f(\x;\bbeta^{*})}\}^2}
  \|\dot{g}(\x;\btheta^{*})\|^2
  \le e^{f(\x;\bbeta^{*})}\|\dot{g}(\x;\btheta^{*})\|^2
  \quad\text{ with }\quad
  \Exp\{e^{f(\x;\bbeta^{*})}\|\dot{g}(\x;\btheta^{*})\|^2\}\le\infty.
\end{align*}
Note that $\dot{g}(\x;\btheta^{*})=\{1,\dot{f}\tp(\x_i;\bbeta^{*})\}\tp$ does not depend on $N$. 
Thus, from the dominated convergence theorem,
\begin{align*}
  \Var\{a_N^{-1}\dot\ell(\btheta^{*})\}\longrightarrow
  \Exp\big\{e^{f(\x;\bbeta^{*})}\dot{g}^{\otimes}(\x;\btheta^{*})\big\}.
\end{align*}
Now we check the Lindeberg-Feller condition. For any $\epsilon>0$,

\begin{align*}
  &\sumN\Exp\Big[\|\{y_i-p(\x_i;\btheta^{*})\}\dot{g}(\x_i;\btheta^{*})\|^2
    I(\|\{y_i-p(\x_i;\btheta^{*})\}\dot{g}(\x_i;\btheta^{*})\|>a_N\epsilon)\Big]\\
  &=N\Exp\Big[\|\{y-p(\x;\btheta^{*})\}\dot{g}(\x;\btheta^{*})\|^2
    I(\|\{y-p(\x;\btheta^{*})\}\dot{g}(\x;\btheta^{*})\|>a_N\epsilon)\Big]\\
  &=N\Exp\big[p(\x;\btheta^{*})\{1-p(\x;\btheta^{*})\}^2\|\dot{g}(\x;\btheta^{*})\|^2
     I(\|\{1-p(\x;\btheta^{*})\}\dot{g}(\x;\btheta^{*})\|>a_N\epsilon)\big]\\
  &\quad+N\Exp\big[\{1-p(\x;\btheta^{*})\}\{p(\x;\btheta^{*})\}^2\|
    \dot{g}(\x;\btheta^{*})\|^2
    I(\|p(\x;\btheta^{*})\dot{g}(\x;\btheta^{*})\|>a_N\epsilon)\big]\\
  &\le N\Exp\big[p(\x;\btheta^{*})\|\dot{g}(\x;\btheta^{*})\|^2
    I(\|\dot{g}(\x;\btheta^{*})\|>a_N\epsilon)\big]\\
    &\quad+N\Exp\big[\{p(\x;\btheta^{*})\}^2\|\dot{g}(\x;\btheta^{*})\|^2
    I(\|p(\x;\btheta^{*})\dot{g}(\x;\btheta^{*})\|>a_N\epsilon)\big]\\
  &\le a_N^2\Exp\{e^{f(\x;\bbeta^{*})}\|\dot{g}(\x;\btheta^{*})\|^2
    I(\|\dot{g}(\x;\btheta^{*})\|>a_N\epsilon)\}\\
    &\quad+a_N^2\Exp\{e^{f(\x;\bbeta^{*})}\|\dot{g}(\x;\btheta^{*})\|^2
    I(\|\dot{g}(\x;\btheta^{*})\|>a_N\epsilon)\}\\
  &=o(a_N^2),
\end{align*}
where the last step is from the dominated convergence theorem. Thus, applying the Lindeberg-Feller central limit theorem \citep[Section $^*$2.8 of][]{Vaart:98}, we finish the proof of \eqref{eq:s3p}.

Now we prove \eqref{eq:s4p}.
This is done by noting that
\begin{align*}
  &a_N^{-2}\sumN\phi(\x_i;\btheta^{*})\dot{g}^{\otimes2}(\x_i;\btheta^{*})\\
  &=\frac{1}{Ne^{\alpha^{*}}}\sumN
     \frac{e^{\alpha^{*}+f(\x_i;\bbeta^{*})}}
     {(1+e^{\alpha^{*}+f(\x_i;\bbeta^{*})})^2}\dot{g}^{\otimes2}(\x_i;\btheta^{*})\\
  &=\frac{1}{N}\sumN\frac{e^{f(\x_i;\bbeta^{*})}}
     {(1+e^{\alpha^{*}+f(\x_i;\bbeta^{*})})^2}\dot{g}^{\otimes2}(\x_i;\btheta^{*})
  =\Exp\{e^{f(\x;\bbeta^{*})}\dot{g}^{\otimes2}(\x;\btheta^{*})\}+\op,
\end{align*}
where the last step is from Lemma 28 of \cite{wang2019more}.

\section{Proof of Theorem~\ref{thm:2}}
  First, note that $\V_w$ can be written as 
\begin{equation*}
    \V_w=\Exp(e^{f(\x;\bbeta^{*})})\M_{\f}^{-1}\M_w\M_{\f}^{-1},
\end{equation*}
where
\begin{equation*}
  \M_w
  =\Exp\bigg[\bigg\{1+\frac{ce^{f(\x;\bbeta^{*})}}{\varphi(\x)}
    \bigg\}e^{f(\x;\bbeta^{*})}\dot{g}^{\otimes2}(\x;\btheta^{*})\bigg].
\end{equation*}

We then point out that under assumptions~\ref{asmp:1} and \ref{asmp:2} the condition $\Exp[\{\varphi(\x)+\varphi^{-1}(\x)\}B^2(\x)]<\infty$ implies the following:
\begin{align}
  &\Exp\big[\big\{\varphi(\x)e^{f(\x;\bbeta^{*})}\big\}
    \|\dot{g}(\x;\btheta^{*})\|^2\big]<\infty,\label{eq:38}\\
  &\Exp\big[\big\{1+\varphi^{-1}(\x)\big\}
    e^{2f(\x;\bbeta^{*})}\|\dot{g}(\x;\btheta^{*})\|^4\big]<\infty,\label{eq:39}\\
  &\Exp\big\{\varphi^{-1}(\x)e^{2f(\x;\bbeta^{*})}
    \|\ddot{g}(\x;\btheta^{*})\|^2\big\}<\infty.\label{eq:40}
\end{align}

The estimator $\htheta_w$ is the maximizer of
\begin{align*}
  \ell_w(\btheta)=\sumN\frac{\delta_i}{\pi(\x_i,y_i)}
  \big[y_ig(\x_i;\btheta)-\log\{1+e^{g(\x_i;\btheta)}\}\big],
\end{align*}
so $a_N(\htheta_w-\btheta^{*})$ is the maximizer of $\gamma_w(\u)=\ell_w(\btheta^{*}+a_N^{-1}\u)-\ell_w(\btheta^{*})$. By Taylor's expansion,
\begin{align*}
  \gamma_w(\u)
  &=\frac{1}{a_N}\u\tp\dot\ell_w(\btheta^{*})
    +\frac{1}{2a_N^2}\sumN\frac{\delta_i}{\pi(\x_i,y_i)}
    \phi(\x_i;\btheta^{*})(\z_i\tp\u)^2
    +\Delta_{\ell_w}+R_w,
\end{align*}
where 
\begin{align*}
  \dot\ell_w(\btheta)
  &=\frac{\partial\ell_w(\btheta)}{\partial\btheta}
    =\sumN\frac{\delta_i}{\pi(\x_i,y_i)}\big\{y_i-p(\x_i;\btheta)\big\}
    \dot{g}(\x_i,\btheta),\\
  \Delta_{\ell_w}
  &=\frac{1}{2a_N^2}\sumN\frac{\delta_i}{\pi(\x_i,y_i)}
  \big\{y_i-p(\x_i;\btheta^{*})\big\}\u\tp\ddot{g}(\x_i,\btheta^{*})\u,
\end{align*}
and $R_w$ is the remainder. 
Similarly to the proof of Theorem~\ref{thm:1}, we only need to show that
\begin{align}\label{eq:s9}
  a_N^{-1}\dot\ell_w(\btheta^{*}) \longrightarrow
  \Nor(\0,\ \M_w),
\end{align}
in distribution,
\begin{align}\label{eq:s10}
  a_N^{-2}\sumN\frac{\delta_i}{\pi(\x_i,y_i)}
  \phi(\x_i;\btheta^{*})\dot{g}^{\otimes2}(\x_i;\btheta^{*})
  \longrightarrow \M_{\f},
\end{align}
in probability for any $\u$, and $\Delta_{\ell_w}=\op$ and $R_w=\op$. 

We prove~\eqref{eq:s9} first. Let $\eta_i=\frac{\delta_i}{\pi(\x_i,y_i)}\{y_i-p(\x_i;\btheta^{*})\}\dot{g}(\x_i;\btheta^{*})$, we know that $\eta_i$, $i=1, ..., N$, are i.i.d., with the underlying distribution of $\eta_i$ being dependent on $N$. From direct calculation, we have that $\Exp(\eta_i\mid\x_i)=\0$ and
\begin{align*}
  \Var(\eta_i\mid\x_i)
  &=\Exp\bigg[\frac{\{y_i-p(\x_i;\btheta^{*})\}^2}
    {\{y_i+(1-y_i)\rho\varphi(\x_i)\}}
    \biggm|\x_i\bigg]\dot{g}^{\otimes2}(\x_i;\btheta^{*})\\
  &=p(\x_i;\btheta^{*})\{1-p(\x_i;\btheta^{*})\}^2
    \dot{g}^{\otimes2}(\x_i;\btheta^{*})
  +\{1-p(\x_i;\btheta^{*})\}
    \frac{p^2(\x_i;\btheta^{*})}{\rho\varphi(\x_i)}
    \dot{g}^{\otimes2}(\x_i;\btheta^{*})\\
  &\le
    e^{\alpha^{*}}e^{f(\x_i;\bbeta^{*})}\dot{g}^{\otimes2}(\x_i;\btheta^{*})
    +\rho^{-1}e^{2\alpha^{*}}
    \frac{e^{2f(\x_i;\bbeta^{*})}}{\varphi(\x_i)}\dot{g}^{\otimes2}(\x_i;\btheta^{*}).
\end{align*}
Thus, by the dominated convergence theorem, we obtain that 
\begin{align*}
  \Var(\eta_i)
  &=\Exp\{\Var(\eta_i\mid\x_i)\}
  =e^{\alpha^{*}}\Exp\bigg[
    \bigg\{1+\frac{ce^{f(\x_i;\bbeta^{*})}}{\varphi(\x_i)}
    \bigg\}e^{f(\x_i;\bbeta^{*})}\dot{g}^{\otimes2}(\x_i;\btheta^{*})\bigg]\{1+o(1)\}.
\end{align*}

Now we check the Lindeberg-Feller condition \citep[Section $^*$2.8 of][]{Vaart:98}. Denote 
$\delta=y+(1-y)I\{u\le \rho\varphi(\x)\}$ where $u\sim\mathbb{U}(0,1)$.
For any $\epsilon>0$,
\begin{align*}
  &\sumN\Exp\big\{\|\eta_i\|^2I(\|\eta_i\|>a_N\epsilon)\big\}\\
  &=N\Exp\big[\|\pi^{-1}(\x,y)\delta\{y-p(\x;\btheta^{*})\}
    \dot{g}(\x;\btheta^{*})\|^2
    I(\|\pi^{-1}(\x,y)\delta\{y-p(\x;\btheta^{*})\}
    \dot{g}(\x;\btheta^{*})\|>a_N\epsilon)\big]\\
  &=\rho N\Exp\Big[\varphi(\x)\|\pi^{-1}(\x,y)
    \{y-p(\x;\btheta^{*})\}\dot{g}(\x;\btheta^{*})\|^2
    I(\|\pi^{-1}(\x,y)\{y-p(\x;\btheta^{*})\}\dot{g}(\x;\btheta^{*})\|
    >a_N\epsilon)\Big]\\
  &\quad+N\Exp\Big[\{1-\rho\varphi(\x)\}\|\pi^{-1}(\x,y)
    y\{y-p(\x;\btheta^{*})\}\dot{g}(\x;\btheta^{*})\|^2\\
  &\hspace{4cm}\times
    I(\|\pi^{-1}(\x,y)y\{y-p(\x;\btheta^{*})\}\dot{g}(\x;\btheta^{*})\|
    >a_N\epsilon)\Big]\\
  &=\rho N\Exp\Big[\varphi(\x)p(\x;\btheta^{*})
    \|\{1-p(\x;\btheta^{*})\}\dot{g}(\x;\btheta^{*})\|^2
    I(\|\{1-p(\x;\btheta^{*})\}\dot{g}(\x;\btheta^{*})\|
    >a_N\epsilon)\Big]\\
  &+N\Exp\Big[\{1-p(\x;\btheta^{*})\}\rho^{-1}\varphi^{-1}(\x)
    \|p(\x;\btheta^{*})\dot{g}(\x;\btheta^{*})\|^2\\
  &\hspace{4cm}\times
    I(\|\rho^{-1}\varphi^{-1}(\x)\{y-p(\x;\btheta^{*})\}\dot{g}(\x;\btheta^{*})\|
    >a_N\epsilon)\Big]\\
  &\quad+N\Exp\Big[\{1-\rho\varphi(\x)\}p(\x;\btheta^{*})
    \|\{1-p(\x;\btheta^{*})\}\dot{g}(\x;\btheta^{*})\|^2
    I(\|\{1-p(\x;\btheta^{*})\}\dot{g}(\x;\btheta^{*})\|
    >a_N\epsilon)\Big]\\
  &\le\rho Ne^{\alpha^{*}}\Exp\Big[\varphi(\x)e^{f(\x;\bbeta^{*})}
    \|\dot{g}(\x;\btheta^{*})\|^2
    I(\|\dot{g}(\x;\btheta^{*})\|
    >a_N\epsilon)\Big]\\
  &\quad+Ne^{2\alpha^{*}}\rho^{-1}\Exp\Big[
    \varphi^{-1}(\x)\|e^{f(\x;\bbeta^{*})}\dot{g}(\x;\btheta^{*})\|^2
    I(\|\rho^{-1}\varphi^{-1}(\x)\{y-p(\x;\btheta^{*})\}\dot{g}(\x;\btheta^{*})\|
    >a_N\epsilon)\Big]\\
  &\quad+Ne^{\alpha^{*}}\Exp\Big[e^{f(\x;\bbeta^{*})}
    \|\dot{g}(\x;\btheta^{*})\|^2
    I(\|\dot{g}(\x;\btheta^{*})\|
    >a_N\epsilon)\Big]\\
  &=o(Ne^{\alpha^{*}}),%
\end{align*}
where the last step is due to Assumptions~\ref{asmp:1},  (\ref{eq:38})-(\ref{eq:40}), the dominated convergence theorem, and the facts that $a_N\rightarrow\infty$ and $\lim_{N\rightarrow\infty}e^\alpha/\rho=c<\infty$. Thus, applying the Lindeberg-Feller central limit theorem \citep[Section $^*$2.8 of][]{Vaart:98} finishes the proof of \eqref{eq:s9}.

Now we prove~\eqref{eq:s10}. Let
\begin{align*}
  H_{w}
  &\equiv a_N^{-2}\sumN\frac{\delta_i}{\pi(\x_i,y_i)}\phi(\x_i;\btheta^{*})
  \dot{g}^{\otimes2}(\x_i;\btheta^{*})\\
  &=\onen\sumN\frac{\delta_i}{\pi(\x_i,y_i)}
    \frac{e^{f(\x_i;\bbeta^{*})}}{(1+e^{\alpha^{*}+f(\x_i;\bbeta^{*})})^2}
    \dot{g}^{\otimes2}(\x_i;\btheta^{*})\\
  &=\onen\sumN\frac{y_i+(1-y_i)I\{u_i\le \rho\varphi(\x_i)\}}
    {y_i+(1-y_i)\rho\varphi(\x_i)}
    \frac{e^{f(\x_i;\bbeta^{*})}}{(1+e^{\alpha^{*}+f(\x_i;\bbeta^{*})})^2}
    \dot{g}^{\otimes2}(\x_i;\btheta^{*}).
\end{align*}
We notice that
\begin{align}\label{eq:s20}
  \Exp(H_{w})=
  &\Exp\bigg\{\frac{e^{f(\x;\bbeta^{*})}}
     {(1+e^{\alpha^{*}+f(\x;\bbeta^{*})})^2}\dot{g}^{\otimes2}(\x;\btheta^{*})\bigg\}
  =\Exp\big\{e^{f(\x;\bbeta^{*})}\dot{g}^{\otimes2}(\x;\btheta^{*})\big\}+o(1),
\end{align}
where the last step is by the dominated convergence theorem. In addition, the variance of each component of $H_{w}$ is bounded by 
\begin{align}
  &\oneN\Exp\bigg\{
    \frac{\delta}{\pi^2(\x,y)}
    e^{2f(\x;\bbeta^{*})}\|\dot{g}(\x;\btheta^{*})\|^4\bigg\}\notag\\
  &\le\oneN\Exp\bigg\{
    \frac{1}{\pi(\x,y)}
    e^{2f(\x;\bbeta^{*})}\|\dot{g}(\x;\btheta^{*})\|^4\bigg\}\notag\\
  &\le\frac{1}{N\rho}\Exp\bigg[
    \Big\{\rho+\frac{1}{\varphi(\x)}\Big\}
    e^{2f(\x;\bbeta^{*})}\|\dot{g}(\x;\btheta^{*})\|^4\bigg]
    =o(1).\label{eq:s21}
\end{align}
where the last step is because of  (\ref{eq:39}) and the fact that $Ne^{\alpha^{*}}\rightarrow\infty$ and $e^{\alpha^{*}}/\rho\rightarrow c<\infty$ imply that $N\rho\rightarrow\infty$. 
From (\ref{eq:s20}) and (\ref{eq:s21}), Chebyshev's inequality implies that $H_{w}\rightarrow \Exp\big\{e^{f(\x;\bbeta^{*})}\dot{g}^{\otimes2}(\x;\btheta^{*})\big\}$ in probability. 

In the following we finish the proof by showing that $\Delta_{\ell_w}=\op$ and $R_w=\op$.
For $\Delta_{\ell_w}$, the mean $\Exp(\Delta_{\ell_w})=\0$ and the variance satisfies that
\begin{align*}
  \Var(\Delta_{\ell_w})
  &\le\frac{\|\u\|^4N}{4a_N^4}\Exp\bigg[
    \frac{\{y-p(\x;\btheta^{*})\}^2\|\ddot{g}(\x;\btheta^{*})\|^2}
    {y+(1-y)\rho\varphi(\x)}\bigg]\\
  &\le\frac{\|\u\|^4N}{4a_N^4}\Exp\bigg[
    \bigg\{1+\frac{p(\x;\btheta^{*})}{\rho\varphi(\x)}\bigg\}
    p(\x;\btheta^{*})\|\ddot{g}(\x;\btheta^{*})\|^2\bigg]\\
  &\le\frac{\|\u\|^4}{4a_N^2}\Exp\bigg[
    \bigg\{1+\frac{e^{\alpha^{*}}e^{f(\x;\bbeta^{*})}}{\rho\varphi(\x)}\bigg\}
    e^{f(\x;\bbeta^{*})}\|\ddot{g}(\x;\btheta^{*})\|^2\bigg]
  =\op,
\end{align*}
so $\Delta_{\ell_w}=\op$.
For the remainder term $R_w$. By direct calculation,
\begin{align*}
  R_w
  &=\frac{1}{6a_N^3}\sumN\frac{\delta_i}{\pi(\x_i,y_i)}
    \phi(\x_i;\btheta^{*}+a_N^{-1}\Acute\u)
    \big\{1-2p(\x_i;\btheta^{*}+a_N^{-1}\Acute\u)\big\}
    \{\u\tp\dot{g}(\x_i;\btheta^{*}+a_N^{-1}\Acute\u)\}^3\\
  &+\frac{1}{6a_N^3}\sumN\frac{\delta_i}{\pi(\x_i,y_i)}
    \phi(\x_i;\btheta^{*}+a_N^{-1}\Acute\u)
    \{\u\tp\dot{g}(\x_i;\btheta^{*}+a_N^{-1}\Acute\u)\}
    \{\u\tp\ddot{g}(\x_i;\btheta^{*}+a_N^{-1}\Acute\u)\u\}\\
  &+\frac{1}{6a_N^3}\sumN\frac{\delta_i}{\pi(\x_i,y_i)}
    \bigg[\big\{y_i-p(\x_i;\btheta^{*}+a_N^{-1}\Acute\u)\big\}
    \sum_{j_1j_2}^du_{j_1}u_{j_2}
    \dddot{g}_{j_1j_2}(\x_i;\btheta^{*}+a_N^{-1}\Acute\u)\u\bigg],
\end{align*}
where $\Acute\u$ lies between $\0$ and $\u$.
We see that
\begin{align*}
  |R_w|
  &\le\frac{\|\u\|^3}{6a_N^3}\sumN\frac{\delta_i}{\pi(\x_i,y_i)}
    p(\x_i;\btheta^{*}+a_N^{-1}\Acute\u) C(\x_i,\btheta^{*}+a_N^{-1}\Acute\u)
  +\frac{d\|\u\|^3}{6a_N^3}\sumN\frac{\delta_i}{\pi(\x_i,y_i)}y_i B(\x_i)\\
  &\le\frac{\|\u\|^3e^{a_N^{-1}\|\u\|}}{6Na_N}
    \sumN\frac{\delta_i}{\pi(\x_i,y_i)}
    \exp\{f(\x_i;\bbeta+a_N^{-1}\Acute\u_{(-1)})\}C(\x_i,\btheta^{*}+a_N^{-1}\Acute\u)\\
  &\quad+\frac{d\|\u\|^3}{6a_N^3}\sumN\frac{\delta_i}{\pi(\x_i,y_i)} y_i B(\x_i)\\
  &\le\frac{d^3\|\u\|^3e^{a_N^{-1}\|\u\|}}{6Na_N}
    \sumN\frac{\delta_i}{\pi(\x_i,y_i)} B(\x_i)
    +\frac{d\|\u\|^3}{6a_N^3}\sumN\frac{\delta_i}{\pi(\x_i,y_i)} y_i B(\x_i)\\
  &\equiv\Delta_{R_w1}+\Delta_{R_w2},
\end{align*}
where $C(\x,\btheta)$ is defined in (\ref{eq:12})
and $\Acute\u_{(-1)}$ is $\Acute\u$ with the first element removed. From Assumption~\ref{asmp:1}, $\Exp(\Delta_{R_w1})\rightarrow0$ and
\begin{align*}
  \Exp(\Delta_{R_w2})
  \le\frac{d\|\u\|^3}{6a_N}
  \Exp\big[e^{f(\x;\bbeta^{*})} B(\x)\big]\rightarrow0.
\end{align*}
Since $\Delta_{R_w1}$ and $\Delta_{R_w2}$ are both positive, Markov's inequality shows that they are both $\op$ so $R_w=\op$.

\section{Proof of Theorem~\ref{thm:3}}
Since only $\V_{\mathrm{sub}}$ in $\V_w$ is affect by subsampling, the probability that minimizes $\tr(\V_{\mathrm{sub}})$ also minimizes $\tr(\V_w)$. Let $\Lambda_{\mathrm{sub}}=\Exp\{\varphi^{-1}(\x)e^{2f(\x;\bbeta^{*})}\dot{g}^{\otimes2}(\x;\btheta^{*})\}$.
 We notice that
\begin{align*}
  \V_{\mathrm{sub}}
  &=c\Exp(e^{f(\x;\bbeta^{*})})\M_{\f}^{-1}\Lambda_{\mathrm{sub}}\M_{\f}^{-1}\\
  &=c\Exp(e^{f(\x;\bbeta^{*})})e^{-2\alpha^{*}}\{1+\op\}
    \M_{\f}^{-1}\Exp\{\varphi^{-1}(\x)p^2(\x;\btheta^{*})
    \dot{g}^{\otimes2}(\x;\btheta^{*})\}\M_{\f}^{-1},
\end{align*}
where the term $c\Exp(e^{f(\x;\bbeta^{*})})e^{-2\alpha^{*}}$ does not depend on $\varphi(\x)$. Thus we can focus on minimizing the term $\M_{\f}^{-1}\Exp\{\varphi^{-1}(\x)p^2(\x;\btheta^{*})
    \dot{g}^{\otimes2}(\x;\btheta^{*})\}\M_{\f}^{-1}$.

Note that 
\begin{equation*}
  \tr[p^2(\x;\btheta^{*})\M_{\f}^{-1}
    \dot{g}^{\otimes2}(\x;\btheta^{*})\M_{\f}^{-1}]
  =\|p(\x;\btheta)\M_{\f}^{-1}\dot{g}(\x;\btheta)\|^2
  =t^2(\x;\btheta^{*}).
\end{equation*}
Thus the problem is to find $\varphi(\x)$ that minimizes
\begin{equation*}
  \Exp\bigg\{\frac{t^2(\x;\btheta^{*})}{\varphi(\x)}\bigg\},
  \quad\text{subject to}\quad
  0<\varphi(\x)\le \rho^{-1} \text{ and } \Exp\{\varphi(\x)\}=1.
\end{equation*}
To facilitate the presentation, denote
\begin{equation*}
  \zeta=\frac{1}{\Exp[\min\{t(\x;\btheta^{*}), T\}]},
\end{equation*}
which is non-random. We notice that
\begin{align*}
  &\Exp\bigg\{\frac{\zeta^2t^2(\x;\btheta^{*})}{\varphi(\x)}\bigg\}\\
  &=\Exp\bigg(\frac{\zeta^2[\min\{t(\x;\btheta^{*}), T\}]^2}{\varphi(\x)}\bigg)
    +\Exp\bigg[\frac{\zeta^2\{t^2(\x;\btheta^{*})-T^2\}}{\varphi(\x)}
    I\{t(\x;\btheta^{*})>T\}\bigg]\\
  &=\Exp\Bigg(\bigg[\sqrt{\varphi(\x)}
    -\frac{\zeta\min\{t(\x;\btheta^{*}), T\}}{\sqrt{\varphi(\x)}}\bigg]^2\Bigg)
  -\Exp\{\varphi(\x)\}+2\Exp\big[\zeta\min\{t(\x;\btheta^{*}), T\}\big]\\
  &\quad+\Exp\bigg[\frac{\zeta^2\{t^2(\x;\btheta^{*})-T^2\}}
    {\varphi(\x)}I\{t(\x;\btheta^{*})>T\}\bigg]\\
  &=\Exp\Bigg(\bigg[\sqrt{\varphi(\x)}
    -\frac{\zeta\min\{t(\x;\btheta^{*}), T\}}{\sqrt{\varphi(\x)}}\bigg]^2\Bigg)
    +\Exp\bigg[\frac{\zeta^2\{t^2(\x;\btheta^{*})-T^2\}}{\varphi(\x)}
    I\{t(\x;\btheta^{*})>T\}\bigg] + 1\\
  &\equiv E_1+E_2 +1,
\end{align*}
where $I(\cdot)$ is the indicator function. 
Note that $E_i$'s are non-negative and $E_1=0$ if and only if
\begin{equation*}
  \varphi(\x)=\zeta\min\{t(\x;\btheta^{*}), T\}
  =\frac{\min\{t(\x;\btheta^{*}), T\}}{\Exp[\min\{t(\x;\btheta^{*}), T\}]},
\end{equation*}
which is attainable because
\begin{align}\label{eq:s12}
  \rho\min\{t(\x;\btheta^{*}), T\}\le\Exp[\min\{t(\x;\btheta^{*}), T\}],
  \quad\text{almost surely}.
\end{align}
If $t(\x;\btheta^{*})\le T$ almost surely, then $E_2=0$ and the proof finishes. If $\Pr\{t(\x;\btheta^{*})>T\}>0$ then 
\begin{equation*}
  \frac{\rho T}{\Exp[\min\{t(\x;\btheta^{*}), T\}]}=1,
\end{equation*}
because if this is not true then we can find a larger $T$ that satisfies (\ref{eq:s12}). This means that if $t(\x;\btheta^{*})>T$, then
\begin{equation*}
  \varphi(\x)=\frac{\min\{t(\x;\btheta^{*}), T\}}{\Exp[\min\{t(\x;\btheta^{*}), T\}]}=\rho^{-1},
\end{equation*}
which minimizes $E_2$ as well since $\varphi(\x)$ is in the denominator. This finishes the proof.

\section{Derivation of corrected model (\ref{eq:2})}
Note that $\pi(\x,1)=1$ and $\pi(\x,0)=\pi(\x)$. 
By direct calculation, we have
\begin{align*}
  &\Pr(y=1\mid\x,\delta=1)\\
  &=\frac{\Pr(y=1,\delta=1\mid\x)}{\Pr(\delta=1\mid\x)}\\
  &=\frac{\Pr(y=1,\delta=1\mid\x)}
  {\Pr(y=1,\delta=1\mid\x)+\Pr(y=0,\delta=1\mid\x)}\\
  &=\frac{\Pr(y=1\mid\x)\Pr(\delta=1\mid\x,y=1)}
    {\Pr(y=1\mid\x)\Pr(\delta=1\mid\x,y=1)
    +\Pr(y=0\mid\x)\Pr(\delta=1\mid\x,y=0)}\\
  &=\frac{p(\x;\btheta^{*})\pi(\x,1)}{p(\x;\btheta^{*})\pi(\x,1)
    +\{1-p(\x;\btheta^{*})\}\pi(\x,0)}\\
  &=\frac{e^{g(\x;\btheta^{*})}}{e^{g(\x;\btheta^{*})}+\pi(\x)}\\
  &=\frac{e^{g(\x;\btheta^{*})+l}}{1+e^{g(\x;\btheta^{*})+l}}.
\end{align*}

\section{Proof of Theorem~\ref{thm:4}}
The estimator $\htheta_{\lik}$ is the maximizer of $\ell_{\lik}(\btheta)$, so $\u_N=a_N(\htheta_{\lik}-\btheta)$ is the maximizer of
\begin{equation*}
  \gamma_{\lik}(\u)=\ell_{\lik}(\btheta+a_N^{-1}\u)-\ell_{\lik}(\btheta).
\end{equation*}
By Taylor's expansion,
\begin{equation*}
  \gamma_{\lik}(\u) =a_N^{-1}\u\tp\dot\ell_{\lik}(\btheta^{*})
    +0.5a_N^{-2}\sumN \phi_{\pi}(\x_i;\btheta^{*})
    \{\u\tp\dot{g}(\x_i;\btheta^{*})\}^2
    +\Delta_{\ell_{\lik}}+R_{\lik},
\end{equation*}
where $\phi_{\pi}(\x;\btheta)=p_{\pi}(\x;\btheta)\{1-p_{\pi}(\x;\btheta)\}$,
\begin{equation}
\label{eq:s8}
  p_{\pi}(\x;\btheta)
  =\frac{e^{g(\x;\btheta)+l}}{1+e^{g(\x;\btheta)+l}}
  \quad\text{with}\quad l=-\log\{\rho\varphi(\x)\},
\end{equation}
\begin{align*}
  \dot\ell_{\lik}(\btheta)
  &=\frac{\partial\ell_{\lik}(\btheta)}{\partial\btheta}
  =\sumN\delta_i\big\{y_i-p_{\pi}(\x_i;\btheta)\big\}\dot{g}(\x_i;\btheta),\\
  \Delta_{\ell_{\lik}}
  &=\frac{1}{2a_N^2}\sumN\delta_i\big\{y_i-p_{\pi}(\x_i;\btheta^{*})\big\}
  \u\tp\ddot{g}(\x_i;\btheta^{*})\u,
\end{align*}
and $R_{\lik}$ is the remainder term. By direct calculation,
\begin{align*}
  R_{\lik}
  &=\frac{1}{6a_N^3}\sumN\delta_i
    \phi_{\pi}(\x_i;\btheta^{*}+a_N^{-1}\Acute\u)
    \big\{1-2p_{\pi}(\x_i;\btheta^{*}+a_N^{-1}\Acute\u)\big\}
    \{\u\tp\dot{g}(\x_i;\btheta^{*}+a_N^{-1}\Acute\u)\}^3\\
  &+\frac{1}{6a_N^3}\sumN\delta_i
    \phi_{\pi}(\x_i;\btheta^{*}+a_N^{-1}\Acute\u)
    \{\u\tp\dot{g}(\x_i;\btheta^{*}+a_N^{-1}\Acute\u)\}
    \{\u\tp\ddot{g}(\x_i;\btheta^{*}+a_N^{-1}\Acute\u)\u\}\\
  &+\frac{1}{6a_N^3}\sumN\delta_i
    \bigg[\big\{y_i-p_{\pi}(\x_i;\btheta^{*}+a_N^{-1}\Acute\u)\big\}
    \sum_{j_1,j_2=1}^du_{j_1}u_{j_2}
    \dddot{g}_{j_1j_2}(\x_i;\btheta^{*}+a_N^{-1}\Acute\u)\u\bigg],
\end{align*}
where $\Acute\u$ lies between $\0$ and $\u$.
We see that %
\begin{align*}
  |R_{\lik}|
  &\le\frac{\|\u\|^3}{6a_N^3}\sumN\delta_i
    p_{\pi}(\x_i;\btheta^{*}+a_N^{-1}\Acute\u) C(\x_i,\btheta^{*}+a_N^{-1}\Acute\u)
  +\frac{d\|\u\|^3}{6a_N^3}\sumN\delta_iy_i B(\x_i)\\ %
  &\le\frac{\|\u\|^3e^{a_N^{-1}\|\u\|}}{6Na_N}\sumN\delta_i
    \exp[f(\x_i;\bbeta+a_N^{-1}\Acute\u_{(-1)})-\log\{\rho\varphi(\x_i)\}]
    C(\x_i,\btheta^{*}+a_N^{-1}\Acute\u)\\
  &\qquad+\frac{d\|\u\|^3}{6a_N^3}\sumN\delta_iy_i B(\x_i)\\
  &=\frac{\|\u\|^3e^{a_N^{-1}\|\u\|}}{6Na_N\rho}\sumN\delta_i
    \varphi^{-1}(\x_i)\exp[f(\x_i;\bbeta+a_N^{-1}\Acute\u_{(-1)})]
    C(\x_i,\btheta^{*}+a_N^{-1}\Acute\u)\\
  &\qquad+\frac{d\|\u\|^3}{6a_N^3}\sumN\delta_iy_i B(\x_i)\\
  &\le\frac{d^3\|\u\|^3e^{a_N^{-1}\|\u\|}}{6Na_N\rho}
    \sumN\delta_i \varphi^{-1}(\x_i)B(\x_i)
    +\frac{d\|\u\|^3}{6a_N^3}\sumN y_i B(\x_i)\\ %
  &\equiv\Delta_{R_{\lik}1}+\Delta_{R2}=\Delta_{R_{\lik}1}+\op,
\end{align*}
where $\Acute\u_{(-1)}$ is $\Acute\u$ with the first element removed,
$C(\x,\btheta)$ and $\Delta_{R2}=\op$ are defined in~(\ref{eq:12}).
For $\Delta_{R_{\lik}1}$, we see that %
\begin{align*}
  \Exp\Big\{(N\rho)^{-1}\sumN\delta_i \varphi^{-1}(\x_i)B(\x_i)\Big\}
  &=\rho^{-1}\Exp\Big([\{p(\x_i;\btheta^{*})\varphi^{-1}(\x)
  +\rho\{1-p(\x_i;\btheta^{*})\}]B(\x)\Big)\\
  &\le e^{\alpha^{*}}\rho^{-1}\Exp\{e^{f(\x;\bbeta^{*})}\varphi^{-1}(\x)B(\x)\}
  +\Exp\{B(\x)\}.
\end{align*}
Therefore, $\Exp(\Delta_{R_{\lik}1})\rightarrow0$. Since $\Delta_{R_{\lik}1}$ is positive, Markov's inequality shows that $\Delta_{R_{\lik}1}=\op$ and thus $R_{\lik}=\op$.

For $\Delta_{\ell_{\lik}}$, the mean $\Exp(\Delta_{\ell_{\lik}})=\0$ and the variance satisfies that
\begin{align*}
  \Var(\Delta_{\ell_{\lik}})
  &\le\frac{\|\u\|^4}{4a_N^4}\sumN\Exp\{\delta_ip_{\pi}(\x_i;\btheta^{*})
  \|\ddot{g}(\x_i;\btheta^{*})\|^2\}
  \le\frac{\|\u\|^4}{4a_N^2}\Exp[e^{f(\x;\bbeta^{*})}
  \|\ddot{g}(\x,\btheta^{*})\|^2]\rightarrow0,
\end{align*}
where the last step is because Assumption~\ref{asmp:1} implies that 
  $e^{f(\x;\bbeta^{*})}\|\ddot{g}(\x,\btheta^{*})\|^2$ is integrable, 
so $\Delta_{\ell_{\lik}}=\op$. 

If we can show that
\begin{align}\label{eq:13}
  a_N^{-1}\dot\ell_{\lik}(\btheta^{*})
  \longrightarrow \Nor\big(\0,\ \bLambda_{\lik}\big),
\end{align}
in distribution, %
and
\begin{align}\label{eq:14}
  a_N^{-2}\sumN\delta_i\phi_{\pi}(\x_i;\btheta^{*})
  \dot{g}^{\otimes2}(\x_i;\btheta^{*})
 \longrightarrow \bLambda_{\lik},
\end{align}
in probability, then from the Basic Corollary in page 2 of \cite{hjort2011asymptotics}, we know that $a_N(\htheta_{\lik}-\btheta^{*})$, the maximizer of $\gamma_{\lik}(\u)$, satisfies that
\begin{equation}\label{eq:41}
  a_N(\htheta_{\lik}-\btheta^{*})
  =\bLambda_{\lik}^{-1}\times a_N^{-1}\dot\ell_{\lik}(\btheta^{*})+\op.
\end{equation}
Slutsky's theorem together with (\ref{eq:13}) and (\ref{eq:41}) implies the result in Theorem~\ref{thm:1}. We prove \eqref{eq:13} and \eqref{eq:14} in the following. 

Note that $\dot\ell_{\lik}(\btheta^{*})$
is a summation of i.i.d. quantities, $\delta_i\big\{y_i-p_{\pi}(\x_i;\btheta^{*})\big\}\dot{g}(\x_i;\btheta^{*})$'s,  whose distribution depends on $N$.  %
Note that
\begin{align}
  &\Exp[\delta\{y-p_{\pi}(\x;\btheta^{*})\}\dot{g}(\x;\btheta^{*})]\notag\\
  &=\Exp[\pi(\x,y)\{y-p_{\pi}(\x;\btheta^{*})\}\dot{g}(\x;\btheta^{*})]\notag\\
  &=\Exp\Big(\big[p(\x;\btheta^{*})\{1-p_{\pi}(\x;\btheta^{*})\}
    -\{1-p(\x;\btheta^{*})\}\rho\varphi(\x)p_{\pi}(\x;\btheta^{*})\big]
    \dot{g}(\x;\btheta^{*})\Big)
  =\0,\label{eq:43}
\end{align}
where the last step is by a direct calculation from (\ref{eq:s8}) to obtain
the following equality
\begin{align}\label{eq:20}
  p(\x;\btheta^{*})\{1-p_{\pi}(\x;\btheta^{*})\}
    =\{1-p(\x;\btheta^{*})\}\rho\varphi(\x)p_{\pi}(\x;\btheta^{*}).
\end{align}
Thus we know that $\Exp\{a_N^{-1}\dot\ell_{\lik}(\btheta^{*})\}=\0$.
The variance of $a_N^{-1}\dot\ell_{\lik}(\btheta^{*})$ satisfies that 
\begin{align}
  &\Var\{a_N^{-1}\dot\ell_{\lik}(\btheta^{*})\}\notag\\
  &=a_N^{-2}N\Exp\big[\delta\{y-p_{\pi}(\x;\btheta^{*})\}^2
    \dot{g}^{\otimes2}(\x;\btheta^{*})\big]\notag\\
  &=e^{-\alpha^{*}}\Exp\big[\{y+(1-y)\rho\varphi(\x)\}
    \{y-p_{\pi}(\x;\btheta^{*})\}^2
    \dot{g}^{\otimes2}(\x;\btheta^{*})\big]\notag\\
  &=e^{-\alpha^{*}}\Exp\Big(\big[
    p(\x;\btheta^{*})\{1-p_{\pi}(\x;\btheta^{*})\}^2
    +\{1-p(\x;\btheta^{*})\}\rho\varphi(\x)p_{\pi}^2(\x;\btheta^{*})\big]
    \dot{g}^{\otimes2}(\x;\btheta^{*})\Big)\label{eq:17}\\
  &=e^{-\alpha^{*}}\Exp\Big(\big[
    p(\x;\btheta^{*})\{1-p_{\pi}(\x;\btheta^{*})\}^2
    +p(\x;\btheta^{*})\{1-p_{\pi}(\x;\btheta^{*})\}
    p_{\pi}(\x;\btheta^{*})\big]
    \dot{g}^{\otimes2}(\x;\btheta^{*})\Big)\\
  &=e^{-\alpha^{*}}\Exp\Big(\big[
    \{1-p_{\pi}(\x;\btheta^{*})\}^2
    +\{1-p_{\pi}(\x;\btheta^{*})\}p_{\pi}(\x;\btheta^{*})\big]p(\x;\btheta^{*})
    \dot{g}^{\otimes2}(\x;\btheta^{*})\Big)\\
  &=e^{-\alpha^{*}}\Exp\Big(\big[
    \{1-p_{\pi}(\x;\btheta^{*})\}^2
    +\phi_{\pi}(\x;\btheta^{*})\big]p(\x;\btheta^{*})
    \dot{g}^{\otimes2}(\x;\btheta^{*})\Big)\\
  &=\Exp\Big(\big[\{1-p_{\pi}(\x;\btheta^{*})\}^2
    +\phi_{\pi}(\x;\btheta^{*})\big]\{1-p(\x;\btheta^{*})\}
    e^{f(\x;\bbeta^{*})}\dot{g}^{\otimes2}(\x;\btheta^{*})\Big)\label{eq:15}
\end{align}
where the forth equality uses (\ref{eq:20}).
Note that elements of the term in the expectation of \eqref{eq:15} are all bounded by integrable random variable $e^{f(\x;\bbeta^{*})}\|\dot{g}(\x;\btheta^{*})\|^2$. 
Thus, from the dominated convergence theorem and the fact that
\begin{align*}
  &\big[\{1-p_{\pi}(\x;\btheta^{*})\}^2
  +\phi_{\pi}(\x;\btheta^{*})\big]\{1-p(\x;\btheta^{*})\}\\
  &= 1-p_{\pi}(\x;\btheta^{*}) + \op
    = \frac{1}{1+c\varphi^{-1}(\x)e^{f(\x;\bbeta^{*})}} + \op,
\end{align*}
we have
\begin{align}
  \Var\{a_N^{-1}\dot\ell_{\lik}(\btheta^{*})\}  \longrightarrow
  &\Exp\bigg[\frac{e^{f(\x;\bbeta^{*})}\dot{g}^{\otimes2}(\x;\btheta^{*})}
    {1+c\varphi^{-1}(\x)e^{f(\x;\bbeta^{*})}}\bigg]=\bLambda_{\lik}.
    \label{eq:18}
\end{align}

Now we check the Lindeberg-Feller condition. For any $\epsilon>0$,
\begin{align*}
  &\sumN\Exp\Big[\|\delta_i\{y_i-p_{\pi}(\x_i;\btheta^{*})\}
    \dot{g}(\x_i;\btheta^{*})\|^2
    I(\|\delta_i\{y_i-p_{\pi}(\x_i;\btheta^{*})\}
    \dot{g}(\x_i;\btheta^{*})\|>a_N\epsilon)\Big]\\
  &=N\Exp\Big[\delta\|\{y-p_{\pi}(\x;\btheta^{*})\}\dot{g}(\x;\btheta^{*})\|^2
    I(\delta\|\{y-p_{\pi}(\x;\btheta^{*})\}
    \dot{g}(\x;\btheta^{*})\|>a_N\epsilon)\Big]\\
  &=N\Exp\big[p(\x;\btheta^{*})\{1-p_{\pi}(\x;\btheta^{*})\}^2\|
    \dot{g}(\x;\btheta^{*})\|^2I(\{1-p_{\pi}(\x;\btheta^{*})\}
    \|\dot{g}(\x;\btheta^{*})\|>a_N\epsilon)\big]\\
  &\quad+N\Exp\big[\{1-p(\x;\btheta^{*})\}\rho\varphi(\x)
    \{p_{\pi}(\x;\btheta^{*})\}^2\|\dot{g}(\x;\btheta^{*})\|^2
    I(\rho\varphi(\x)p_{\pi}(\x;\btheta^{*})
    \|\dot{g}(\x;\btheta^{*})\|>a_N\epsilon)\big]\\
  &\le N\Exp\big[e^{\alpha^{*}+f(\x;\bbeta^{*})}
    \|\dot{g}(\x;\btheta^{*})\|^2
    I(\|\dot{g}(\x;\btheta^{*})\|>a_N\epsilon)\big]\\
  &\quad+N\Exp\big[
    e^{\alpha^{*}+f(\x;\bbeta^{*})}
    p_{\pi}(\x;\btheta^{*})\|\dot{g}(\x;\btheta^{*})\|^2
    I(e^{\alpha^{*}+f(\x;\bbeta^{*})}\|\dot{g}(\x;\btheta^{*})\|>a_N\epsilon)\big]\\
  &\le a_N^2\Exp\big[e^{f(\x;\bbeta^{*})}\|\dot{g}(\x;\btheta^{*})\|^2
    I(\|\dot{g}(\x;\btheta^{*})\|>a_N\epsilon)\big]\\
  &\quad+a_N^2\Exp\big[e^{f(\x;\bbeta^{*})}\|\dot{g}(\x;\btheta^{*})\|^2
    I(e^{\alpha^{*}+f(\x;\bbeta^{*})}\|\dot{g}(\x;\btheta^{*})\|>a_N\epsilon)\big]\\
  &=o(a_N^2),
\end{align*}
where the inequality in the third step uses the following fact derived from (\ref{eq:s8})
\begin{equation}\label{eq:35}
  \rho\varphi(\x)p_{\pi}(\x;\btheta^{*})
  =e^{\alpha^{*}+f(\x;\bbeta^{*})}\{1-p_{\pi}(\x;\btheta^{*})\}
  \le e^{\alpha^{*}+f(\x;\bbeta^{*})},
\end{equation}
and the last step is from the dominated convergence theorem. Thus, applying the Lindeberg-Feller central limit theorem \citep[Section $^*$2.8 of][]{Vaart:98}, we finish the proof of \eqref{eq:13}.

Now we prove \eqref{eq:14}. Denote
\begin{align*}
  H_{\lik}=a_N^{-2}\sumN\delta_i\phi_{\pi}(\x_i;\btheta^{*})
    \dot{g}^{\otimes2}(\x_i;\btheta^{*}).
\end{align*}

We notice that
\begin{align}
  \Exp(H_{\lik})
  &=Na_N^{-2}\Exp[ \{y+(1-y)\rho\varphi(\x)\}
    \phi_{\pi}(\x;\btheta^{*}) \dot{g}^{\otimes2}(\x;\btheta^{*})]\notag\\
  &=e^{-\alpha^{*}}\Exp[\{1-p(\x;\btheta^{*})\}\rho\varphi(\x)
    p_{\pi}(\x;\btheta^{*})\{1-p_{\pi}(\x;\btheta^{*})\}
    \dot{g}^{\otimes2}(\x;\btheta^{*})]\notag\\
  &\qquad+e^{-\alpha^{*}}\Exp[p(\x;\btheta^{*})
    p_{\pi}(\x;\btheta^{*})\{1-p_{\pi}(\x;\btheta^{*})\}
    \dot{g}^{\otimes2}(\x;\btheta^{*})]\notag\\
  &=e^{-\alpha^{*}}\Exp[p(\x;\btheta^{*})\{1-p_{\pi}(\x;\btheta^{*})\}^2
    \dot{g}^{\otimes2}(\x;\btheta^{*})]\notag\\
  &\qquad+e^{-\alpha^{*}}\Exp[
    \{1-p(\x;\btheta^{*})\}\rho\varphi(\x)p_{\pi}^2(\x;\btheta^{*})
    \dot{g}^{\otimes2}(\x;\btheta^{*})]\notag\\
  &=\Var\{a_N^{-1}\dot\ell_{\lik}(\btheta^{*})\}=\bLambda_{\lik}+o(1),\label{eq:21}
\end{align}
where the third equality is obtained by applying \eqref{eq:20} and the forth equality is from~\eqref{eq:17}.
In addition, the variance of each component of $H_{\lik}$ is bounded by
\begin{align}
  &a_N^{-4}N\Exp\{\delta\phi_{\pi}^2(\x_i;\btheta^{*})
  \|\dot{g}(\x_i;\btheta^{*})\|^4\}\notag\\
  &\le a_N^{-4}N\Exp[\{y+(1-y)\rho\varphi(\x)\}
    p_{\pi}(\x_i;\btheta^{*})
    \|\dot{g}(\x_i;\btheta^{*})\|^4]\notag\\
  &\le a_N^{-4}N\Exp[\{p(\x;\btheta^{*})+\rho\varphi(\x)\}
    p_{\pi}(\x_i;\btheta^{*})
    \|\dot{g}(\x_i;\btheta^{*})\|^4]\notag\\
  &\le 2a_N^{-4}Ne^{\alpha^{*}}\Exp\{e^{f(\x;\bbeta^{*})}
    \|\dot{g}(\x_i;\btheta^{*})\|^4\}
    =o(1),\label{eq:22}
\end{align}
where the last inequality is because of (\ref{eq:35}) and the last step if because Assumption~\ref{asmp:1} implies that 
  $e^{f(\x;\bbeta^{*})}\|\dot{g}(\x_i;\btheta^{*})\|^4$ is integrable. %
From (\ref{eq:21}) and (\ref{eq:22}), Chebyshev's inequality implies that $H_{\lik}\rightarrow\bLambda_{\lik}$ in probability. 

\section{Proof of Theorem~\ref{thm:5}}
Let $\kappa(\x)=1+c\varphi^{-1}(\x)e^{f(\x;\bbeta^{*})}$. We first notice that
\begin{align*}
  [\Exp\{e^{f(\x;\bbeta^{*})}\}]^{-1}\V_w
  &=\M_{\f}^{-1}\M_{\f}\M_{\f}^{-1}
    +\M_{\f}^{-1}c\Lambda_{\mathrm{sub}}\M_{\f}^{-1}\\
  &=\M_{\f}^{-1}\Exp[\kappa(\x)
    e^{f(\x;\bbeta^{*})}\dot{g}^{\otimes2}(\x;\btheta^{*})]\M_{\f}^{-1},
\end{align*}
and 
\begin{equation*}
  [\Exp\{e^{f(\x;\bbeta^{*})}\}]^{-1}\V_{\lik}=\bLambda_{\lik}^{-1}
  =\big[\Exp\big\{\kappa^{-1}(\x)e^{f(\x;\bbeta^{*})}
  \dot{g}^{\otimes2}(\x;\btheta^{*})\big\}\big]^{-1}
\end{equation*}
We only need to show that
\begin{equation*}
  [\Exp\{e^{f(\x;\bbeta^{*})}\}]^{-1}\V_w\ge\bLambda_{\lik}^{-1}.
\end{equation*}
This is proved by the following calculation
\begin{align*}
  \0\le &\Exp
  \Big(\big[\{\kappa^{1/2}(\x)\M_{\f}^{-1}-\kappa^{-1/2}(\x)\bLambda_{\lik}^{-1}\}
  e^{f(\x;\bbeta^{*})/2}\dot{g}(\x;\btheta^{*})\big]^{\otimes2}\Big)\\
  =&\Exp\Big\{\M_{\f}^{-1}\kappa(\x)e^{f(\x;\bbeta^{*})}
     \dot{g}^{\otimes2}(\x;\btheta^{*})\M_{\f}^{-1}\Big\}
  +\Exp\Big\{\bLambda_{\lik}^{-1}\kappa^{-1}(\x)
  e^{f(\x;\bbeta^{*})}\dot{g}^{\otimes2}(\x;\btheta^{*})\bLambda_{\lik}^{-1}\Big\}\\
  &-2\Exp\Big\{\M_{\f}^{-1}e^{f(\x;\bbeta^{*})}
    \dot{g}^{\otimes2}(\x;\btheta^{*})\bLambda_{\lik}^{-1}\Big\}\\
  =&[\Exp\{e^{f(\x;\bbeta^{*})}\}]^{-1}\V_w + \bLambda_{\lik}^{-1}
  -2\bLambda_{\lik}^{-1}
  =[\Exp\{e^{f(\x;\bbeta^{*})}\}]^{-1}\V_w - \bLambda_{\lik}^{-1}.
\end{align*}
If $c=0$ then $\kappa(\x)=1$, and we can directly verify that 
\begin{equation}
  [\Exp\{e^{f(\x;\bbeta^{*})}\}]^{-1}\V_w=\bLambda_{\lik}^{-1}=\M_{\f}^{-1}.
\end{equation}

\section{Proof of Theorem~\ref{thm:6}}
For sampled data, (\ref{eq:5}) tell us that the joint density w.r.t. the product counting measure of the responses given the features is $\exp\big\{\ell_{\lik}(\btheta)+\sumN\delta_iy_il_i\big\}$, whose support has a finite number of possible values. Thus the expectation $\Exp\{\U(\btheta;\Dd)\mid\X\}$ is a sum of finite number of elements, and therefore the partial derivatives w.r.t. $\btheta$ can be passed under the integration sign in $\Exp\{\U(\btheta;\Dd)\mid\X\}$. Therefore we have
\begin{align*}
  \I&=\frac{\partial}{\partial\btheta\tp}
    \Exp\{\U(\btheta;\Dd)\mid\X\}\\
  &=\frac{\partial}{\partial\btheta\tp}
    \int\U(\btheta;\Dd)
    \exp\bigg\{\ell_{\lik}(\btheta)+\sumN\delta_iy_il_i\bigg\}\ud\y\\
  &=\int\dot\U(\btheta;\Dd)
    \exp\bigg\{\ell_{\lik}(\btheta)+\sumN\delta_iy_il_i\bigg\}\ud\y\\
  &\qquad  +\int\U(\btheta;\Dd)
    \exp\bigg\{\ell_{\lik}(\btheta)+\sumN\delta_iy_il_i\bigg\}
    \dot\ell_{\lik}\tp(\btheta)\ud\y\\
  &=\Exp\{\dot\U(\btheta;\Dd)\mid\X\}
    +\Exp\{\U(\btheta;\Dd)\dot\ell_{\lik}\tp(\btheta)\mid\X\},
\end{align*}
where $\ud\y=\ud y\times\ud y\times ... \ud y$ is a product measure and $\ud y$ is the counting measure. 
This gives us that
\begin{equation}\label{eq:23}
  \Exp\{\U(\btheta;\Dd)\dot\ell_{\lik}\tp(\btheta)\mid\X\}=\I
\end{equation}
Hence we have
\begin{align*}
  &\Var\big\{\U(\btheta;\Dd)
    -a_N^{-2}\bLambda_{\lik}^{-1}\dot\ell_{\lik}(\btheta)
    \bigm|\X\big\}\ge\0\\
  &=\Var\{\U(\btheta;\Dd)\mid\X\}
    +\Var\big\{a_N^{-2}\bLambda_{\lik}^{-1}\dot\ell_{\lik}(\btheta)\bigm|\X\big\}
    -2a_N^{-2}\Exp\{\U(\btheta;\Dd)\dot\ell_{\lik}\tp(\btheta)
    \mid\X\}\bLambda_{\lik}^{-1}\\
  &=\Var\{\U(\btheta;\Dd)\mid\X\}
    +a_N^{-2}\bLambda_{\lik}^{-1}\Var\{a_N^{-1}\dot\ell_{\lik}(\btheta)\bigm|\X\}
    \bLambda_{\lik}^{-1}-2a_N^{-2}\bLambda_{\lik}^{-1}.
\end{align*}
Taking expectation of the above conditional variance and using (\ref{eq:18}), we have
\begin{align*}
  &\Exp[\Var\{\U(\btheta;\Dd)\mid\X\}]
    +a_N^{-2}\bLambda_{\lik}^{-1}
    \Exp[a_N^{-1}\Var\{\dot\ell_{\lik}(\btheta)\bigm|\X\}]\bLambda_{\lik}^{-1}
    -2a_N^{-2}\bLambda_{\lik}^{-1}\\
  &=\Var\{\U(\btheta;\Dd)\} + a_N^{-2}\bLambda_{\lik}^{-1}\{1+\op\}
    - 2a_N^{-2}\bLambda_{\lik}^{-1}\\
  &=\Var\{\U(\btheta;\Dd)\} - a_N^{-2}\bLambda_{\lik}^{-1}\{1+\op\}\\
  &=\Var\{\U(\btheta;\Dd)\} - N_1^{-1}\V_{\lik}\{1+\op\}\ge\0,
\end{align*}
where the last equality is because \eqref{eq:26} implies that $a_N^{-2}=N_1^{-1}\Exp\{e^{f(\x;\bbeta^{*})}\}\{1+\op\}$. 
This finishes the proof.

\section{Proof of Theorem~\ref{thm:7}}
The outline of the proof is similar to that of the proof of Theorem~\ref{thm:2}. Write $\pi_{\varrho}^{\os}(\x,y;\tilde\bvtheta)
  =y+(1-y)\pi_{\varrho}^{\os}(\x;\tilde\bvtheta)$ and $\delta^{\tilde\bvtheta}=y+(1-y)I\{u\le\pi_{\varrho}^{\os}(\x;\tilde\bvtheta)\}$ where $u\sim\mathbb{U}(0,1)$. 
The estimator $\htheta_w^{\tilde\bvtheta}$ is the maximizer of
\begin{align*}
  \ell_w^{\tilde\bvtheta}(\btheta)
  =\sumN\frac{\delta_i^{\tilde\bvtheta}}{\pi_{\varrho}^{\os}(\x,y;\tilde\bvtheta)}
  \big[y_ig(\x_i;\btheta)-\log\{1+e^{g(\x_i;\btheta)}\}\big],
\end{align*}
so $a_N(\htheta_w^{\tilde\bvtheta}-\btheta^{*})$ is the maximizer of $\gamma_w^{\tilde\bvtheta}(\u)=\ell_w^{\tilde\bvtheta}(\btheta^{*}+a_N^{-1}\u)-\ell_w^{\tilde\bvtheta}(\btheta^{*})$. By Taylor's expansion,
\begin{align*}
  \gamma_w^{\tilde\bvtheta}(\u)
  &=\frac{1}{a_N}\u\tp\dot\ell_w^{\tilde\bvtheta}(\btheta^{*})
    +\frac{1}{2a_N^2}\sumN\frac{\delta_i^{\tilde\bvtheta}}{\pi_{\varrho}^{\os}(\x,y;\tilde\bvtheta)}
    \phi(\x_i;\btheta^{*})(\z_i\tp\u)^2
    +\Delta_{\ell_w}^{\tilde\bvtheta}+R_w^{\tilde\bvtheta},
\end{align*}
where 
\begin{align*}
  \dot\ell_w^{\tilde\bvtheta}(\btheta)
  &=\sumN\frac{\delta_i^{\tilde\bvtheta}}{\pi_{\varrho}^{\os}(\x,y;\tilde\bvtheta)}
  \big\{y_i-p(\x_i;\btheta)\big\}\dot{g}(\x_i,\btheta),\\
  \Delta_{\ell_w}^{\tilde\bvtheta}
  &=\frac{1}{2a_N^2}\sumN\frac{\delta_i^{\tilde\bvtheta}}{\pi_{\varrho}^{\os}(\x,y;\tilde\bvtheta)}
  \big\{y_i-p(\x_i;\btheta^{*})\big\}
  \u\tp\ddot{g}(\x_i,\btheta^{*})\u,
\end{align*}
and $R_w^{\tilde\bvtheta}$ is the remainder. 
Similarly to the proof of Theorem~\ref{thm:2}, we only need to show that
\begin{align}\label{eq:27}
  a_N^{-1}\dot\ell_w^{\tilde\bvtheta}(\btheta^{*}) \longrightarrow
  \Nor(\0,\ \Lambda_w^{\plt}),
\end{align}
in distribution with
\begin{align*}
  \Lambda_w^{\plt}=
  \Exp\bigg[e^{f(\x_i;\bbeta^{*})}\dot{g}^{\otimes2}(\x_i;\btheta^{*})
    +\frac{ce^{2f(\x_i;\bbeta^{*})}}{\max\{\varphi_{\plt}(\x),c_l\}}
    \dot{g}^{\otimes2}(\x_i;\btheta^{*})\bigg],
\end{align*}
and for any $\u$,
\begin{align}\label{eq:19}
  H_w^{\tilde\bvtheta}
  :=a_N^{-2}\sumN\frac{\delta_i^{\tilde\bvtheta}}
  {\pi_{\varrho}^{\os}(\x,y;\tilde\bvtheta)}
  \phi(\x_i;\btheta^{*})\dot{g}^{\otimes2}(\x_i;\btheta^{*})
  \longrightarrow \M_{\f},
\end{align}
in probability, and $\Delta_{\ell_w}^{\tilde\bvtheta}=\op$ and $R_w^{\tilde\bvtheta}=\op$. 

We prove~\eqref{eq:27} first. Let $\eta_i^{\tilde\bvtheta}=\frac{\delta_i^{\tilde\bvtheta}}{\pi_{\varrho}^{\os}(\x,y;\tilde\bvtheta)}\{y_i-p(\x_i;\btheta^{*})\}\dot{g}(\x_i;\btheta^{*})$. Given $\tilde\bvtheta$, $\eta_i^{\tilde\bvtheta}$, $i=1, ..., N$, are i.i.d. with mean zero, and the variance satisfies that
\begin{align*}
  &\Var(\eta_i^{\tilde\bvtheta}\mid\tilde\bvtheta)\\
  &=\Exp\bigg[\frac{\{y_i-p(\x_i;\btheta^{*})\}^2}
    {\{y_i+(1-y_i)\pi_{\varrho}^{\os}(\x_i;\tilde\bvtheta)\}}
    \dot{g}^{\otimes2}(\x_i;\btheta^{*})\biggm|\tilde\bvtheta\bigg]\\
  &=\Exp\bigg[p(\x_i;\btheta^{*})\{1-p(\x_i;\btheta^{*})\}^2
    \dot{g}^{\otimes2}(\x_i;\btheta^{*})
    +\{1-p(\x_i;\btheta^{*})\}
    \frac{p^2(\x_i;\btheta^{*})}{\pi_{\varrho}^{\os}(\x_i;\tilde\bvtheta)}
    \dot{g}^{\otimes2}(\x_i;\btheta^{*})\biggm|\tilde\bvtheta\bigg]\\
  &\le\Exp\big\{p(\x_i;\btheta^{*})\dot{g}^{\otimes2}(\x_i;\btheta^{*})
    +\varrho^{-1}p^2(\x_i;\btheta^{*})\dot{g}^{\otimes2}(\x_i;\btheta^{*})
  \bigm|\tilde\bvtheta\big\}\\
  &\le e^{\alpha^{*}}\Exp\big\{
    e^{f(\x_i;\bbeta^{*})}\dot{g}^{\otimes2}(\x_i;\btheta^{*})
    +(\varrho^{-1}e^{\alpha^{*}})e^{2f(\x_i;\bbeta^{*})}
    \dot{g}^{\otimes2}(\x_i;\btheta^{*})\big\}
\end{align*}
Since $\rho^{-1}\varrho\rightarrow c_l>0$, for large enough $N$,
\begin{align*}
  \Var\{a_N^{-1}\dot\ell_w^{\tilde\bvtheta}(\btheta^{*})\mid\tilde\bvtheta\}
  \le \Exp\big\{e^{f(\x_i;\bbeta^{*})}\dot{g}^{\otimes2}(\x_i;\btheta^{*})
    +(1+c_l^{-1}c)e^{2f(\x_i;\bbeta^{*})}
    \dot{g}^{\otimes2}(\x_i;\btheta^{*})\big\}.
\end{align*}
Thus,  we know that
\begin{align*}
  \Var\{a_N^{-1}\dot\ell_w^{\tilde\bvtheta}(\btheta^{*})\}
  \rightarrow \Lambda_w^{\plt}
\end{align*}

Now we check the Lindeberg-Feller condition \citep[Section $^*$2.8 of][]{Vaart:98} given $\tilde\bvtheta$. 
For any $\epsilon>0$,
\begin{align*}
  &\sumN\Exp\big\{\|\eta_i^{\tilde\bvtheta}\|^2
    I(\|\eta_i^{\tilde\bvtheta}\|>a_N\epsilon)\bigm|\tilde\bvtheta\big\}\\
  &=N\Exp\big[\|\{\pi_{\varrho}^{\os}(\x,y;\tilde\bvtheta)\}^{-1}
    \delta^{\tilde\bvtheta}\{y-p(\x;\btheta^{*})\}\dot{g}(\x;\btheta^{*})\|^2\\
  &\hspace{2cm}\times
    I(\|\{\pi_{\varrho}^{\os}(\x,y;\tilde\bvtheta)\}^{-1}
    \delta^{\tilde\bvtheta}\{y-p(\x;\btheta^{*})\}
    \dot{g}(\x;\btheta^{*})\|>a_N\epsilon)\bigm|\tilde\bvtheta\big]\\
  &= N\Exp\Big[\{\pi_{\varrho}^{\os}(\x,y;\tilde\bvtheta)\}^{-1}
    \|\{y-p(\x;\btheta^{*})\}\dot{g}(\x;\btheta^{*})\|^2\\
  &\hspace{2cm}\times I(\|\{\pi_{\varrho}^{\os}(\x,y;\tilde\bvtheta)\}^{-1}
    \{y-p(\x;\btheta^{*})\}\dot{g}(\x;\btheta^{*})\|
    >a_N\epsilon)\bigm|\tilde\bvtheta\Big]\\
  &\quad+N\Exp\Big[\{1-\pi_{\varrho}^{\os}(\x,y;\tilde\bvtheta)\}\|
    \{\pi_{\varrho}^{\os}(\x,y;\tilde\bvtheta)\}^{-1}
    y\{y-p(\x;\btheta^{*})\}\dot{g}(\x;\btheta^{*})\|^2\\
  &\hspace{4cm}\times
    I(\|\{\pi_{\varrho}^{\os}(\x,y;\tilde\bvtheta)\}^{-1}y\{y-p(\x;\btheta^{*})\}
    \dot{g}(\x;\btheta^{*})\| >a_N\epsilon)\bigm|\tilde\bvtheta\Big]\\
  &\le N\Exp\Big[p(\x;\btheta^{*})\|\dot{g}(\x;\btheta^{*})\|^2
    I(\|\dot{g}(\x;\btheta^{*})\| >a_N\epsilon)\Big]\\
  &\quad+N\Exp\Big[\varrho^{-1}
    \|p(\x;\btheta^{*})\dot{g}(\x;\btheta^{*})\|^2
    I(\|\varrho^{-1}p(\x;\btheta^{*})
    \dot{g}(\x;\btheta^{*})\| >a_N\epsilon)\Big]\\
  &\quad+N\Exp\Big[p(\x;\btheta^{*})\|\dot{g}(\x;\btheta^{*})\|^2
  I(\|\dot{g}(\x;\btheta^{*})\|>a_N\epsilon)\Big]\\
  &\le 2a_N^2\Exp\Big[e^{f(\x_i;\bbeta^{*})}\|\dot{g}(\x;\btheta^{*})\|^2
    I(\|\dot{g}(\x;\btheta^{*})\| >a_N\epsilon)\Big]\\
  &\quad+a_N^2e^{\alpha^{*}}\varrho^{-1}\Exp\Big[ e^{2f(\x_i;\bbeta^{*})}
    \|\dot{g}(\x;\btheta^{*})\|^2
    I(\varrho^{-1}e^{\alpha^{*}}\|e^{f(\x_i;\bbeta^{*})}
    \dot{g}(\x;\btheta^{*})\| >a_N\epsilon)\Big]\\
  &=o(a_N^2),
\end{align*}
where the last step is due to Assumptions~\ref{asmp:1},
the dominated convergence theorem, and the facts that $a_N\rightarrow\infty$ and $\lim_{N\rightarrow\infty}e^\alpha/\varrho=c/c_l<\infty$. Thus, applying the Lindeberg-Feller central limit theorem \citep[Section $^*$2.8 of][]{Vaart:98} finishes the proof of \eqref{eq:27}.

Now we prove~\eqref{eq:19}. %
We notice that
\begin{align}\label{eq:36}
  \Exp(H_w^{\tilde\bvtheta}\mid\tilde\bvtheta)=
  &\Exp\bigg\{\frac{e^{f(\x;\bbeta^{*})}}
     {(1+e^{\alpha^{*}+f(\x;\bbeta^{*})})^2}\dot{g}^{\otimes2}(\x;\btheta^{*})\bigg\}
  =\M_{\f}+o(1),
\end{align}
where the last step is by the dominated convergence theorem. In addition, the conditional variance of each component of $H_w^{\tilde\bvtheta}$ is bounded by 
\begin{align}\label{eq:37}
  \oneN\Exp\bigg\{
  \frac{e^{2f(\x;\bbeta^{*})}\|\dot{g}(\x;\btheta^{*})\|^4}
  {\pi_{\varrho}^{\os}(\x,y;\tilde\bvtheta)}
    \biggm|\tilde\bvtheta\bigg\}
  &\le\frac{1}{N\varrho}\Exp\{
    e^{2f(\x;\bbeta^{*})}\|\dot{g}(\x;\btheta^{*})\|^4\}
    =o(1).
\end{align}
where the last step is because of Assumption~\ref{asmp:1} and the fact that $Ne^{\alpha^{*}}\rightarrow\infty$, $e^{\alpha^{*}}/\rho\rightarrow c<\infty$, and $\rho^{-1}\varrho\rightarrow c_l>0$ imply that $N\varrho\rightarrow\infty$. 
From (\ref{eq:36}) and (\ref{eq:37}), applying Chebyshev's inequality finishes the proof of \eqref{eq:19}.

In the following we finish the proof by showing that given $\tilde\bvtheta$, $\Delta_{\ell_w}^{\tilde\bvtheta}=\op$ and $R_w^{\tilde\bvtheta}=\op$.

For $\Delta_{\ell_w}^{\tilde\bvtheta}$, the conditional mean $\Exp(\Delta_{\ell_w}^{\tilde\bvtheta}\mid\tilde\bvtheta)=\0$ and the conditional variance satisfies that
\begin{align*}
  \Var(\Delta_{\ell_w}^{\tilde\bvtheta}\mid\tilde\bvtheta)
  &\le\frac{\|\u\|^4N}{4a_N^4}\Exp\bigg[
    \frac{\{y-p(\x;\btheta^{*})\}^2\|\ddot{g}(\x;\btheta^{*})\|^2}
    {y+(1-y)\pi_{\varrho}^{\os}(\x,y;\tilde\bvtheta)}\biggm|\tilde\bvtheta\bigg]\\
  &\le\frac{\|\u\|^4N}{4a_N^4}\Exp\big[
    \{1+\varrho^{-1}p(\x;\btheta^{*})\}
    p(\x;\btheta^{*})\|\ddot{g}(\x;\btheta^{*})\|^2\big]\\
  &\le\frac{\|\u\|^4}{4a_N^2}\Exp\big[
    \{1+\varrho^{-1}e^{\alpha^{*}}e^{f(\x;\bbeta^{*})}\}
    e^{f(\x;\bbeta^{*})}\|\ddot{g}(\x;\btheta^{*})\|^2\big]
  =o(1),
\end{align*}
so $\Delta_{\ell_w}^{\tilde\bvtheta}=\op$.  %

For the remainder term $R_w^{\tilde\bvtheta}$. By direct calculations similar to those used for the proof of Theorem~\ref{thm:2}, we know that
\begin{align*}
  |R_w^{\tilde\bvtheta}|
  &\le\frac{d^3\|\u\|^3e^{a_N^{-1}\|\u\|}}{6Na_N}
    \sumN\frac{\delta_i^{\tilde\bvtheta}}{\pi_{\varrho}^{\os}(\x,y;\tilde\bvtheta)} B(\x_i)
    +\frac{d\|\u\|^3}{6a_N^3}\sumN\frac{\delta_i^{\tilde\bvtheta}}
    {\pi_{\varrho}^{\os}(\x,y;\tilde\bvtheta)} y_i B(\x_i)\\
    &\equiv\Delta_{R_w1}^{\tilde\bvtheta}+\Delta_{R_w2}^{\tilde\bvtheta}.
\end{align*}

From Assumption~\ref{asmp:1}, $\Exp(\Delta_{R_w1}^{\tilde\bvtheta})\rightarrow0$ and $\Exp(\Delta_{R_w2}^{\tilde\bvtheta})\rightarrow0$.
Since $\Delta_{R_w1}^{\tilde\bvtheta}$ and $\Delta_{R_w2}^{\tilde\bvtheta}$ are both positive, Markov's inequality shows that they are both $\op$ so $R_w^{\tilde\bvtheta}=\op$.

\section{Proof of Theorem~\ref{thm:8}}
The outline of the proof is similar to that of the proof of Theorem~\ref{thm:4}. 
The estimator $\htheta_{\lik}^{\tilde\bvtheta}$ is the maximizer of $\ell_{\lik}^{\tilde\bvtheta}(\btheta)$, so $\u_N^{\tilde\bvtheta}=a_N^{-1}(\htheta_{\lik}^{\tilde\bvtheta}-\btheta)$ 
is the maximizer of
\begin{equation*}
  \gamma_{\lik}^{\tilde\bvtheta}(\u)=\ell_{\lik}^{\tilde\bvtheta}(\btheta+a_N^{-1}\u)-\ell_{\lik}^{\tilde\bvtheta}(\btheta).
\end{equation*}
By Taylor's expansion,
\begin{equation}
  \gamma_{\lik}^{\tilde\bvtheta}(\u) =a_N^{-1}\u\tp\dot\ell_{\lik}^{\tilde\bvtheta}(\btheta^{*})
    +0.5a_N^{-2}\sumN \phi_{\pi}(\x_i;\btheta^{*})
    \{\u\tp\dot{g}(\x_i;\btheta^{*})\}^2
    +\Delta_{\ell_{\lik}}^{\tilde\bvtheta}+R_{\lik}^{\tilde\bvtheta},
\end{equation}
where $\phi_{\pi}^{\tilde\bvtheta}(\x;\btheta)
=p_{\pi}^{\tilde\bvtheta}(\x;\btheta)\{1-p_{\pi}^{\tilde\bvtheta}(\x;\btheta)\}$,
\begin{equation}\label{eq:25}
  p_{\pi}^{\tilde\bvtheta}(\x;\btheta)
    =\frac{e^{g(\x;\btheta)+\tilde{l}}}{1+e^{g(\x;\btheta)+\tilde{l}}},
    \quad\text{with}\quad
    \tilde{l}_i=-\log\{\pi_{\varrho}^{\os}(\x;\tilde\bvtheta)\},
\end{equation}
\begin{equation*}
  \dot\ell_{\lik}^{\tilde\bvtheta}(\btheta)
  =\sumN\delta_i^{\tilde\bvtheta}\big\{y_i-p_{\pi}^{\tilde\bvtheta}(\x_i;\btheta)\big\}
  \dot{g}(\x_i;\btheta),
\end{equation*}
\begin{align}
  \Delta_{\ell_{\lik}}^{\tilde\bvtheta} =\frac{1}{2a_N^2}\sumN\delta_i^{\tilde\bvtheta}
  \big\{y_i-p_{\pi}^{\tilde\bvtheta}(\x_i;\btheta^{*})\big\}
  \u\tp\ddot{g}(\x_i;\btheta^{*})\u,
\end{align}
and $R_{\lik}^{\tilde\bvtheta}$ is the remainder term. By direct calculations similar to those used for the proof of Theorem~\ref{thm:4}, we know that
\begin{align*}
  |R_{\lik}^{\tilde\bvtheta}|
  &\le\frac{\|\u\|^3}{6a_N^3}\sumN
    \delta_i^{\tilde\bvtheta}%
    C(\x_i,\btheta^{*}+a_N^{-1}\Acute\u)
  +\frac{d\|\u\|^3}{6a_N^3}\sumN\delta_i^{\tilde\bvtheta}y_i B(\x_i)
  \le\frac{(d+1)\|\u\|^3}{6a_N^3}\sumN\delta_i^{\tilde\bvtheta}B(\x_i),
\end{align*}
where $\Acute\u$ lies between $\0$ and $\u$.

Since $\varrho=o(\rho)$, we know that for large enough $N$,
\begin{align*}
  &\Exp\Big\{a_N^{-3}\sumN\delta_i^{\tilde\bvtheta}B(\x_i)
    \Bigm|\tilde\bvtheta\Big\}\\
  &=Na_N^{-3}\Exp\Big\{
    [y+(1-y)\pi_{\varrho}^{\os}(\x;\tilde\bvtheta)]
    B(\x)\Bigm|\tilde\bvtheta\Big\}\\
  &\le Na_N^{-3}\Exp\Big\{
    [e^{\alpha^{*}}e^{f(\x;\bbeta^{*})}+ \tilde\omega^{-1}\rho t(\x;\ttheta)]
    B(\x)\Bigm|\tilde\bvtheta\Big\}\\
  &\le a_N^{-1}\Exp\big\{e^{f(\x;\bbeta^{*})}B(\x)\big\}
    +a_N^{-1}\tilde\lambda_{\min}^{-1}(e^{-\alpha^{*}}\rho)
    (e^{\tilde\alpha}\tilde\omega)^{-1}
    \Exp\big\{e^{f(\x;\tilde\btheta)}\|\dot{g}(\x;\tilde\btheta)\|
    B(\x)\bigm|\tilde\bvtheta\big\}\\
  &\le a_N^{-1}\Exp\big\{e^{f(\x;\bbeta^{*})}B(\x)\big\}
    +a_N^{-1}\tilde\lambda_{\min}^{-1}(e^{-\alpha^{*}}\rho)
    (e^{\tilde\alpha}\tilde\omega)^{-1}\Exp\{B^2(\x)\}
  =\op,
\end{align*}
where $\tilde\lambda_{\min}$ is the minimum eigenvalue of $\tilde\M_{\f}$.
Thus, Markov's inequality shows that $R_{\lik}^{\tilde\bvtheta}=\op$. 
For $\Delta_{\ell_{\lik}}^{\tilde\bvtheta}$, the conditional mean $\Exp(\Delta_{\ell_{\lik}}^{\tilde\bvtheta}\mid\tilde\bvtheta)=\0$ and the conditional variance satisfies that
\begin{align*}
  \Var(\Delta_{\ell_{\lik}}^{\tilde\bvtheta}\mid\tilde\bvtheta)
  &\le\frac{\|\u\|^4}{4a_N^4}
    \sumN\Exp\{\delta_i^{\tilde\bvtheta}p_{\pi}^{\tilde\bvtheta}(\x_i;\btheta^{*})
  \|\ddot{g}(\x_i;\btheta^{*})\|^2\mid\tilde\bvtheta\}\\
  &\le\frac{\|\u\|^4}{4a_N^2}\Exp[e^{f(\x;\bbeta^{*})}
  \|\ddot{g}(\x,\btheta^{*})\|^2]\rightarrow0,
\end{align*}
so $\Delta_{\ell_{\lik}}^{\tilde\bvtheta}=\op$.

Now we need to show that given $\tilde\bvtheta$
\begin{align}\label{eq:28}
  a_N^{-1}\dot\ell_{\lik}^{\tilde\bvtheta}(\btheta^{*})
  \longrightarrow \Nor\big(\0,\ \bLambda_{\lik}^{\plt}\big),
\end{align}
in distribution, with
\begin{align*}
  \bLambda_{\lik}^{\plt}
  =\Exp\bigg[\frac{e^{f(\x;\bbeta^{*})}\dot{g}^{\otimes2}(\x;\btheta^{*})}
  {1+c\varphi_{\plt}^{-1}(\x) e^{f(\x;\bbeta^{*})}}\bigg],
\end{align*}
and
\begin{align}\label{eq:29}
  H_{\lik}^{\tilde\bvtheta}
  :=a_N^{-2}\sumN\delta_i^{\tilde\bvtheta}\phi_{\pi}^{\tilde\bvtheta}(\x_i;\btheta^{*})
  \dot{g}^{\otimes2}(\x_i;\btheta^{*})
 \longrightarrow \bLambda_{\lik}^{\plt},
\end{align}
in probability, then from the Basic Corollary in page 2 of \cite{hjort2011asymptotics}, we know that $a_N(\htheta_{\lik}^{\tilde\bvtheta}-\btheta^{*})$, the maximizer of $\gamma_{\lik}^{\tilde\bvtheta}(\u)$, satisfies that
\begin{equation}\label{eq:42}
  a_N(\htheta_{\lik}^{\tilde\bvtheta}-\btheta^{*})
  =(\bLambda_{\lik}^{\plt})^{-1}\times a_N^{-1}\dot\ell_{\lik}^{\tilde\bvtheta}(\btheta^{*})+\op.
\end{equation}
Slutsky's theorem together with (\ref{eq:28}) and (\ref{eq:42}) implies the result in Theorem~\ref{thm:1}. We prove \eqref{eq:28} and \eqref{eq:29} in the following. 

Given $\tilde\bvtheta$, $\dot\ell_{\lik}^{\tilde\bvtheta}(\btheta^{*})$ is a summation of i.i.d. quantities, $\delta_i^{\tilde\bvtheta}\big\{y_i-p_{\pi}^{\tilde\bvtheta}(\x_i;\btheta^{*})\big\}\dot{g}(\x_i;\btheta^{*})$'s. Using a similar approach to obtain (\ref{eq:43})
we know that $\Exp\{a_N^{-1}\dot\ell_{\lik}^{\tilde\bvtheta}(\btheta^{*})\mid\tilde\bvtheta\}=\0$.
The conditional variance of $a_N^{-1}\dot\ell_{\lik}^{\tilde\bvtheta}(\btheta^{*})$ satisfies that
\begin{align}
  &\Var\{a_N^{-1}\dot\ell_{\lik}^{\tilde\bvtheta}(\btheta^{*})
    \mid\tilde\bvtheta\}\notag\\
  &=e^{-\alpha^{*}}\Exp\big[
    \{y+(1-y)\pi_{\varrho}^{\os}(\x;\tilde\bvtheta)\}
    \{y-p_{\pi}^{\tilde\bvtheta}(\x;\btheta^{*})\}^2
    \dot{g}^{\otimes2}(\x;\btheta^{*})\bigm|\tilde\bvtheta\big]\notag\\
  &=e^{-\alpha^{*}}\Exp\big([
    p(\x;\btheta^{*})\{1-p_{\pi}^{\tilde\bvtheta}(\x;\btheta^{*})\}^2
    +\pi_{\varrho}^{\os}(\x;\tilde\bvtheta)
    \{1-p(\x;\btheta^{*})\}p_{\pi}^2(\x;\btheta^{*})]
    \dot{g}^{\otimes2}(\x;\btheta^{*})\Bigm|\tilde\bvtheta\big)\notag\\
  &=e^{-\alpha^{*}}\Exp\big([\{1-p_{\pi}^{\tilde\bvtheta}(\x;\btheta^{*})\}^2
    +\phi_{\pi}^{\tilde\bvtheta}(\x;\btheta^{*})\{1-p(\x;\btheta^{*})\}^2]
    p(\x;\btheta^{*})\dot{g}^{\otimes2}(\x;\btheta^{*})
    \bigm|\tilde\bvtheta\big)\notag\\
  &=\Exp\big([\{1-p_{\pi}^{\tilde\bvtheta}(\x;\btheta^{*})\}^2
    +\phi_{\pi}^{\tilde\bvtheta}(\x;\btheta^{*})\{1-p(\x;\btheta^{*})\}^2]
    \{1-p(\x;\btheta^{*})\}
    e^{f(\x;\bbeta^{*})}\dot{g}^{\otimes2}(\x;\btheta^{*})
    \bigm|\tilde\bvtheta\big).\notag
\end{align}
The elements of the term on the right hand side of the above equation
are bounded by an integrable random variable $e^{f(\x;\bbeta^{*})}\|\dot{g}(\x;\btheta^{*})\|^2$, so if $\varrho=o(\rho)$ then
\begin{align}
  \Var\{a_N^{-1}\dot\ell_{\lik}^{\tilde\bvtheta}(\btheta^{*})\}
  \longrightarrow \bLambda_{\lik}^{\plt}. \label{eq:32}
\end{align}

Using an similar approach as in checking the Lindeberg-Feller condition in the proof of Theorem~\ref{thm:4}, we obtain that for any $\epsilon>0$,
\begin{align*}
  &\sumN\Exp\Big[\|\delta_i^{\tilde\bvtheta}
    \{y_i-p_{\pi}^{\tilde\bvtheta}(\x_i;\btheta^{*})\}
    \dot{g}(\x_i;\btheta^{*})\|^2
    I(\|\delta_i^{\tilde\bvtheta}\{y_i-p_{\pi}^{\tilde\bvtheta}(\x_i;\btheta^{*})\}
    \dot{g}(\x_i;\btheta^{*})\|>a_N\epsilon)\Bigm|\tilde\bvtheta\Big]\\
  &=N\Exp\Big[\delta^{\tilde\bvtheta}
    \|\{y-p_{\pi}^{\tilde\bvtheta}(\x;\btheta^{*})\}\dot{g}(\x;\btheta^{*})\|^2
    I(\delta^{\tilde\bvtheta}\|\{y-p_{\pi}^{\tilde\bvtheta}(\x;\btheta^{*})\}
    \dot{g}(\x;\btheta^{*})\|>a_N\epsilon)\Bigm|\tilde\bvtheta\Big]\\
  &=N\Exp\big[p(\x;\btheta^{*})\{1-p_{\pi}^{\tilde\bvtheta}(\x;\btheta^{*})\}^2\|
    \dot{g}(\x;\btheta^{*})\|^2I(\{1-p_{\pi}^{\tilde\bvtheta}(\x;\btheta^{*})\}
    \|\dot{g}(\x;\btheta^{*})\|>a_N\epsilon)\bigm|\tilde\bvtheta\big]\\
  &\quad+N\Exp\Big[\{1-p(\x;\btheta^{*})\}\pi_{\varrho}^{\os}(\x;\tilde\bvtheta)
    \{p_{\pi}^{\tilde\bvtheta}(\x;\btheta^{*})\}^2\|\dot{g}(\x;\btheta^{*})\|^2\\
  &\hspace{5cm}\times
    I(\pi_{\varrho}^{\os}(\x;\tilde\bvtheta)p_{\pi}^{\tilde\bvtheta}(\x;\btheta^{*})
    \|\dot{g}(\x;\btheta^{*})\|>a_N\epsilon)\Bigm|\tilde\bvtheta\Big]\\
  &\le N\Exp\big[p(\x;\btheta^{*})\|
    \dot{g}(\x;\btheta^{*})\|^2I(\|\dot{g}(\x;\btheta^{*})\|>a_N\epsilon)
    \bigm|\tilde\bvtheta\big]\\
  &\quad+N\Exp\big[e^{\alpha^{*}+f(\x;\bbeta^{*})}
    p_{\pi}^{\tilde\bvtheta}(\x;\btheta^{*})\|\dot{g}(\x;\btheta^{*})\|^2
    I(e^{\alpha^{*}+f(\x;\bbeta^{*})}
    \|\dot{g}(\x;\btheta^{*})\|>a_N\epsilon)\bigm|\tilde\bvtheta\big]\\
  &\le a_N^2\Exp\{e^{f(\x;\bbeta^{*})}\|\dot{g}(\x;\btheta^{*})\|^2
    I(\|\dot{g}(\x;\btheta^{*})\|>a_N\epsilon)\}\\
  &\quad+a_N^2\Exp\big[e^{f(\x;\bbeta^{*})}\|\dot{g}(\x;\btheta^{*})\|^2
    I(e^{\alpha^{*}+f(\x;\bbeta^{*})}
    \|\dot{g}(\x;\btheta^{*})\|>a_N\epsilon)\big]\\
  &=o(a_N^2),
\end{align*}
where the last step is from the dominated convergence theorem. Thus, applying the Lindeberg-Feller central limit theorem \citep[Section $^*$2.8 of][]{Vaart:98}, we finish the proof of \eqref{eq:28}.

Now we prove \eqref{eq:29}. %
By similar derivations in~(\ref{eq:21}), we know that
\begin{align}
  \Exp(H_{\lik}^{\tilde\bvtheta}\mid\tilde\bvtheta)
  &=\Var\{a_N^{-1}\dot\ell_{\lik}^{\tilde\bvtheta}(\btheta^{*})
    \mid\tilde\bvtheta\}=\bLambda_{\lik}^{\plt}+\op.\label{eq:33}
\end{align}
where the second equality is from~\eqref{eq:32}.
In addition, the conditional variance of each component of $H_{\lik}^{\tilde\bvtheta}$ is bounded by
\begin{align}
  &a_N^{-4}N\Exp\{\delta^{\tilde\bvtheta}
    \{\phi_{\pi}^{\tilde\bvtheta}(\x_i;\btheta^{*})\}^2
    \|\dot{g}(\x_i;\btheta^{*})\|^4\mid\tilde\bvtheta\}\notag\\
  &\le a_N^{-4}N\Exp[\{y+(1-y)\pi_{\varrho}^{\os}(\x;\tilde\bvtheta)\}
    p_{\pi}^{\tilde\bvtheta}(\x_i;\btheta^{*})
    \|\dot{g}(\x_i;\btheta^{*})\|^4\mid\tilde\bvtheta]\notag\\
  &\le a_N^{-4}N\Exp[
    \{p(\x;\btheta^{*})+\pi_{\varrho}^{\os}(\x;\tilde\bvtheta)\}
    p_{\pi}^{\tilde\bvtheta}(\x_i;\btheta^{*})
    \|\dot{g}(\x_i;\btheta^{*})\|^4\mid\tilde\bvtheta]\notag\\
  &\le 2a_N^{-4}Ne^{\alpha^{*}}\Exp\{
    e^{f(\x;\bbeta^{*})}\|\dot{g}(\x_i;\btheta^{*})\|^4\}
  =2a_N^{-2}\Exp\{e^{f(\x;\bbeta^{*})}\|\dot{g}(\x_i;\btheta^{*})\|^4\}
    =o(1).\label{eq:34}
\end{align}
From (\ref{eq:33}) and (\ref{eq:34}), Chebyshev's inequality implies that $H_{\lik}^{\tilde\bvtheta}\rightarrow\bLambda_{\lik}^{\plt}$ in probability. 

\section{Experiments on Pilot Mis-specification}
This section complements Section 6.1 in the main text. We generate all data with a logistic regression model with $g(\x;\btheta)=\alpha+\x\tp\bbeta$. The total number of data is $N=5\times 10^5$ with $N_0$ negative samples and $N_1$ positive samples. When generating data, we tune the bias $\alpha$ to force $N_0/N_1\approx 400$. In order to test the robustness of our methods, we add an artificial biased to the pilot model, which will produce skewed sampling probabilities through the equation
\begin{equation}
  \pi_{\varrho}^{\os}(\x;\tilde\bvtheta)
  =\min[\max\{\rho\tilde\varphi_{\os}(\x),\varrho\},1],
  \quad\text{ where }\quad
  \tilde\varphi_{\os}(\x)=\tilde\omega^{-1}t(\x;\ttheta).
\end{equation}

We try two types of biases: one on the intercept term $\alpha$ and the other on weights $\bbeta$.

\subsection{Perturb the pilot intercept $\tilde\alpha$.}
We perturb $\tilde\alpha\leftarrow \tilde\alpha + \xi\times \mathbb{U}(0,1)\times \log(N_0/N_1)$, where $\xi$ is a hyperparameter controlling the degree of perturbation. We try $\xi=\{0.1, 0.5, 1.0\}$. As the result will show, perturbing the intercept will not affect the relative performance for various types of sampling algorithms.

\begin{figure}[H]\centering
  \begin{minipage}{0.23\textwidth}\centering
    \includegraphics[width=\textwidth]{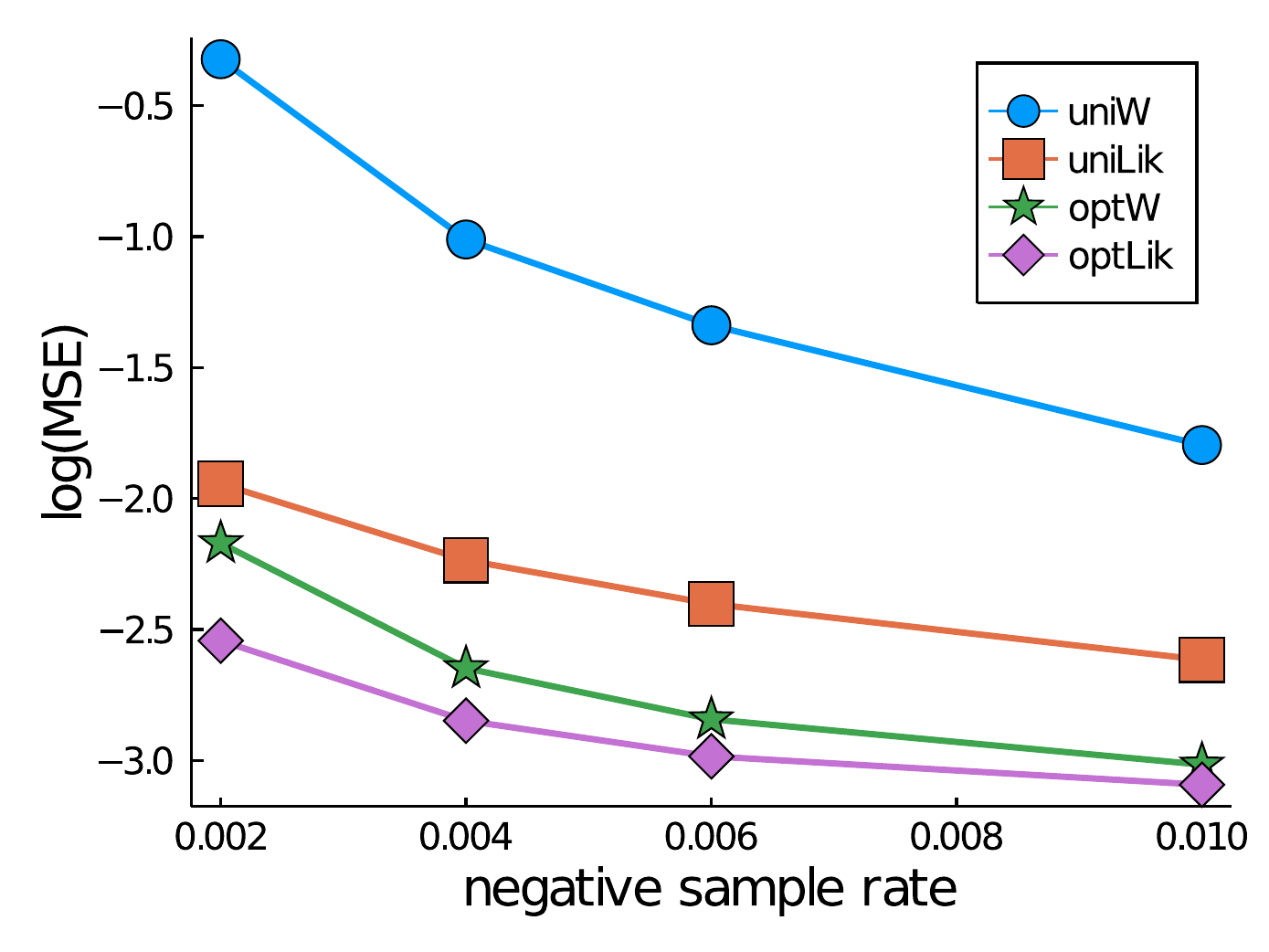}
    (a). {$\x$'s are normal}
  \end{minipage} %
  \begin{minipage}{0.23\textwidth}\centering
    \includegraphics[width=\textwidth]{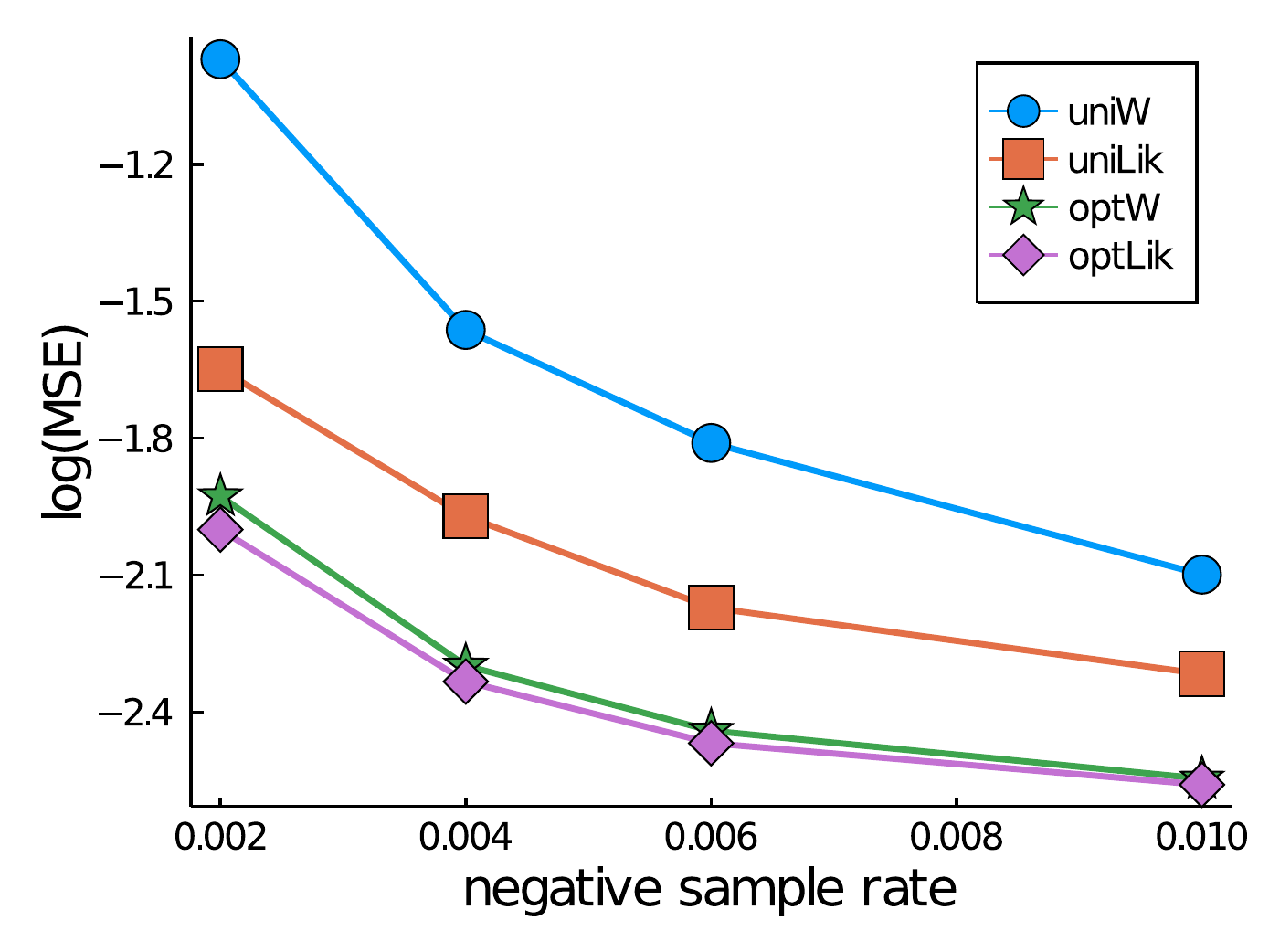}
    (b) {$\x$'s are lognormal}
  \end{minipage} %
  \begin{minipage}{0.23\textwidth}\centering
    \includegraphics[width=\textwidth]{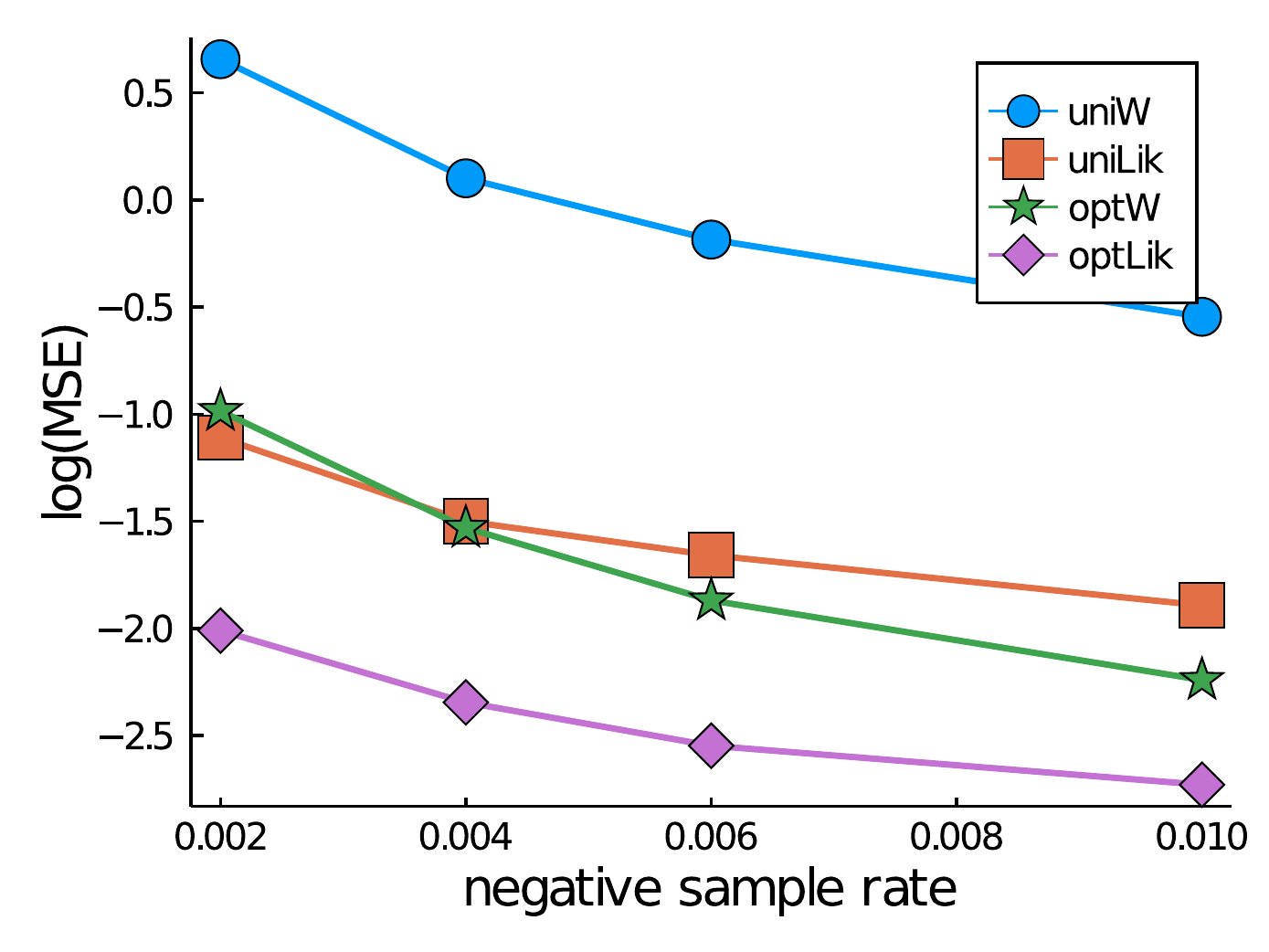}
    (c) {$\x$'s are $t_3$}
  \end{minipage}
  \begin{minipage}{0.23\textwidth}\centering
    \includegraphics[width=\textwidth]{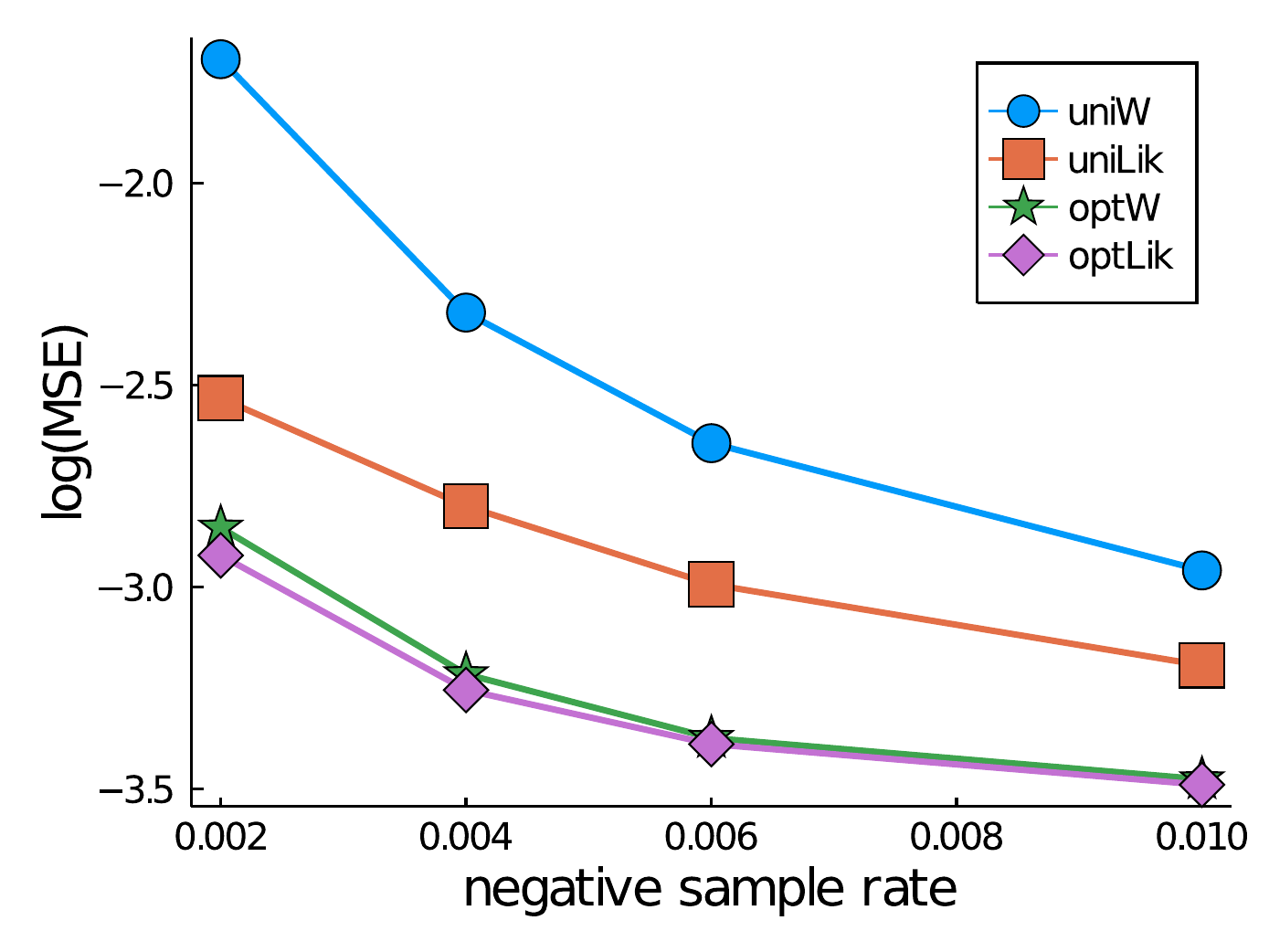}
    (d) {$\x$'s are exponential}
  \end{minipage}
  \caption{Degree of perturbation $\xi=0.1$. Log (MSEs) of subsample estimators for different sample sizes (the smaller the better).}
  \label{fig:supp_perturb_alpha_0_1}
\end{figure}

\begin{figure}[H]\centering
  \begin{minipage}{0.23\textwidth}\centering
    \includegraphics[width=\textwidth]{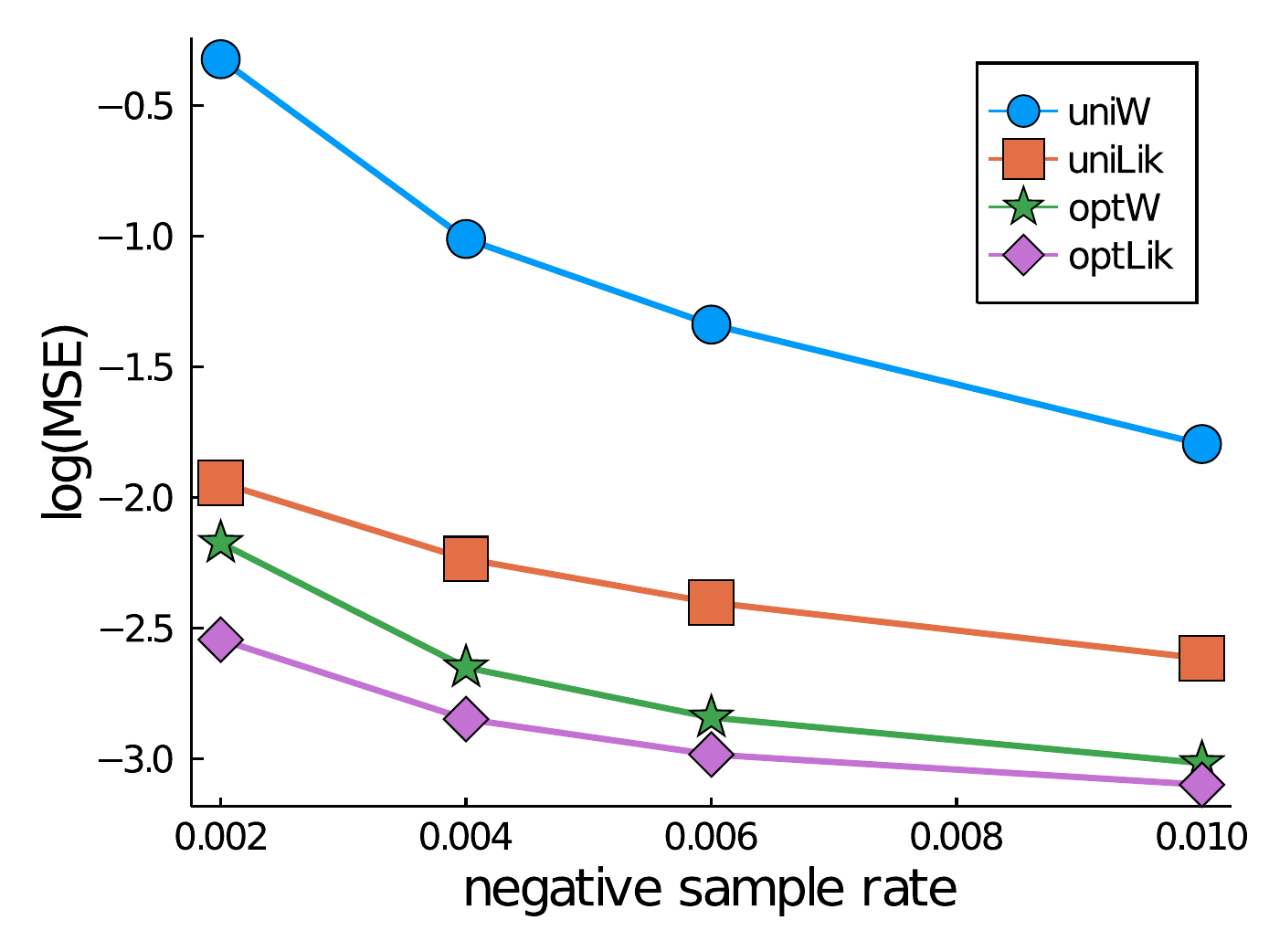}
    (a). {$\x$'s are normal}
  \end{minipage} %
  \begin{minipage}{0.23\textwidth}\centering
    \includegraphics[width=\textwidth]{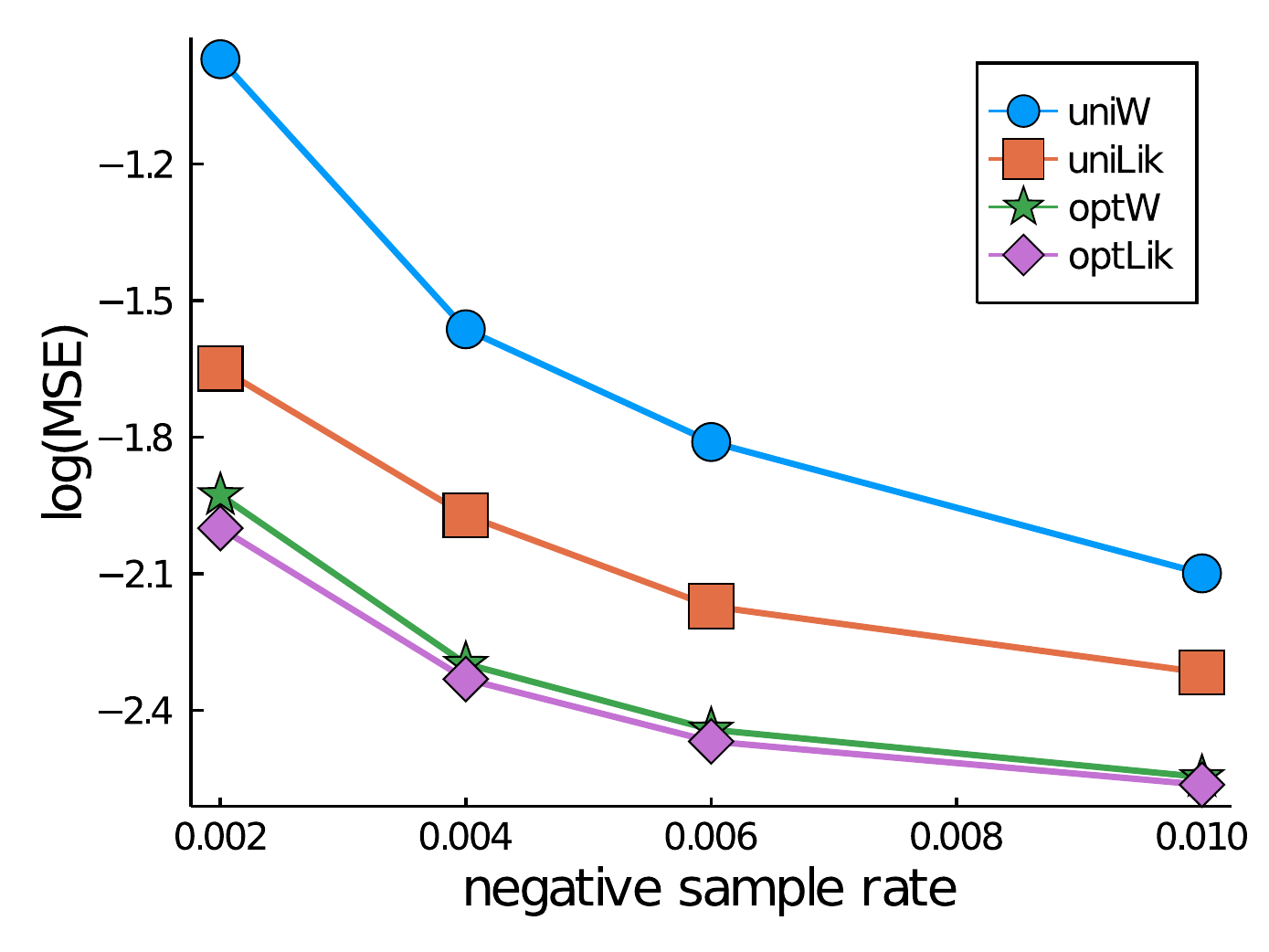}
    (b) {$\x$'s are lognormal}
  \end{minipage} %
  \begin{minipage}{0.23\textwidth}\centering
    \includegraphics[width=\textwidth]{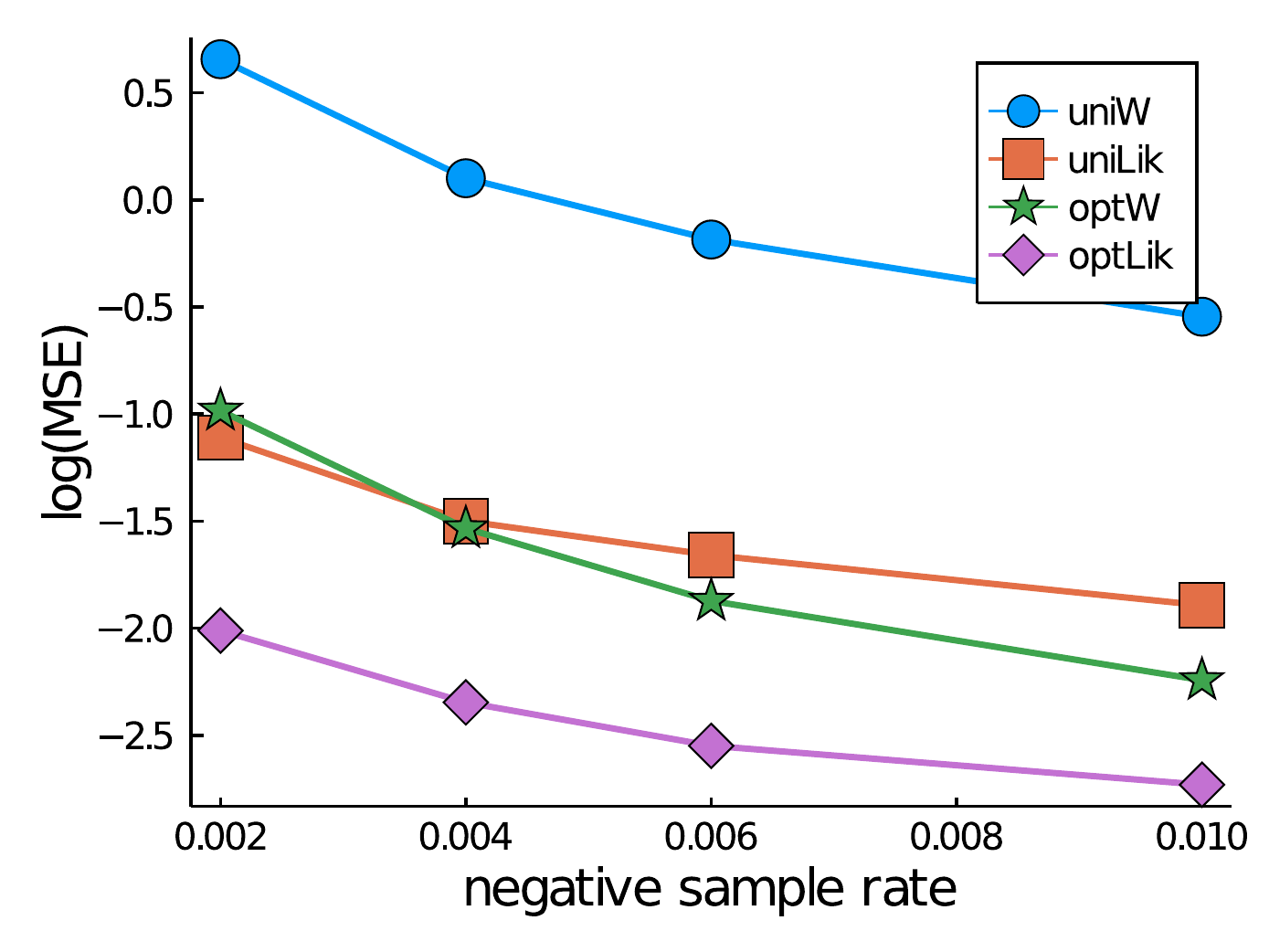}
    (c) {$\x$'s are $t_3$}
  \end{minipage}
  \begin{minipage}{0.23\textwidth}\centering
    \includegraphics[width=\textwidth]{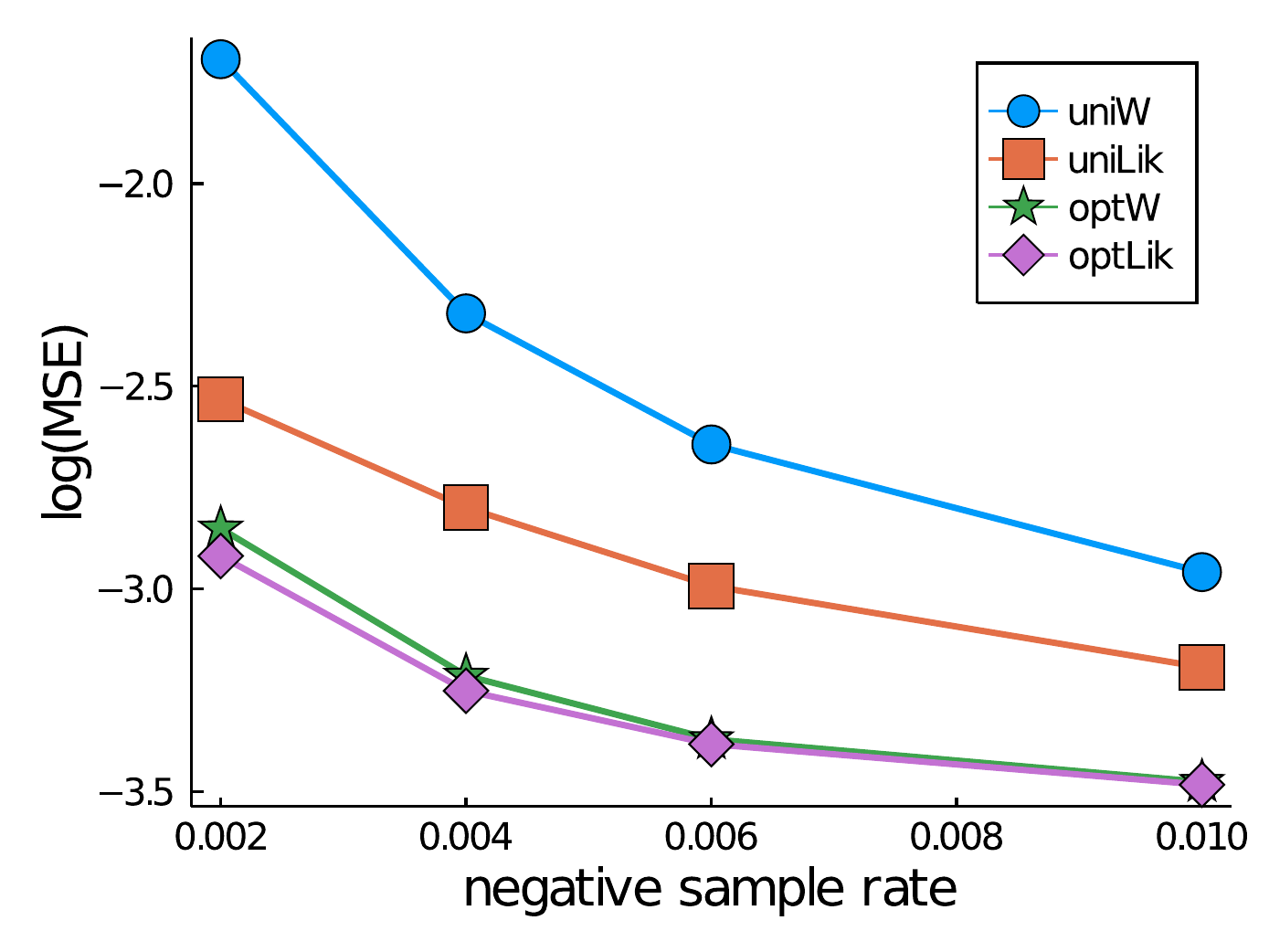}
    (d) {$\x$'s are exponential}
  \end{minipage}
  \caption{Degree of perturbation $\xi=0.5$. Log (MSEs) of subsample estimators for different sample sizes (the smaller the better).}
  \label{fig:supp_perturb_alpha_0_5}
\end{figure}

\begin{figure}[H]\centering
  \begin{minipage}{0.23\textwidth}\centering
    \includegraphics[width=\textwidth]{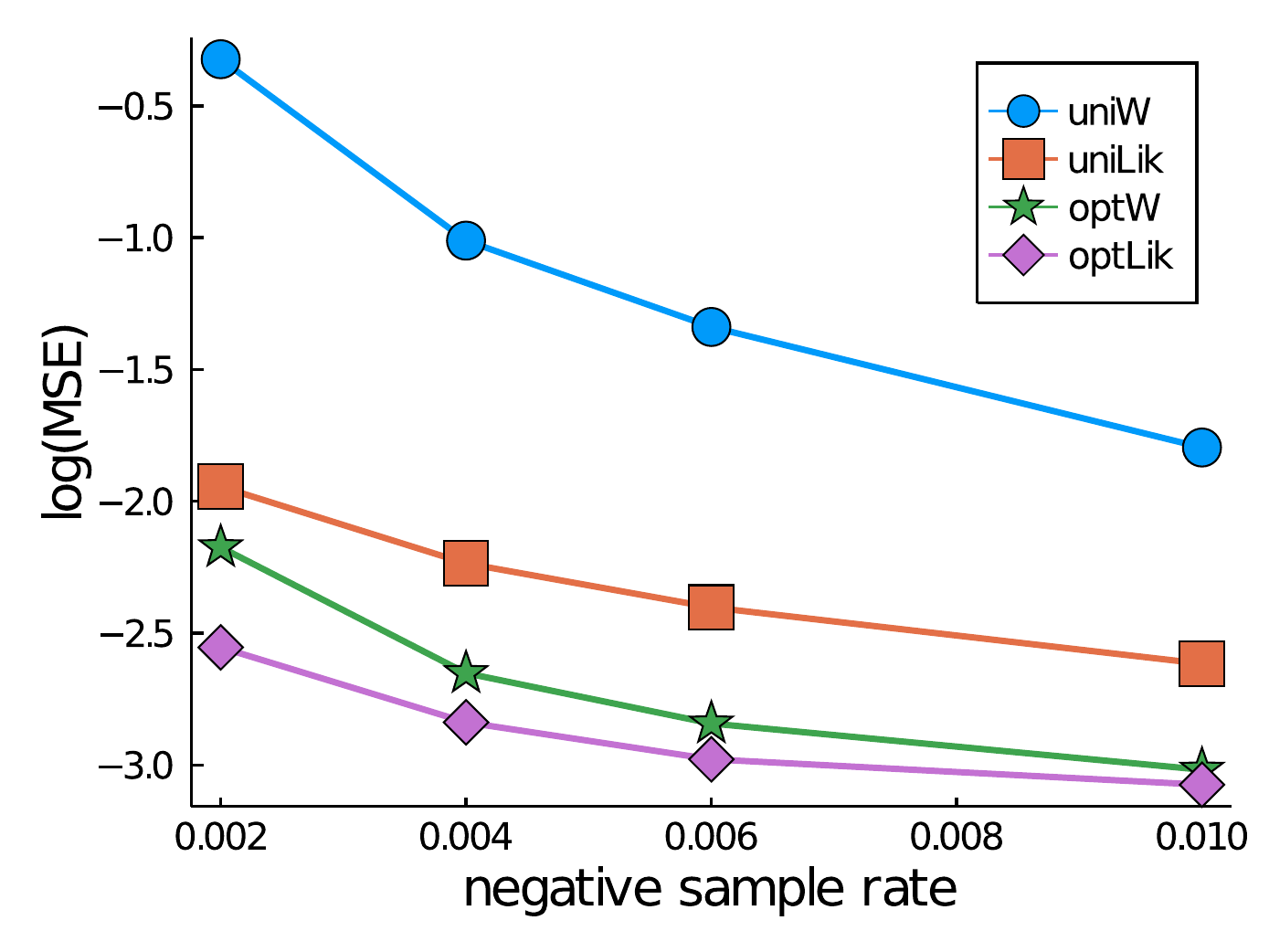}
    (a). {$\x$'s are normal}
  \end{minipage} %
  \begin{minipage}{0.23\textwidth}\centering
    \includegraphics[width=\textwidth]{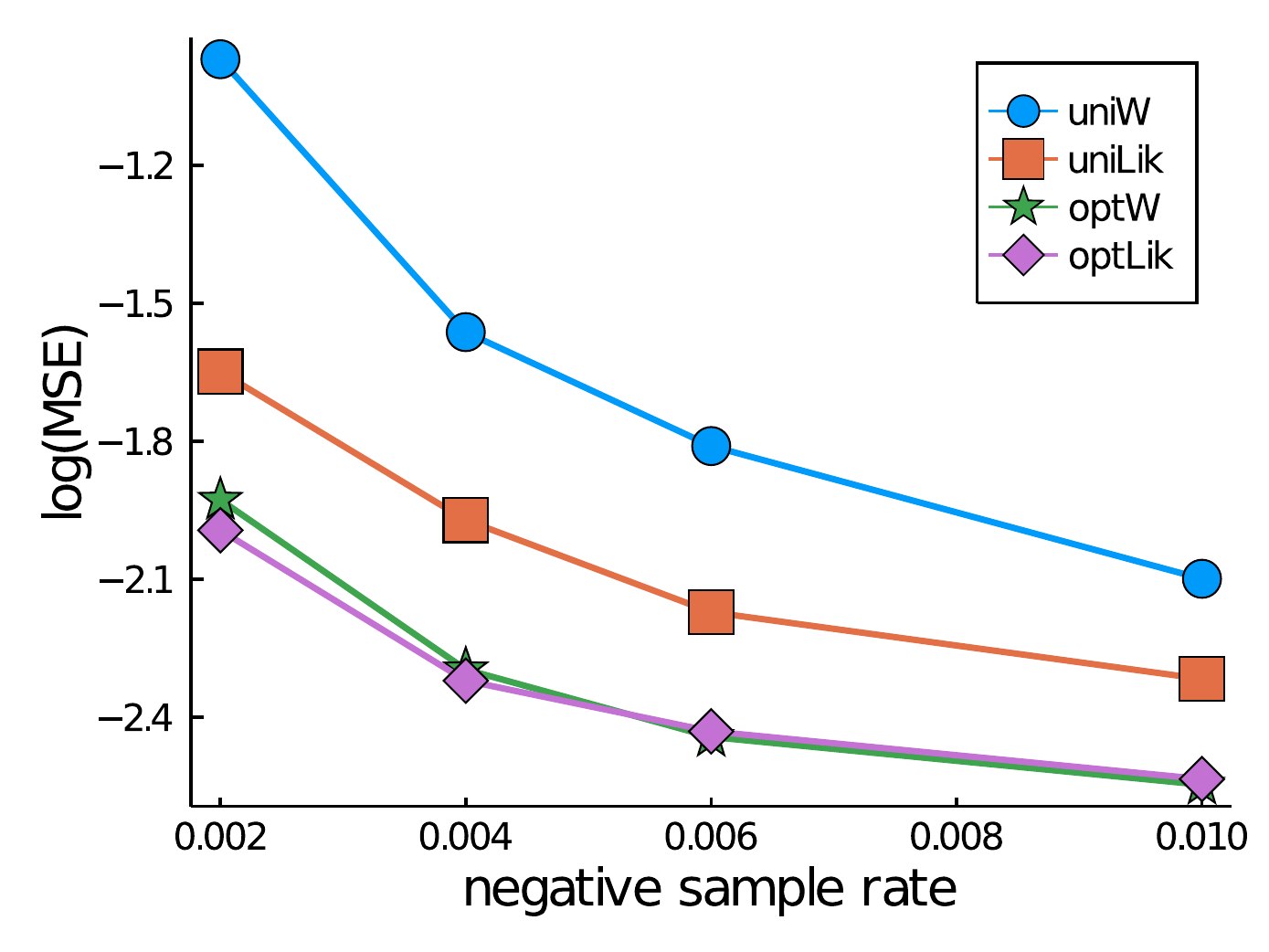}
    (b) {$\x$'s are lognormal}
  \end{minipage} %
  \begin{minipage}{0.23\textwidth}\centering
    \includegraphics[width=\textwidth]{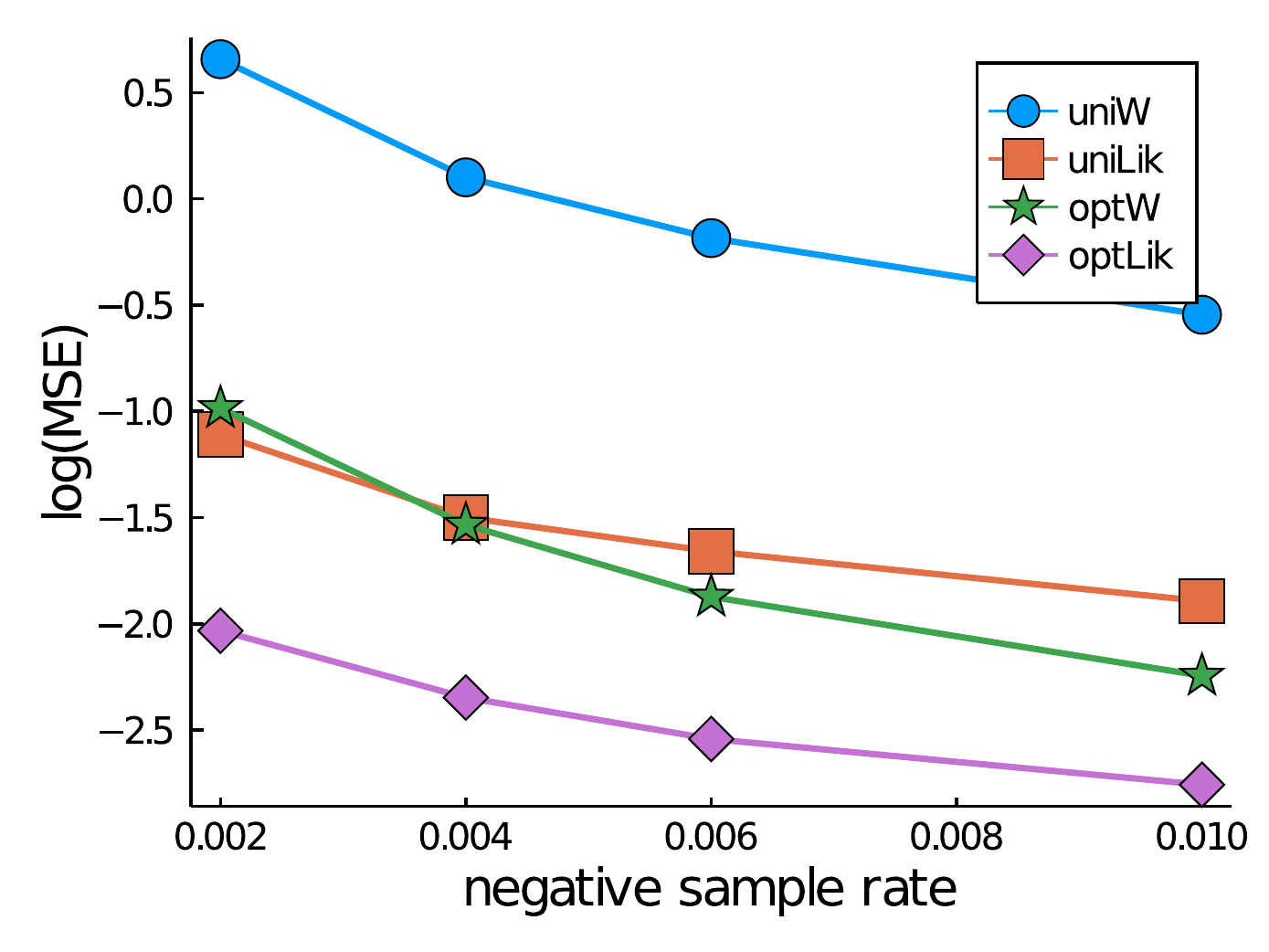}
    (c) {$\x$'s are $t_3$}
  \end{minipage}
  \begin{minipage}{0.23\textwidth}\centering
    \includegraphics[width=\textwidth]{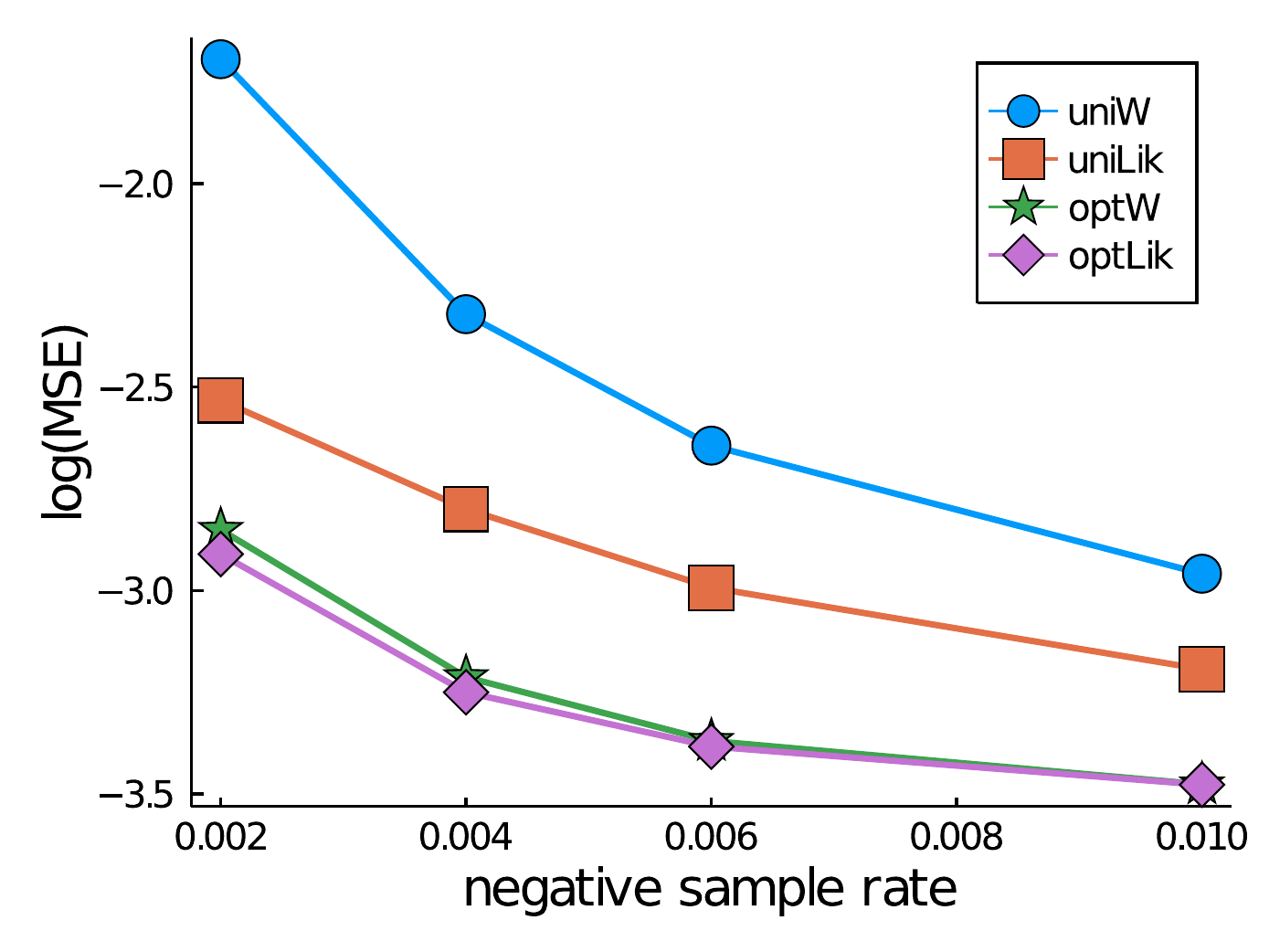}
    (d) {$\x$'s are exponential}
  \end{minipage}
  \caption{Degree of perturbation $\xi=1.0$. Log (MSEs) of subsample estimators for different sample sizes (the smaller the better).}
  \label{fig:supp_perturb_alpha_1}
\end{figure}

\subsection{Perturb the pilot weights $\beta$.}
\begin{figure}[H]\centering
  \begin{minipage}{0.23\textwidth}\centering
    \includegraphics[width=\textwidth]{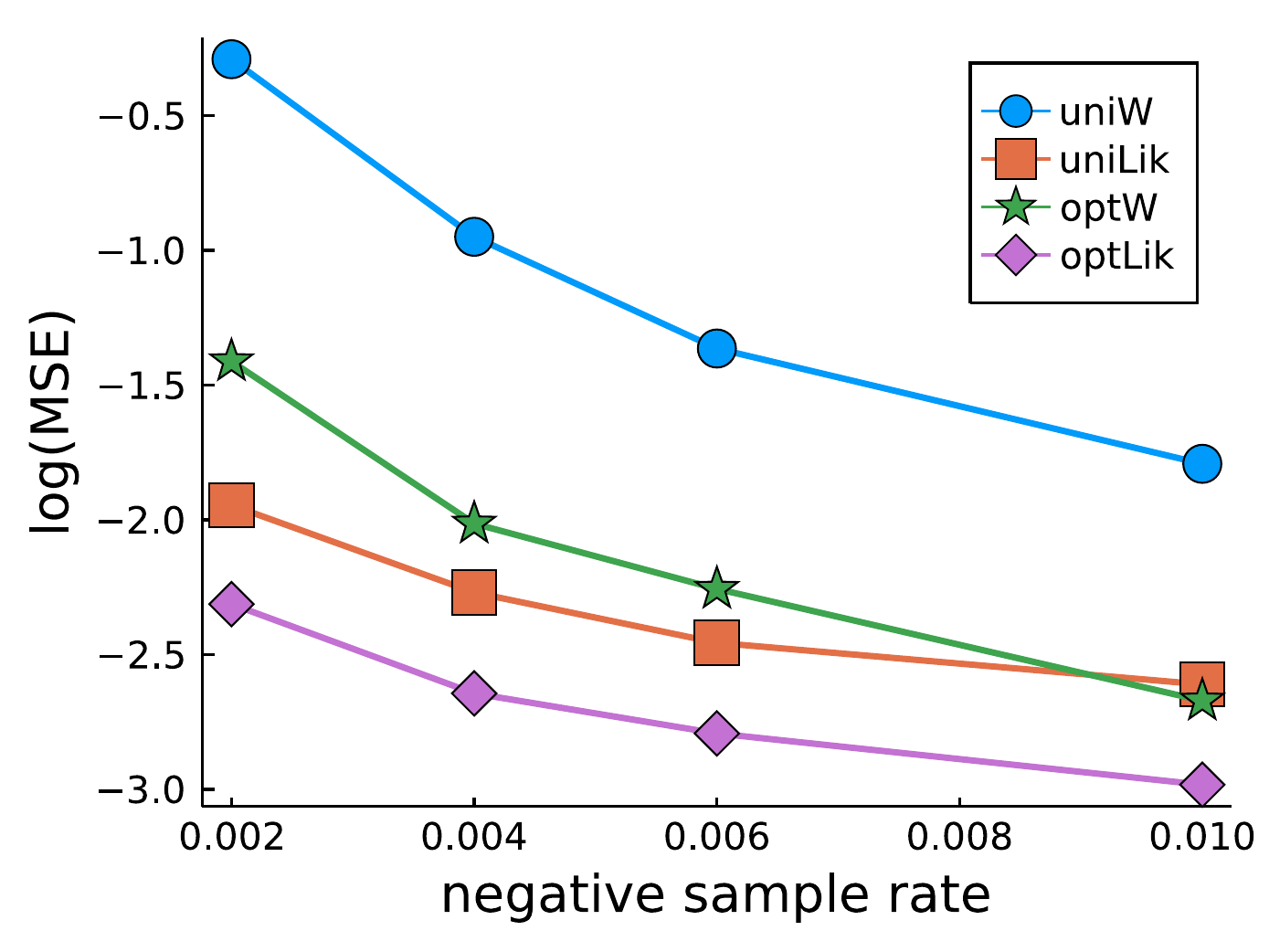}
    (a). {$\x$'s are normal}
  \end{minipage} %
  \begin{minipage}{0.23\textwidth}\centering
    \includegraphics[width=\textwidth]{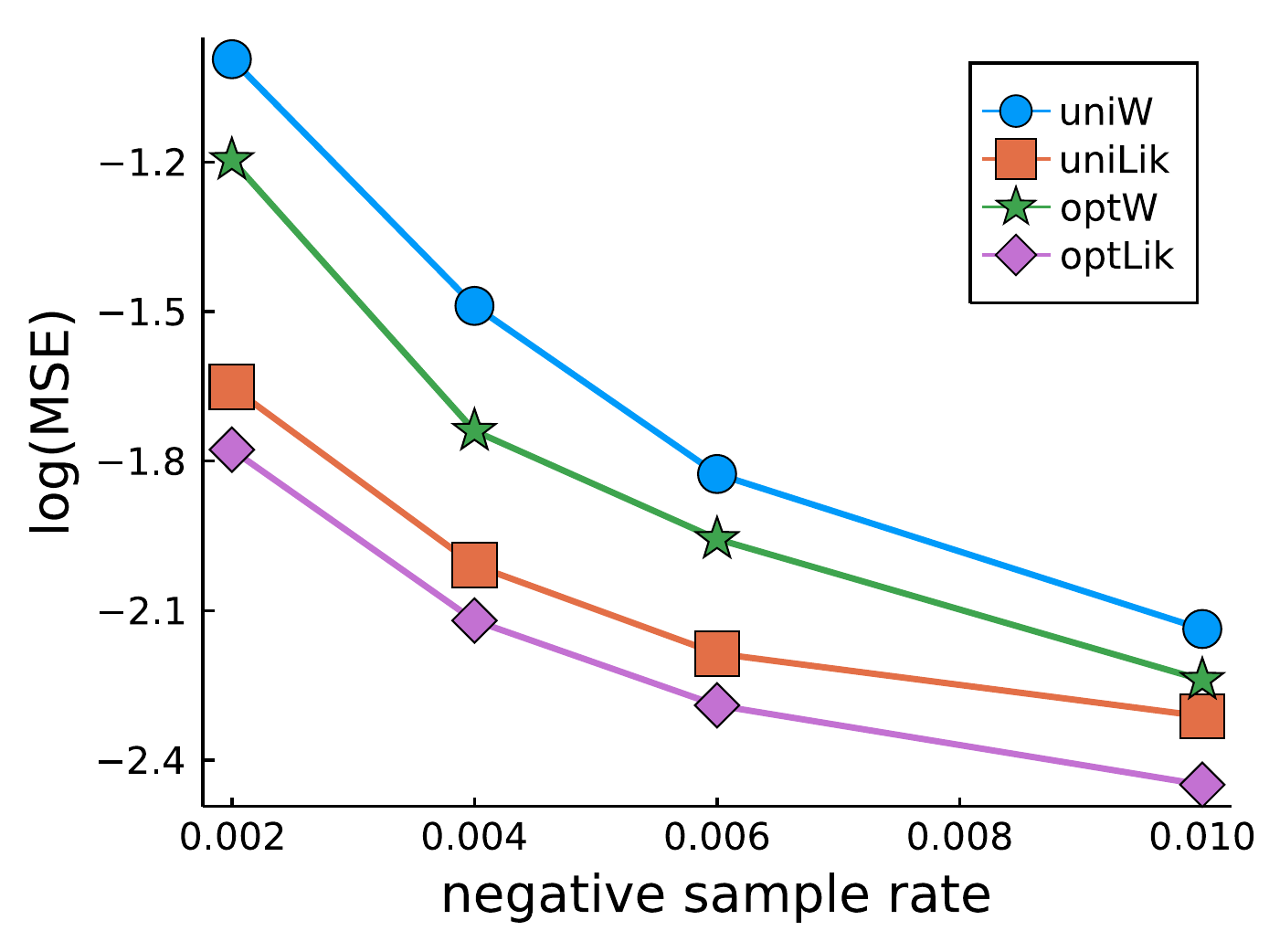}
    (b) {$\x$'s are lognormal}
  \end{minipage} %
  \begin{minipage}{0.23\textwidth}\centering
    \includegraphics[width=\textwidth]{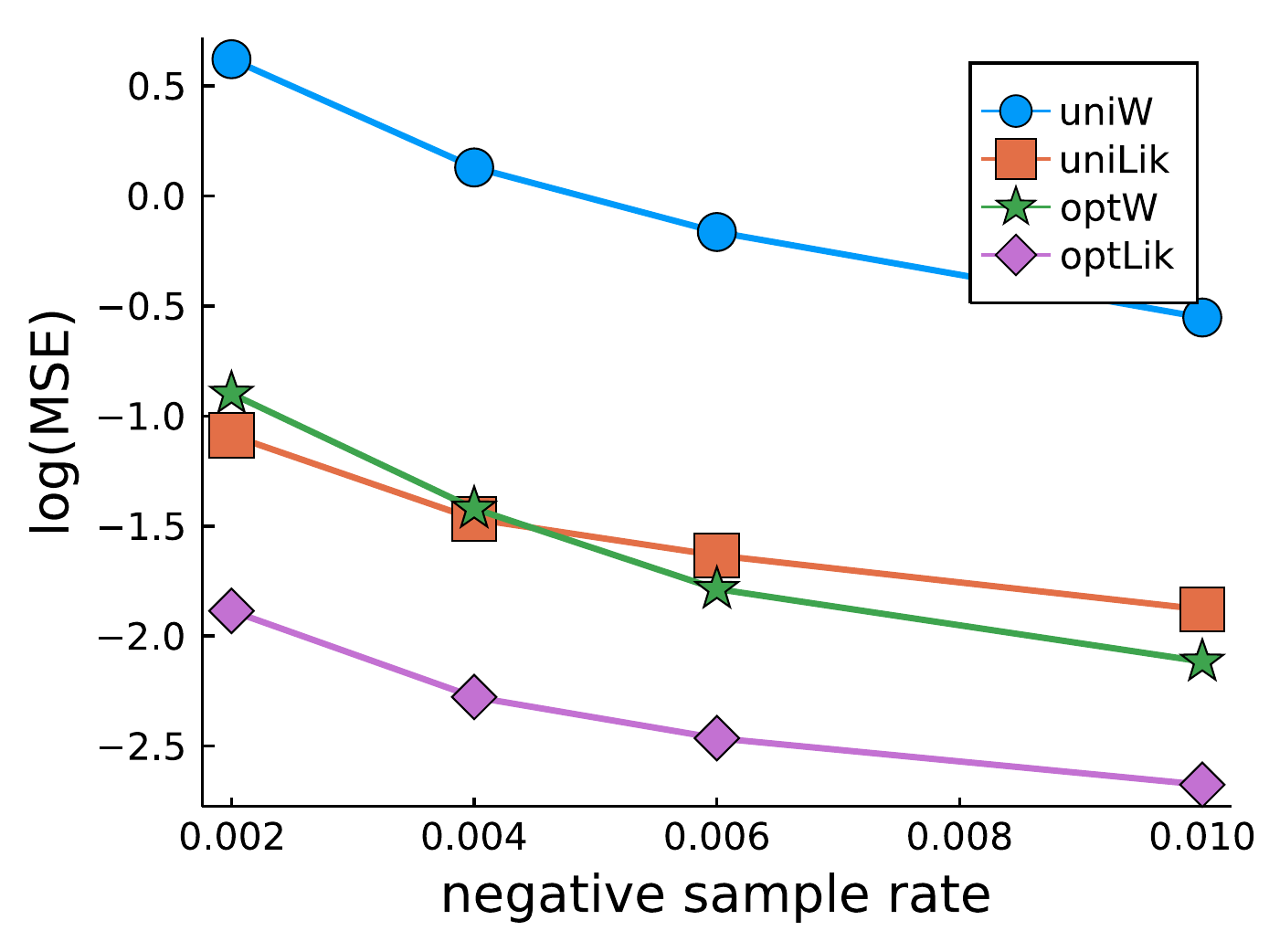}
    (c) {$\x$'s are $t_3$}
  \end{minipage}
  \begin{minipage}{0.23\textwidth}\centering
    \includegraphics[width=\textwidth]{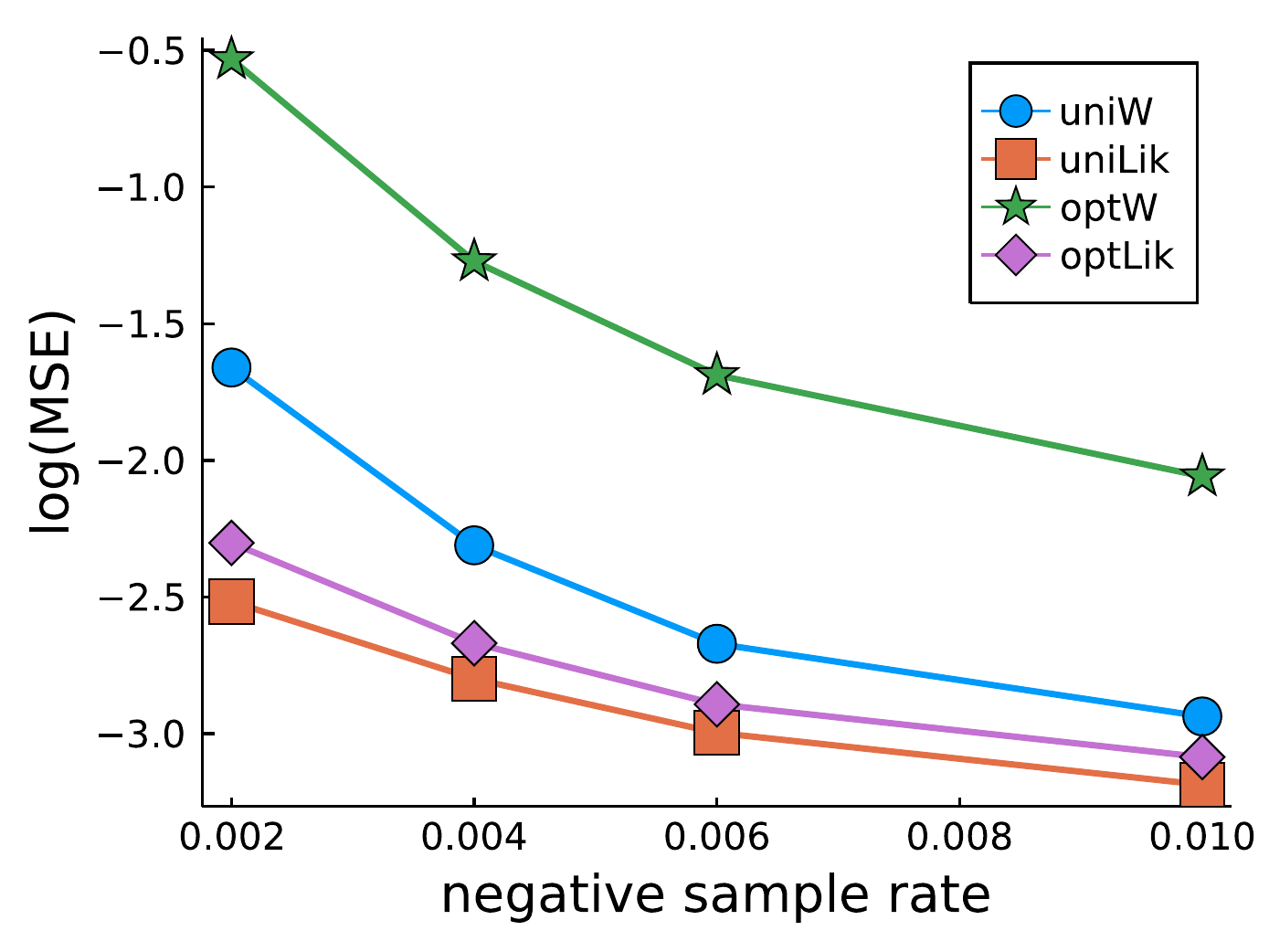}
    (d) {$\x$'s are exponential}
  \end{minipage}
  \caption{Degree of perturbation $\xi=1.0$. Log (MSEs) of subsample estimators for different sample sizes (the smaller the better).}
  \label{fig:supp_perturb_beta_1}
\end{figure}

We perturb $\tilde\bbeta\leftarrow \tilde\bbeta + \xi\times \mathcal{N}(0,I)$, where $\xi$ is a hyperparameter to control the degree of perturbation. We try $\xi=\{0.1, 0.5, 1.0\}$. As the result shows, the IPW estimator is pretty sensitive to the perturbation, and optW can be even worse than uniLik when the perturbation is significant (Figure~\ref{fig:supp_perturb_beta_1}). uniLik may exceed optLik when the perturbation is too large such that the optimal sampling probability is significatly mis-calculated (Figure~\ref{fig:supp_perturb_beta_1}(d)).

\begin{figure}[H]\centering
  \begin{minipage}{0.23\textwidth}\centering
    \includegraphics[width=\textwidth]{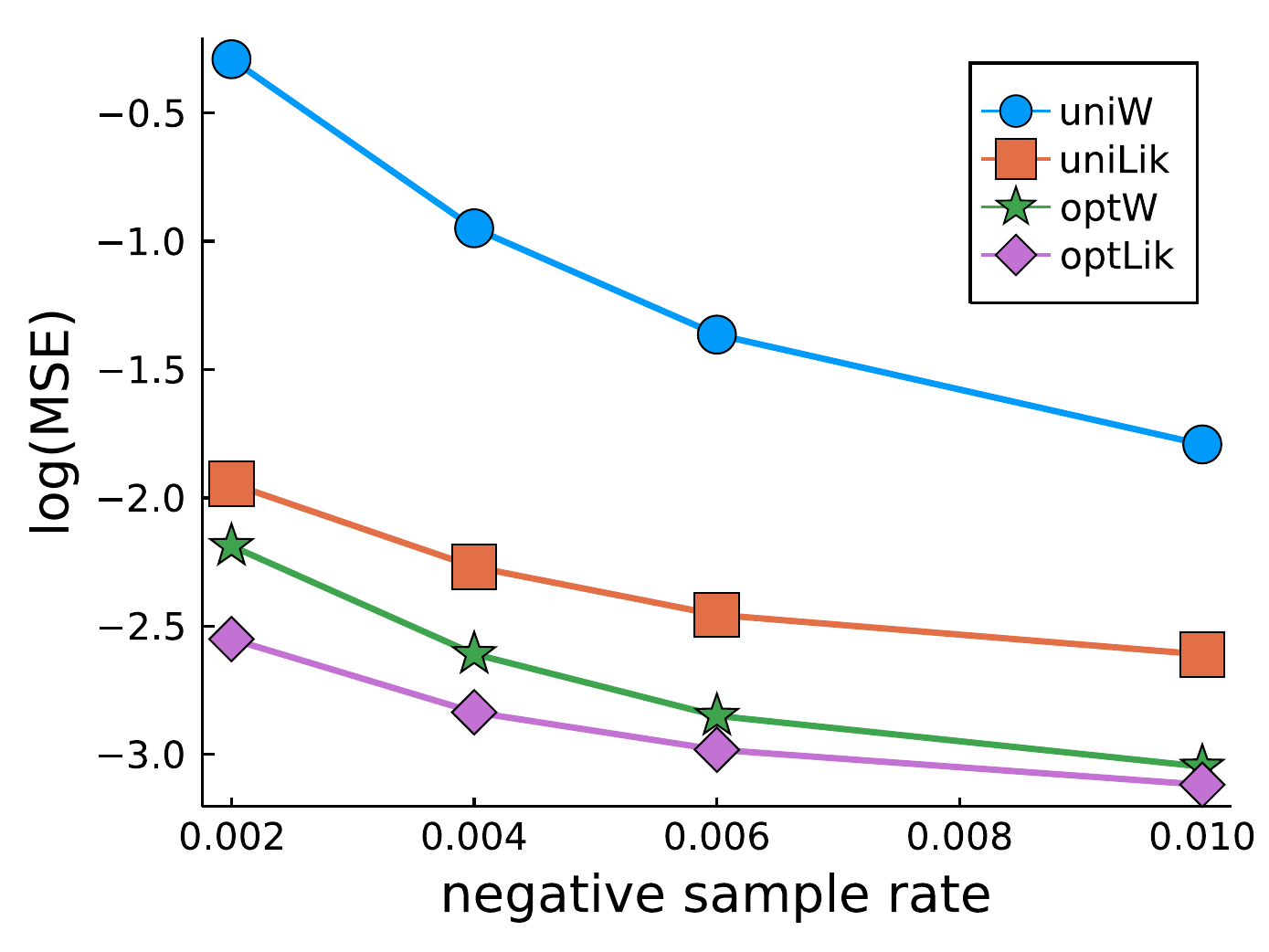}
    (a). {$\x$'s are normal}
  \end{minipage} %
  \begin{minipage}{0.23\textwidth}\centering
    \includegraphics[width=\textwidth]{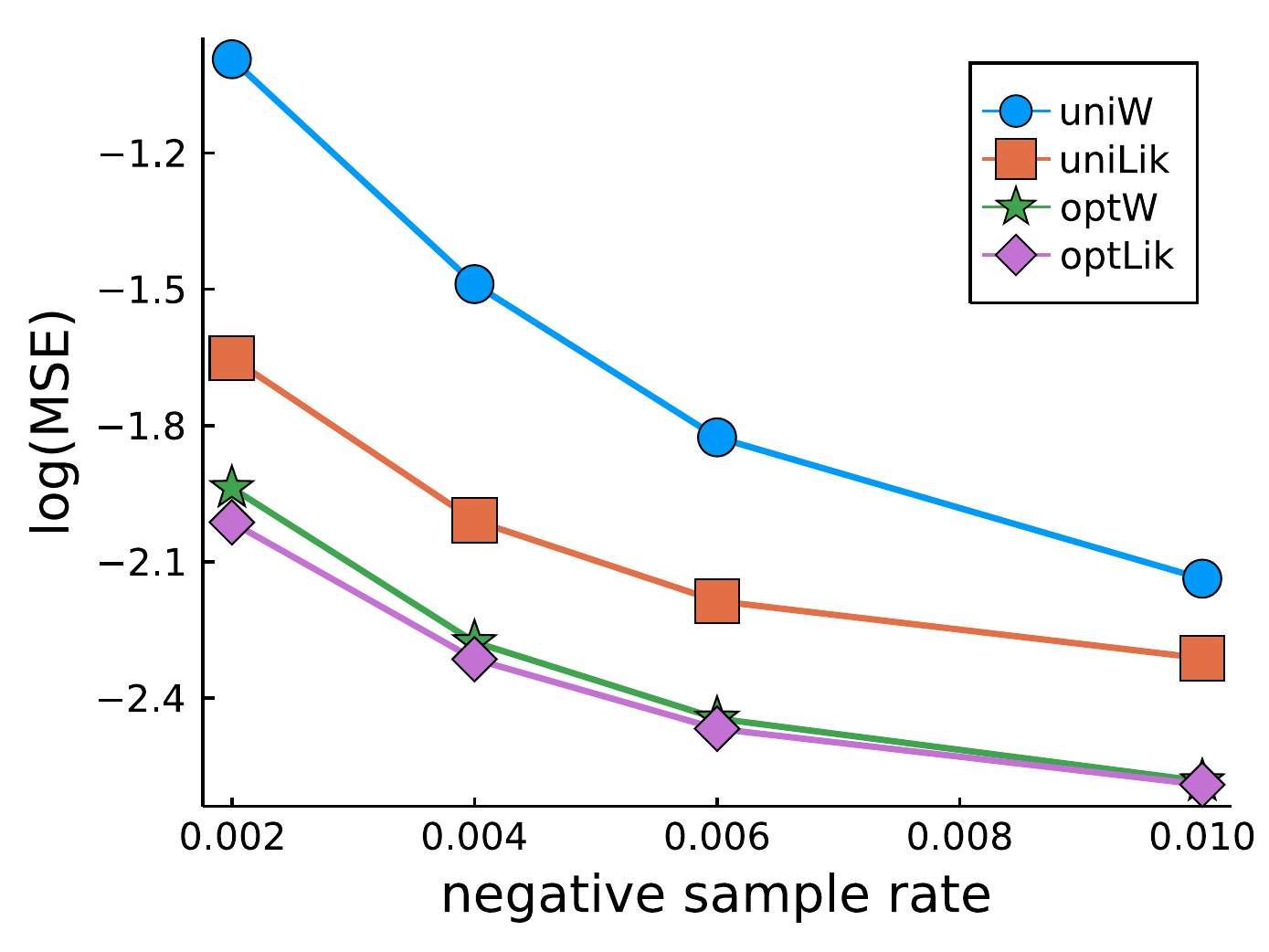}
    (b) {$\x$'s are lognormal}
  \end{minipage} %
  \begin{minipage}{0.23\textwidth}\centering
    \includegraphics[width=\textwidth]{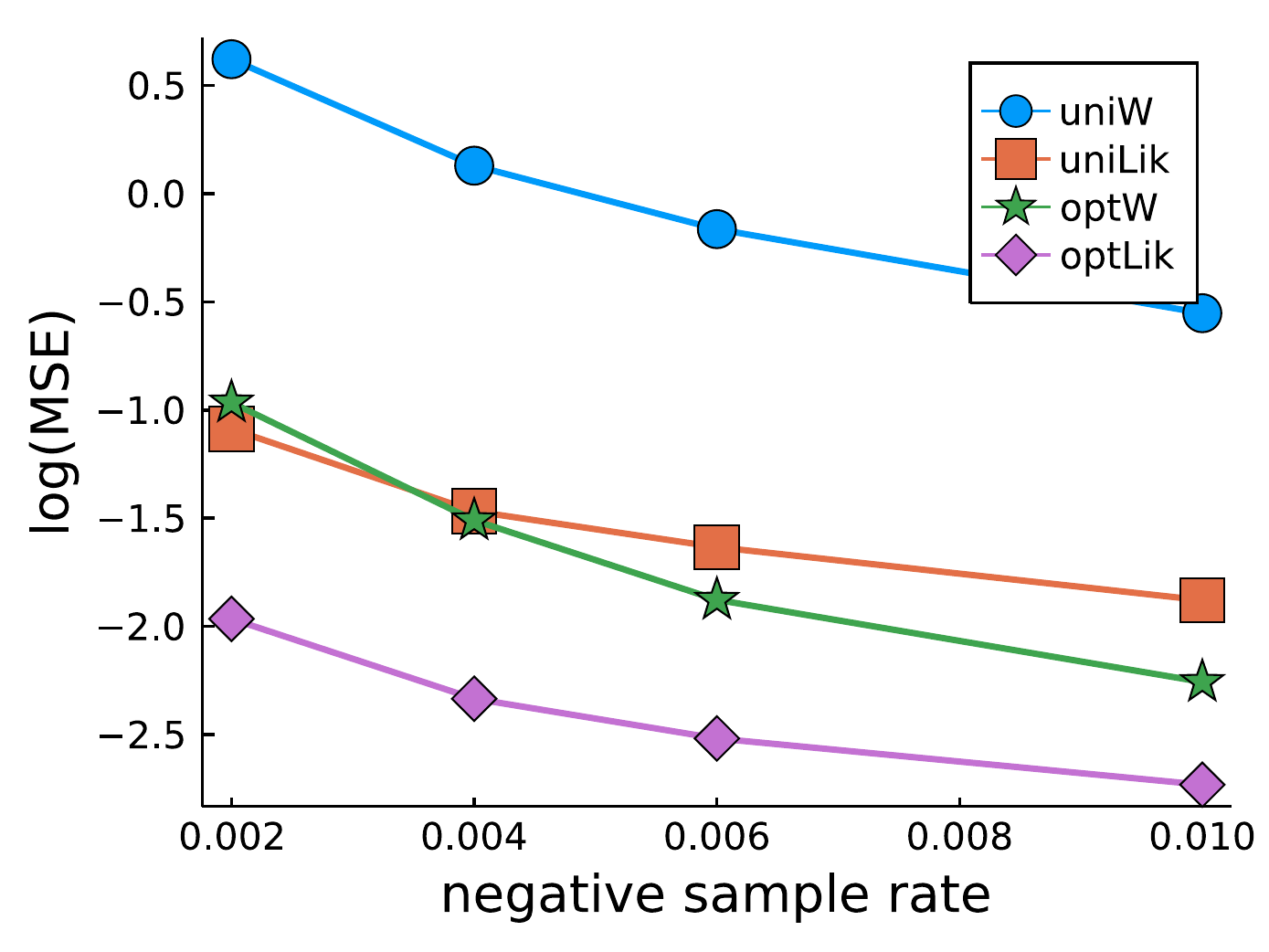}
    (c) {$\x$'s are $t_3$}
  \end{minipage}
  \begin{minipage}{0.23\textwidth}\centering
    \includegraphics[width=\textwidth]{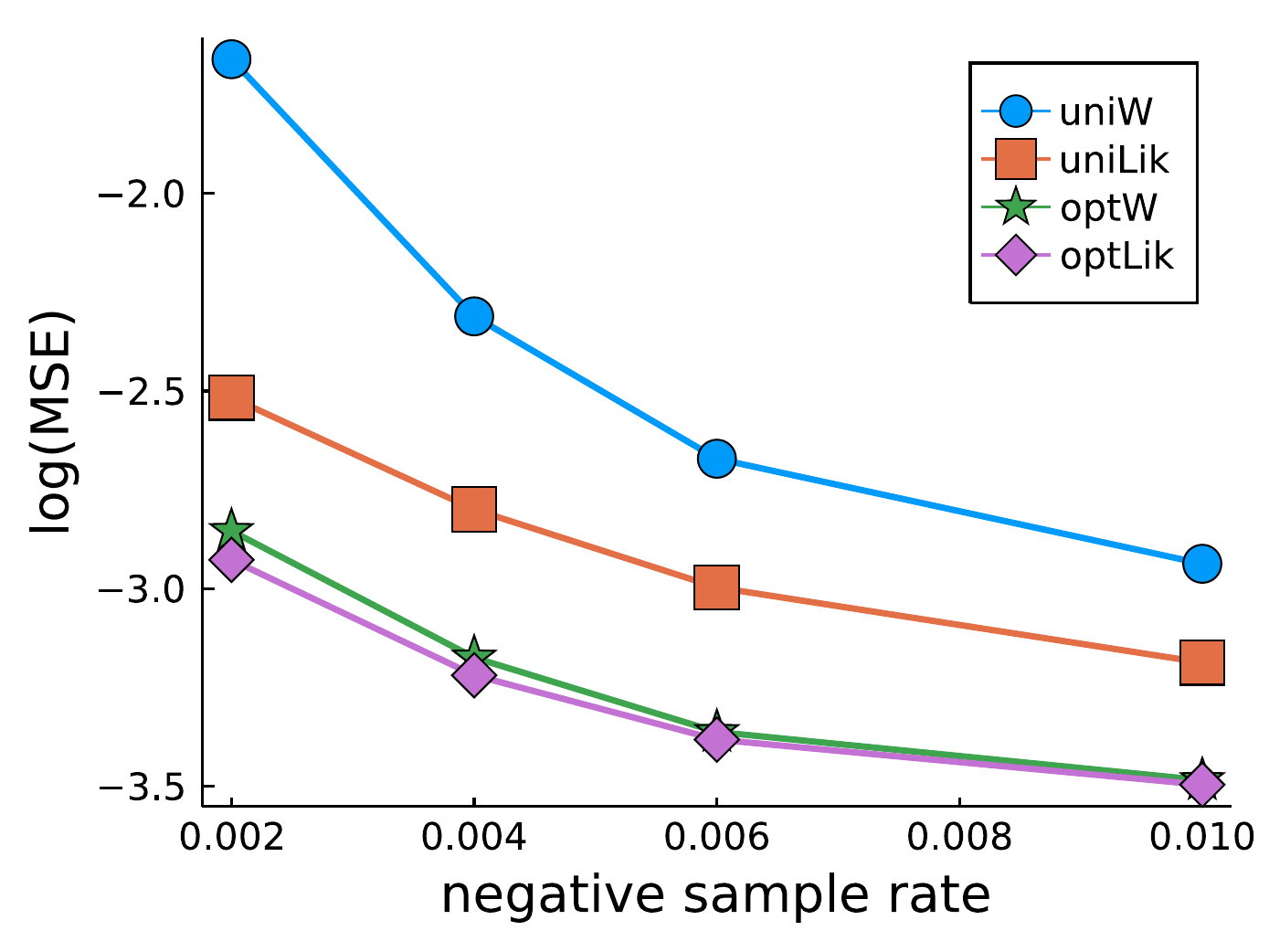}
    (d) {$\x$'s are exponential}
  \end{minipage}
  \caption{Degree of perturbation $\xi=0.1$. Log (MSEs) of subsample estimators for different sample sizes (the smaller the better).}
  \label{fig:supp_perturb_beta_0_1}
\end{figure}

\begin{figure}[H]\centering
  \begin{minipage}{0.23\textwidth}\centering
    \includegraphics[width=\textwidth]{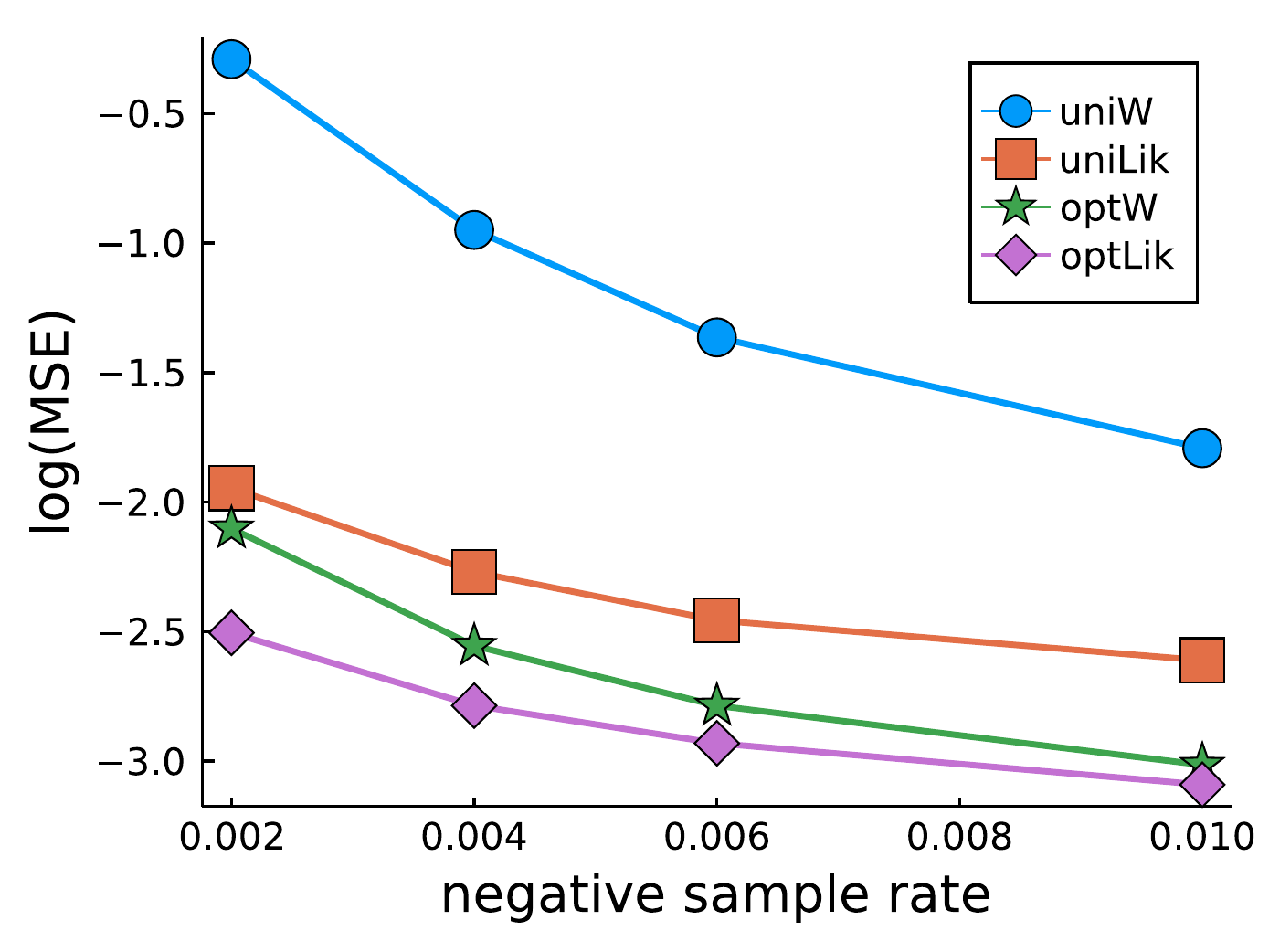}
    (a). {$\x$'s are normal}
  \end{minipage} %
  \begin{minipage}{0.23\textwidth}\centering
    \includegraphics[width=\textwidth]{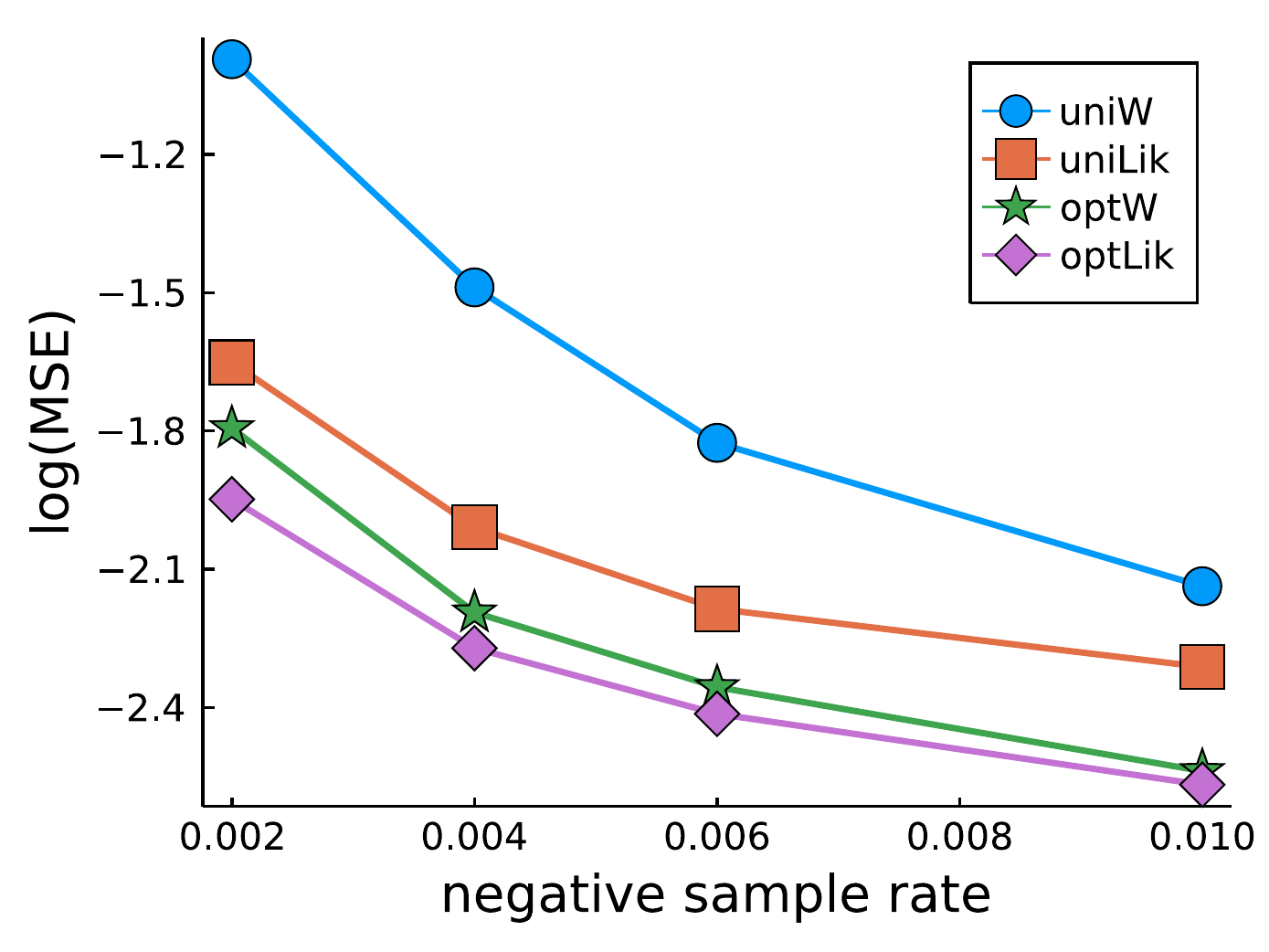}
    (b) {$\x$'s are lognormal}
  \end{minipage} %
  \begin{minipage}{0.23\textwidth}\centering
    \includegraphics[width=\textwidth]{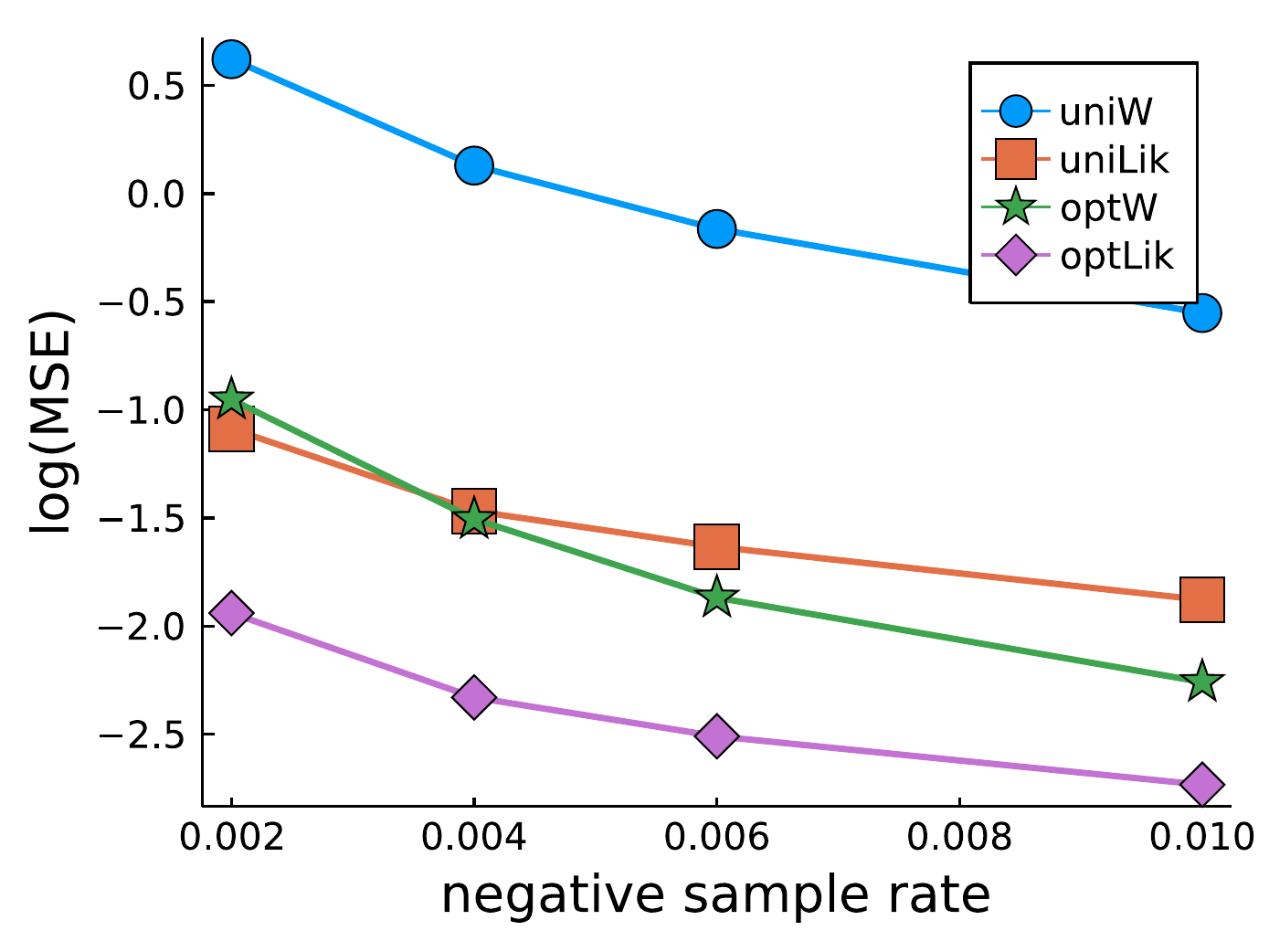}
    (c) {$\x$'s are $t_3$}
  \end{minipage}
  \begin{minipage}{0.23\textwidth}\centering
    \includegraphics[width=\textwidth]{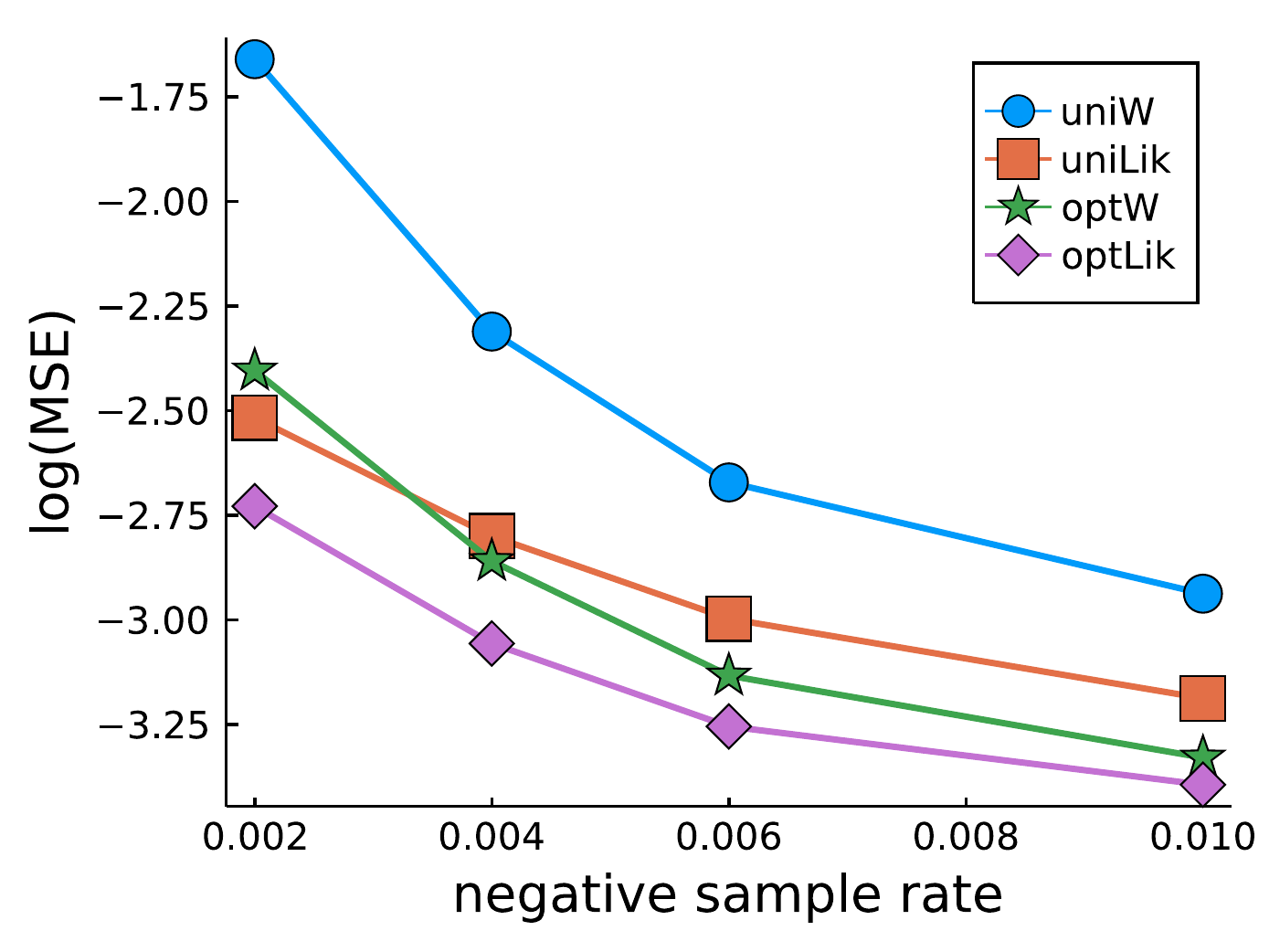}
    (d) {$\x$'s are exponential}
  \end{minipage}
  \caption{Degree of perturbation $\xi=0.5$. Log (MSEs) of subsample estimators for different sample sizes (the smaller the better).}
  \label{fig:supp_perturb_beta_0_5}
\end{figure}

\section{Experiments on Model Mis-specification}
\begin{figure}[H]
    \centering
    \includegraphics[width=.4\textwidth]{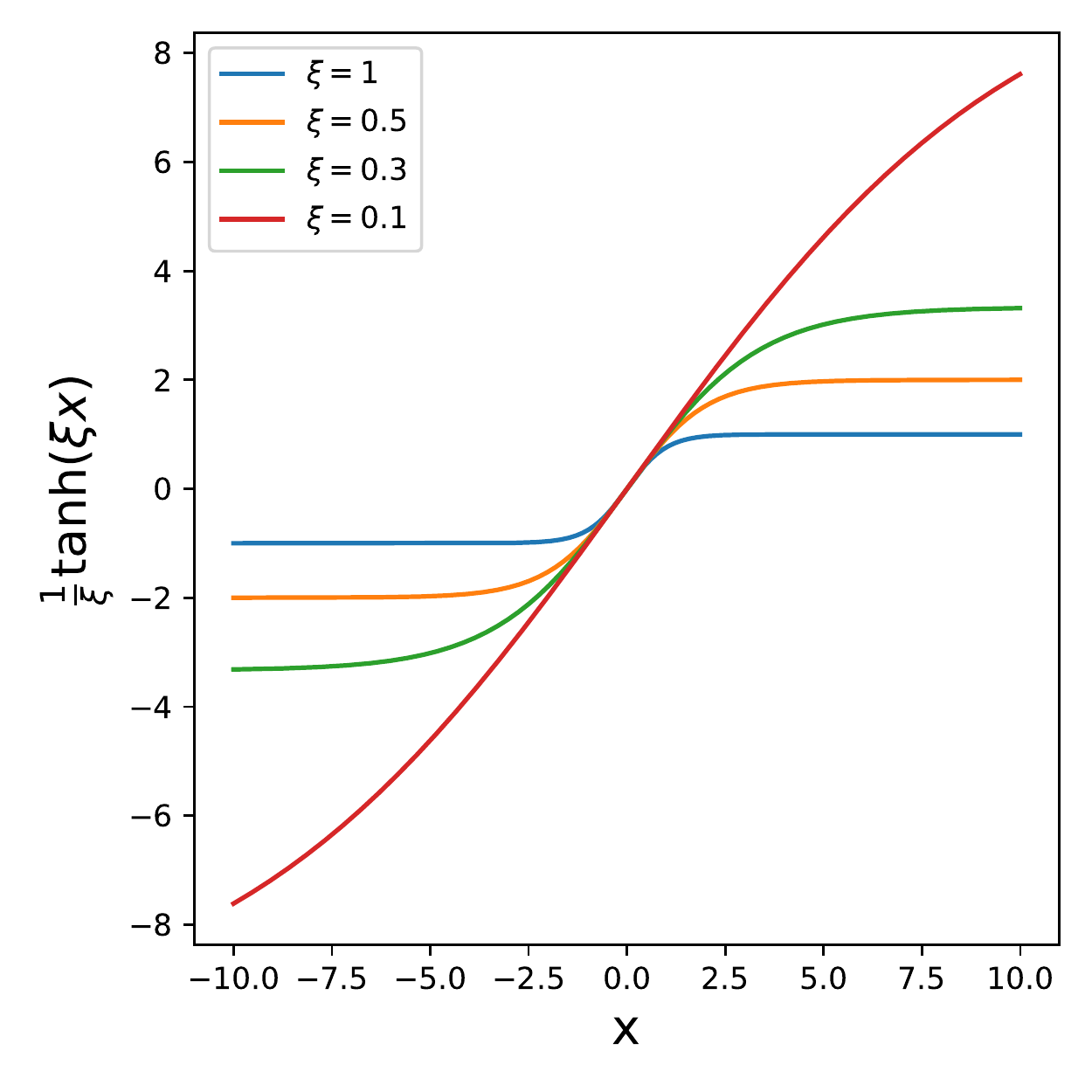}
    \caption{Scaled $\tanh$ function approximates identity transformation when $\xi\rightarrow 0$.}
    \label{fig:supp_scaled_h}
\end{figure}

This section studies the case when model is mis-specified. That is, the ground truth model fall out of the domain of linear logistic regression model. We design the ground truth model as a two-layer neural network with non-linear activations:
\begin{align}
g(\x;\btheta)=\alpha+\frac{1}{\xi}\tanh(\xi\cdot\x\tp\mathbf{W})\bbeta,
\end{align}
where $\xi$ is a constant to control the ``cut-off" threshold of the non-linear activation $\tanh$. Note that as $\xi\rightarrow 0$, the scaled $\tanh$ function $\frac{1}{\xi}\cdot\tanh(\xi\cdot x)\approx x$ as shown in Figure~\ref{fig:supp_scaled_h}. Thus, by tuning $\xi$ we have the control of the deviation of scaled $\tanh$ function from a linear, identity transformation.

We also generate $\mathbf{W}$ to be an approximately identity matrix, by setting the diagonal elements of $\mathbf{W}$ to be 1, and draw off diagonal terms to be i.i.d. Normal distribution scaled by a factor $\xi_w$. Clearly, the operator norm $\|\mathbf{W}-I\|_{2\rightarrow 2}$ is controlled by $\xi_w$. When $\xi_w=0$, $\mathbf{W}=I$.

When perturbing the ground-truth model by setting $\xi$ and $\xi_w$ (thus $\mathbf{W}$), we need to reset the intercept $\alpha$ to match the pos/neg proportion to be $1/400$. For each experiment setup, we record the tuned $\alpha$ in the caption for reproducible research.

\subsection{Ablation study.}
\begin{figure}[H]\centering
  \begin{minipage}{0.23\textwidth}\centering
    \includegraphics[width=\textwidth]{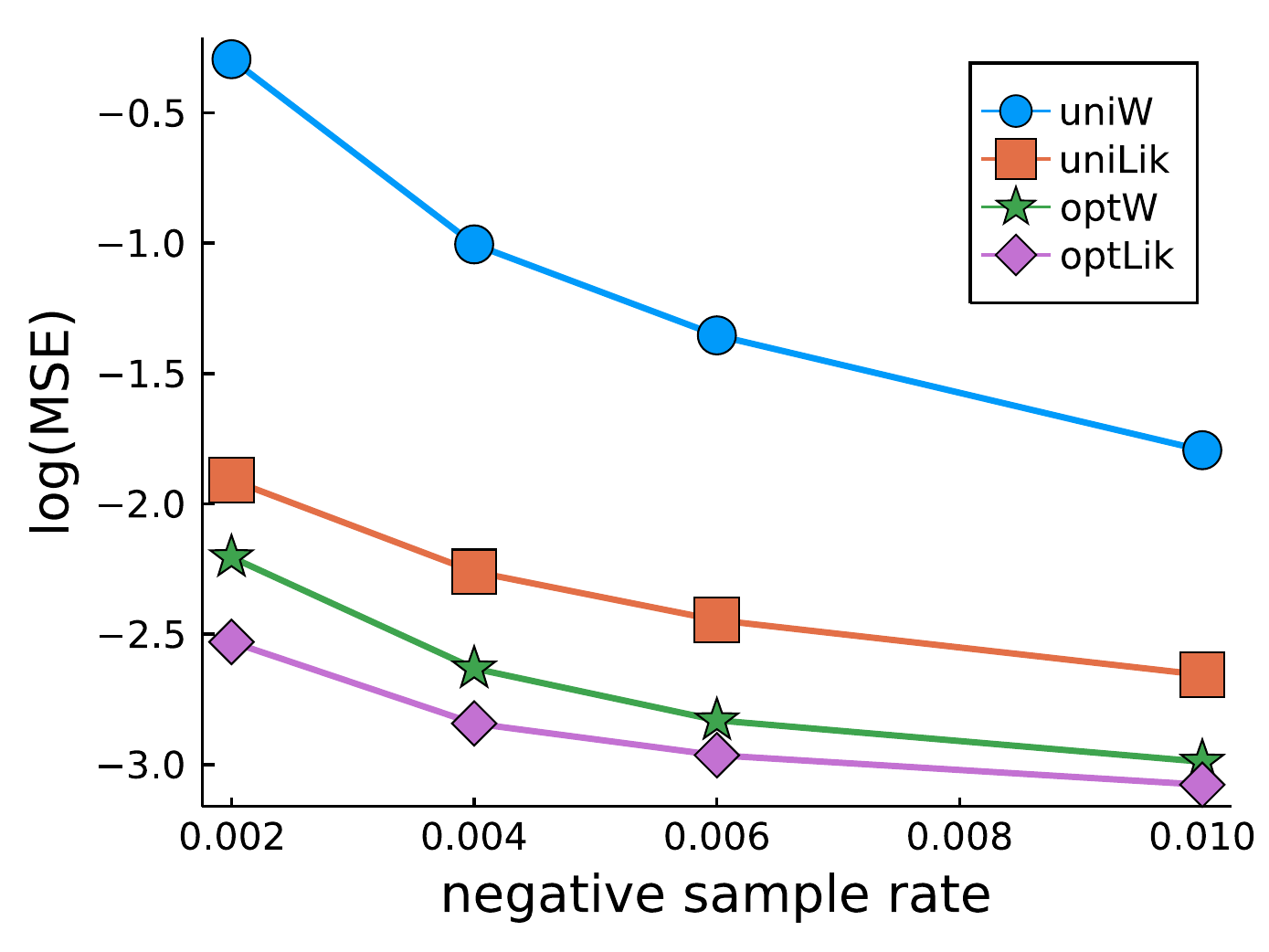}
    (a). {$\x$'s are normal}
  \end{minipage} %
  \begin{minipage}{0.23\textwidth}\centering
    \includegraphics[width=\textwidth]{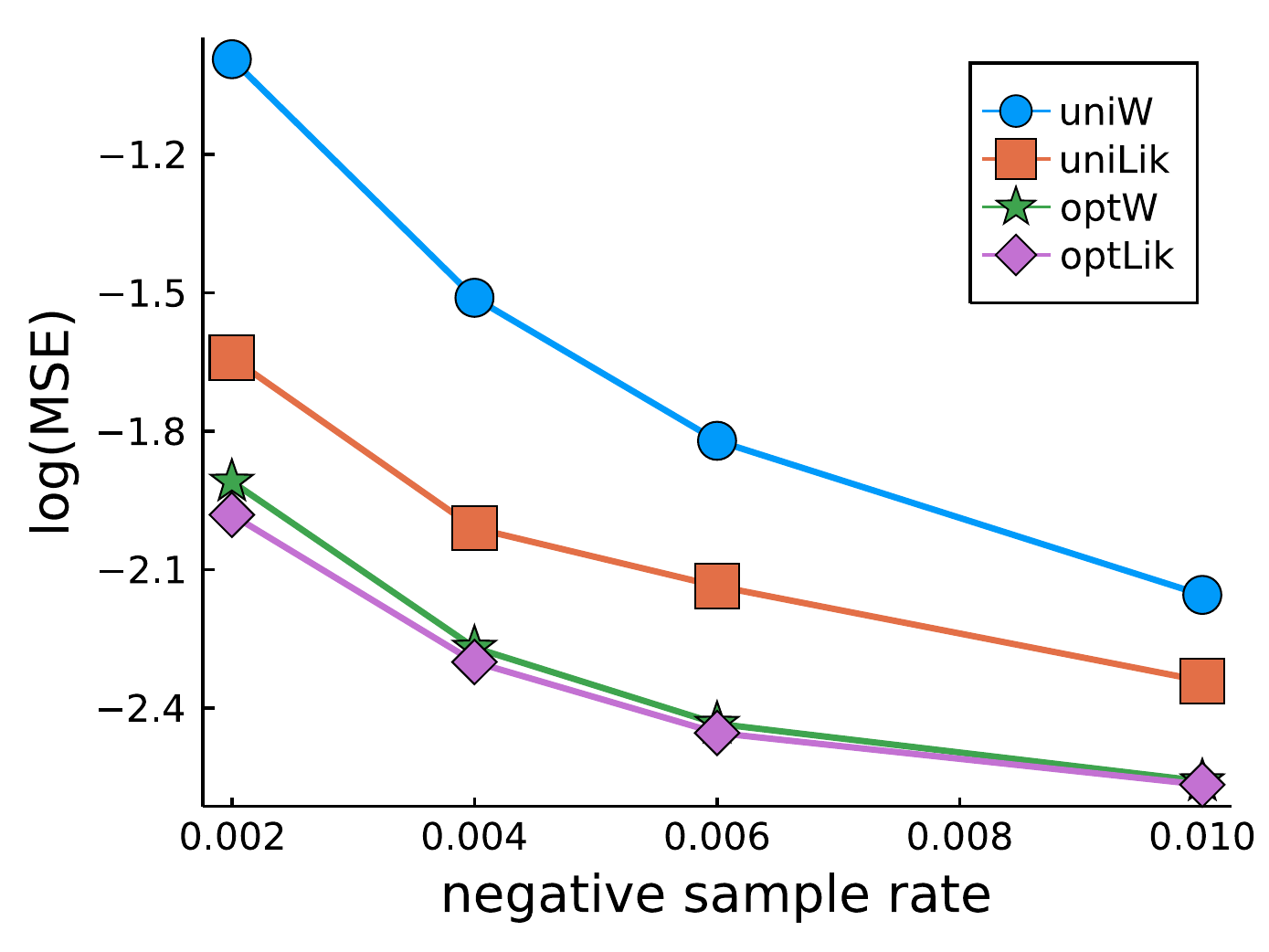}
    (b) {$\x$'s are lognormal}
  \end{minipage} %
  \begin{minipage}{0.23\textwidth}\centering
    \includegraphics[width=\textwidth]{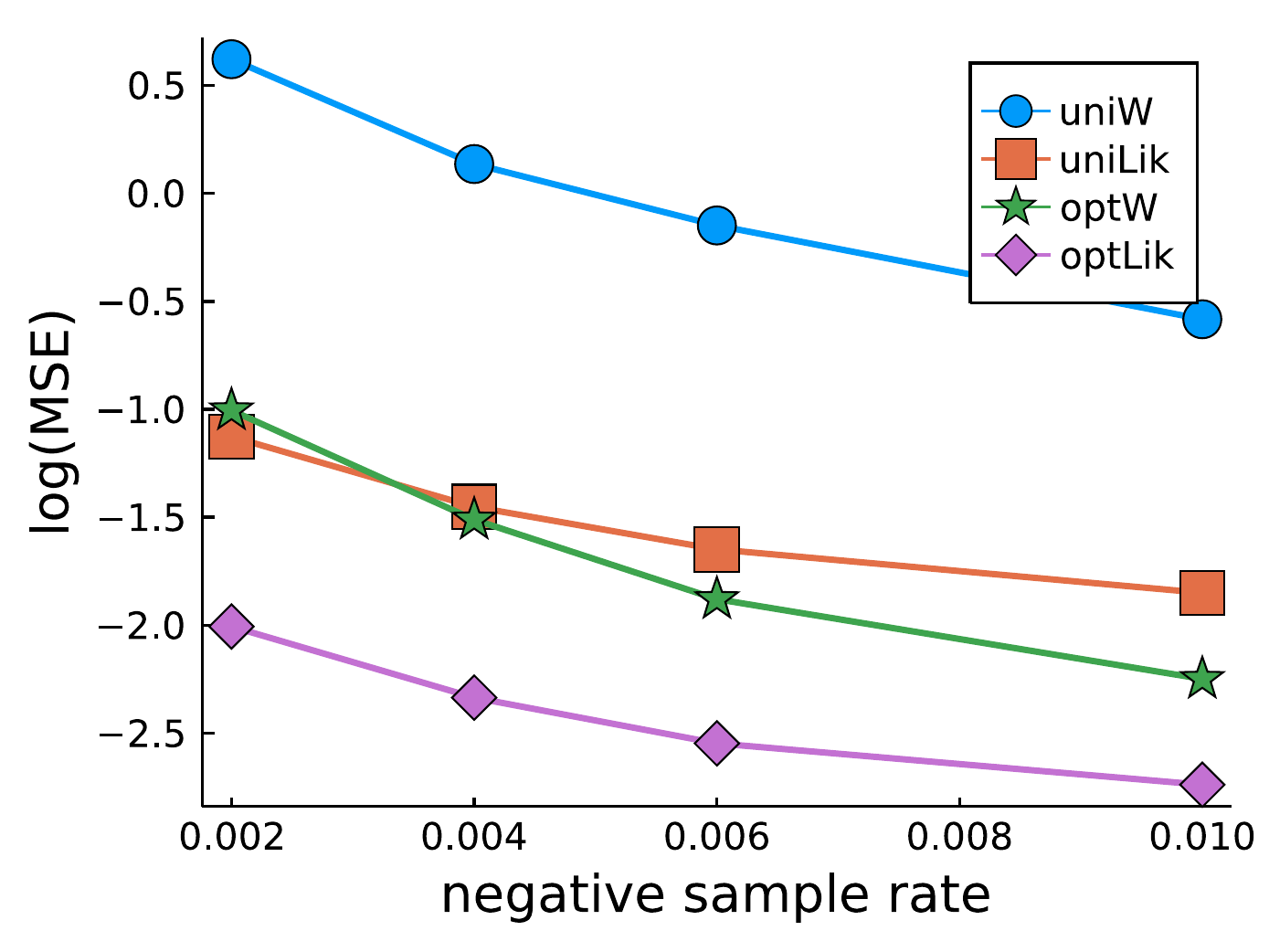}
    (c) {$\x$'s are $t_3$}
  \end{minipage}
  \begin{minipage}{0.23\textwidth}\centering
    \includegraphics[width=\textwidth]{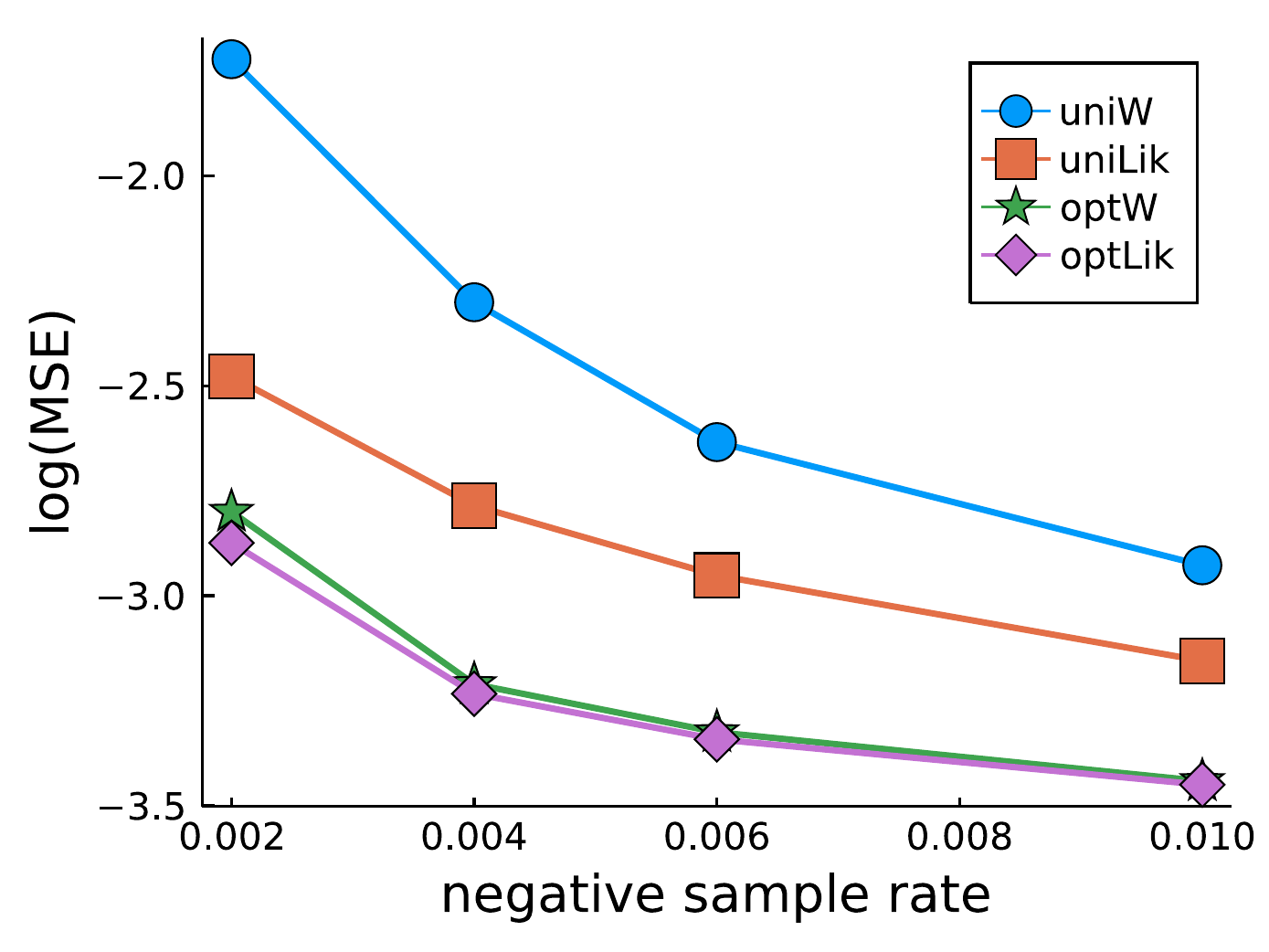}
    (d) {$\x$'s are exponential}
  \end{minipage}
  \caption{Degree of non-linearity $\xi=10^{-3},\xi_w=10^{-3}$. Ground-truth intercepts: (a) $\alpha=-7.65$, (b) $\alpha=-0.5$, (c) $\alpha=-7$, (d) $\alpha=-1.8$. Log (MSEs) of subsample estimators for different sample sizes (the smaller the better).}
  \label{fig:supp_model_mis_ablation}
\end{figure}

When $\xi$ and $\xi_w$ are extremely small, $g(\x;\btheta)\approx \alpha+\x\tp\bbeta$. Empirical results shown in Figure~\ref{fig:supp_model_mis_ablation} match results without model mis-specification.

\subsection{Tune the degree of activation function non-linearity by trying different $\xi$.}
\begin{figure}[H]\centering
  \begin{minipage}{0.23\textwidth}\centering
    \includegraphics[width=\textwidth]{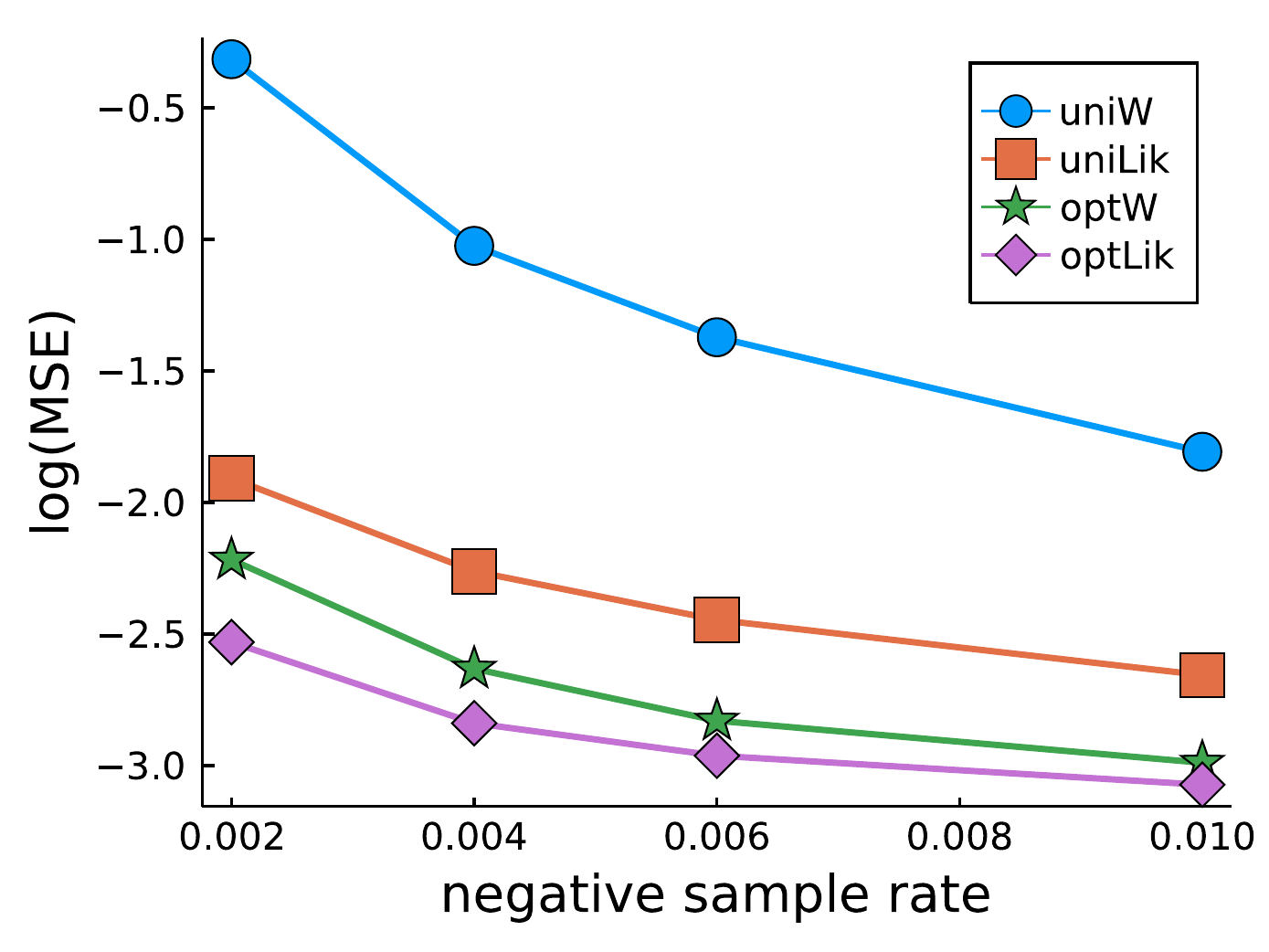}
    (a). {$\x$'s are normal}
  \end{minipage} %
  \begin{minipage}{0.23\textwidth}\centering
    \includegraphics[width=\textwidth]{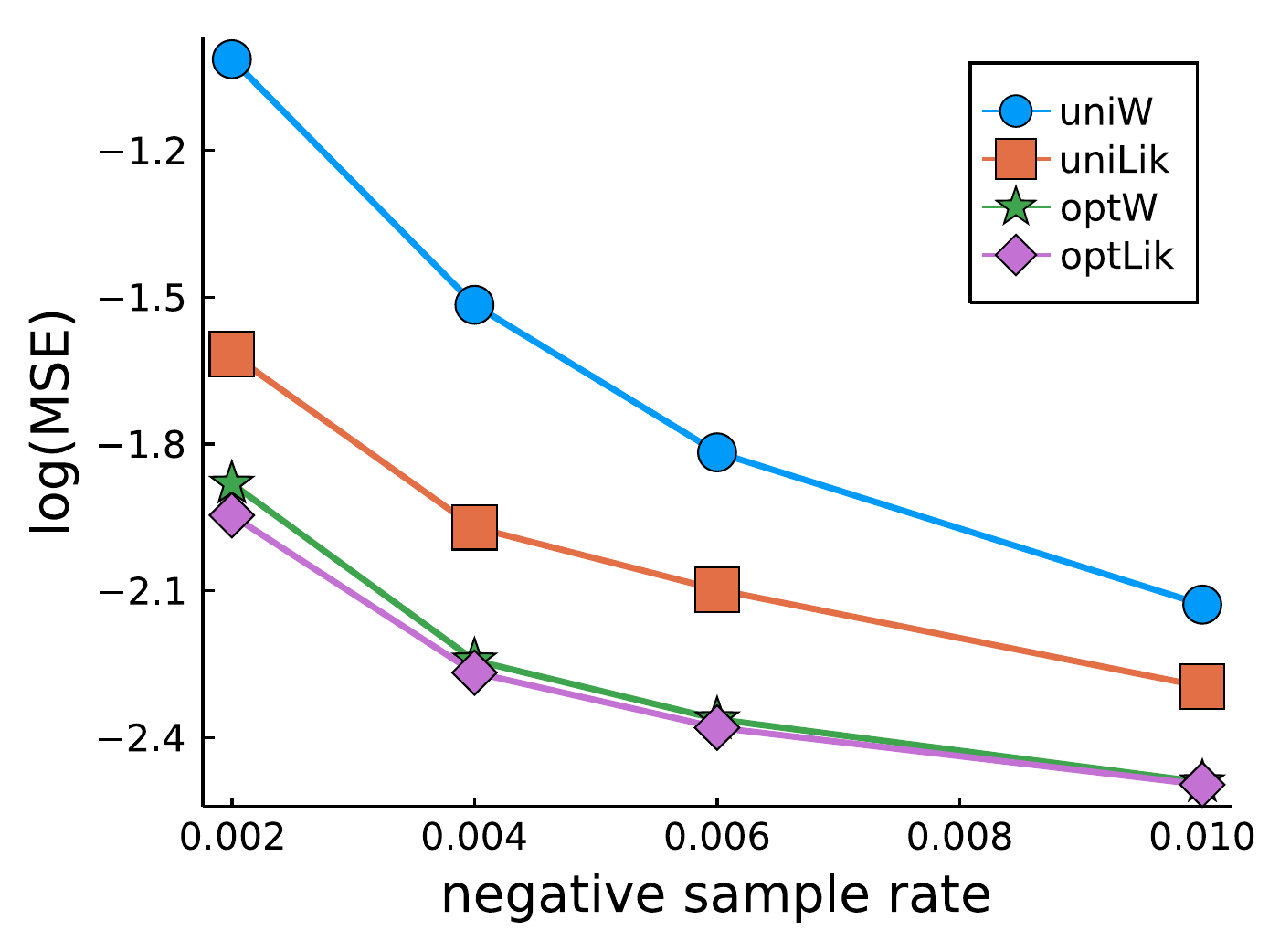}
    (b) {$\x$'s are lognormal}
  \end{minipage} %
  \begin{minipage}{0.23\textwidth}\centering
    \includegraphics[width=\textwidth]{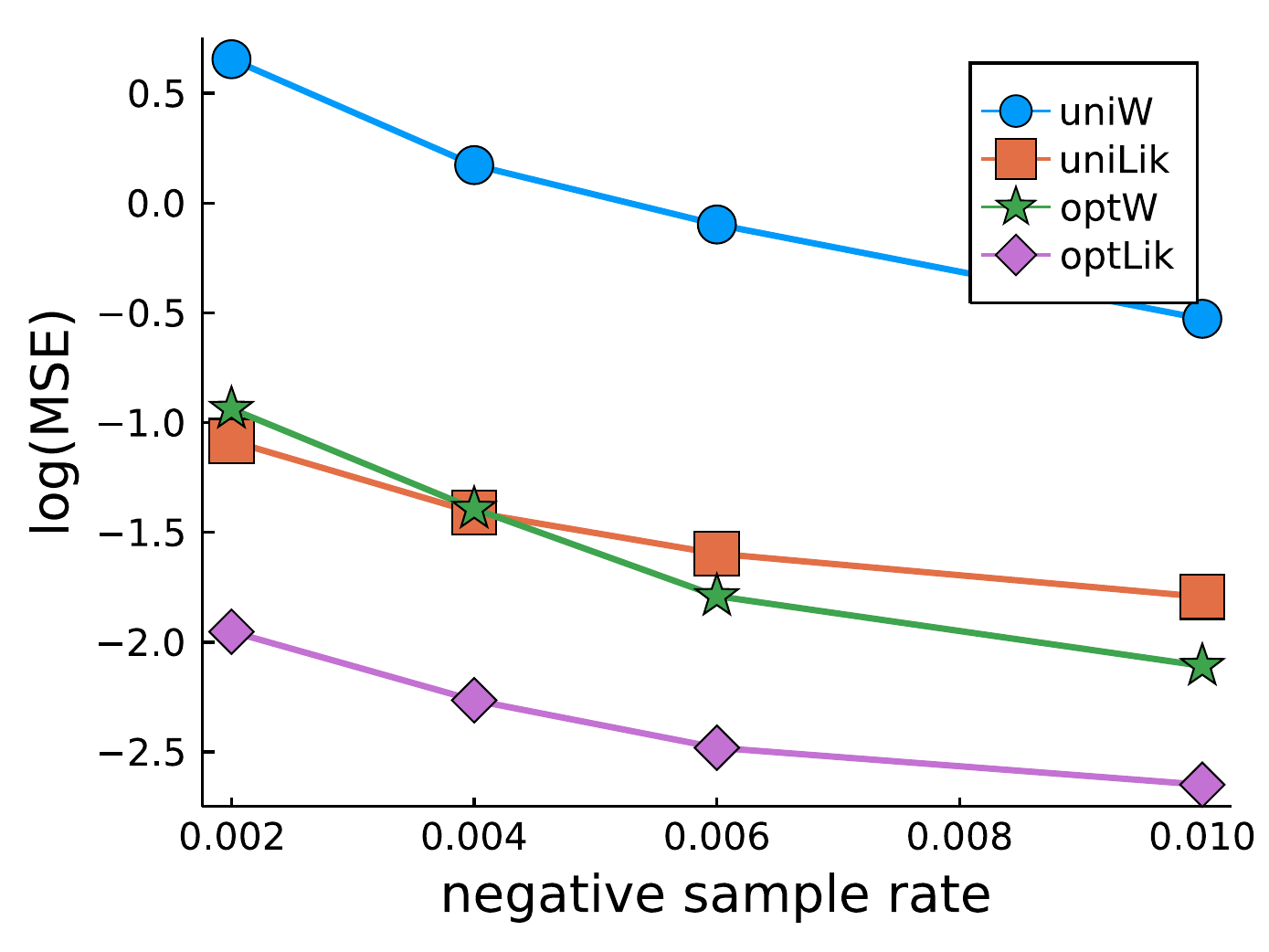}
    (c) {$\x$'s are $t_3$}
  \end{minipage}
  \begin{minipage}{0.23\textwidth}\centering
    \includegraphics[width=\textwidth]{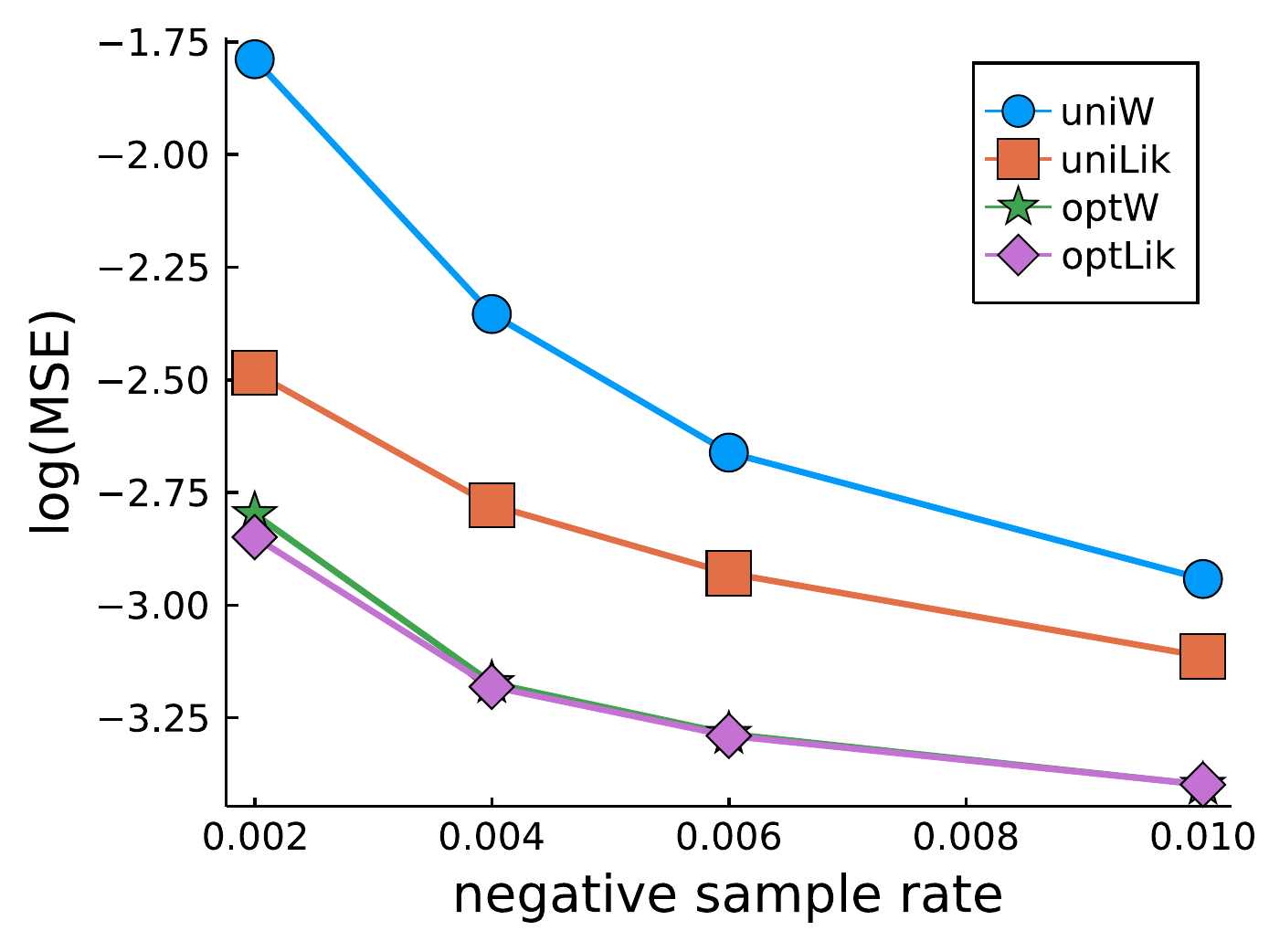}
    (d) {$\x$'s are exponential}
  \end{minipage}
  \caption{Degree of non-linearity $\xi=10^{-1},\xi_w=10^{-3}$. Ground-truth intercepts: (a) $\alpha=-7.7$, (b) $\alpha=-0.6$, (c) $\alpha=-6.9$, (d) $\alpha=-1.9$. Log (MSEs) of subsample estimators for different sample sizes (the smaller the better).}
  \label{fig:supp_model_mis_c_0_1}
\end{figure}

\begin{figure}[H]\centering
  \begin{minipage}{0.23\textwidth}\centering
    \includegraphics[width=\textwidth]{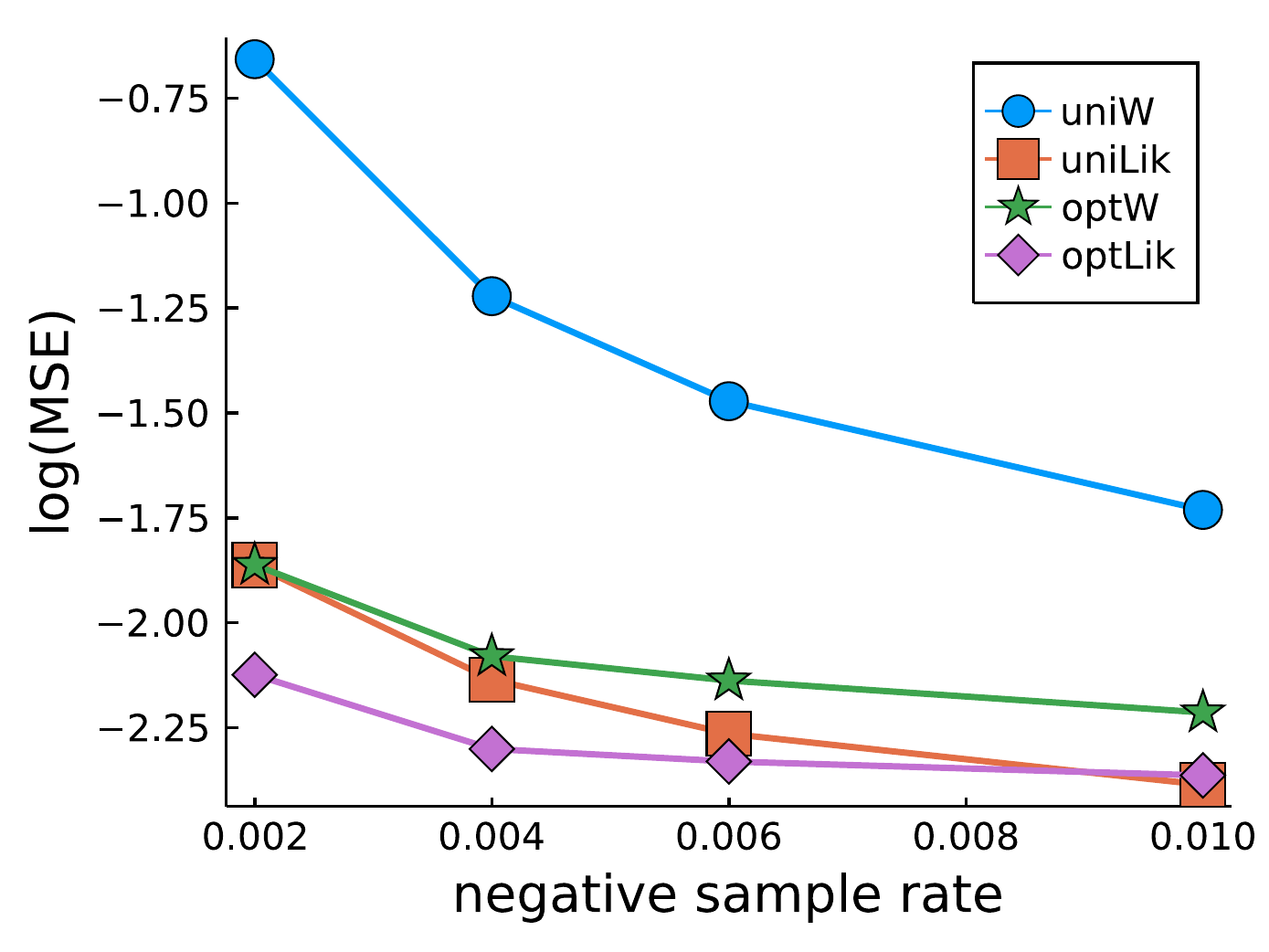}
    (a). {$\x$'s are normal}
  \end{minipage} %
  \begin{minipage}{0.23\textwidth}\centering
    \includegraphics[width=\textwidth]{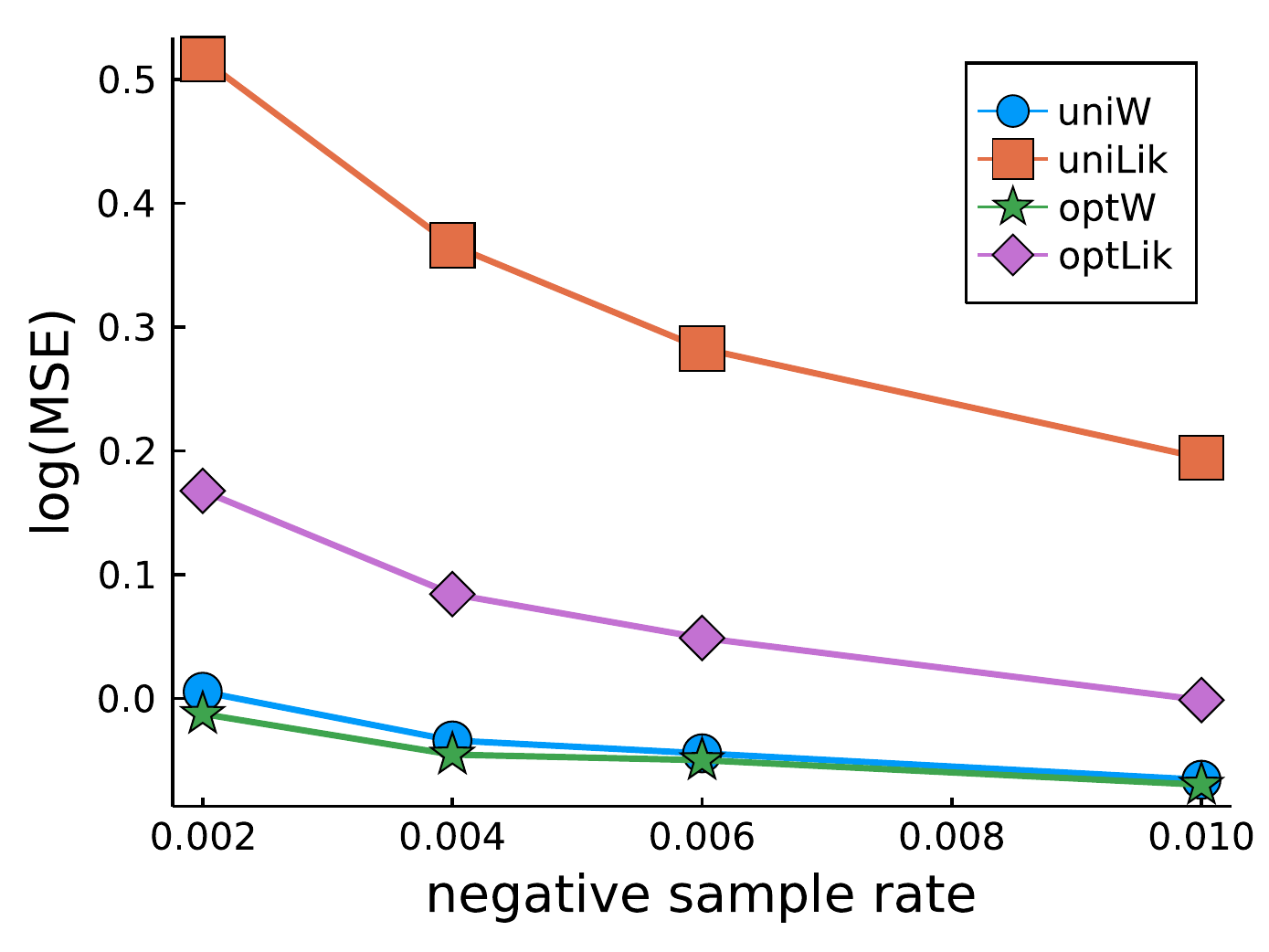}
    (b) {$\x$'s are lognormal}
  \end{minipage} %
  \begin{minipage}{0.23\textwidth}\centering
    \includegraphics[width=\textwidth]{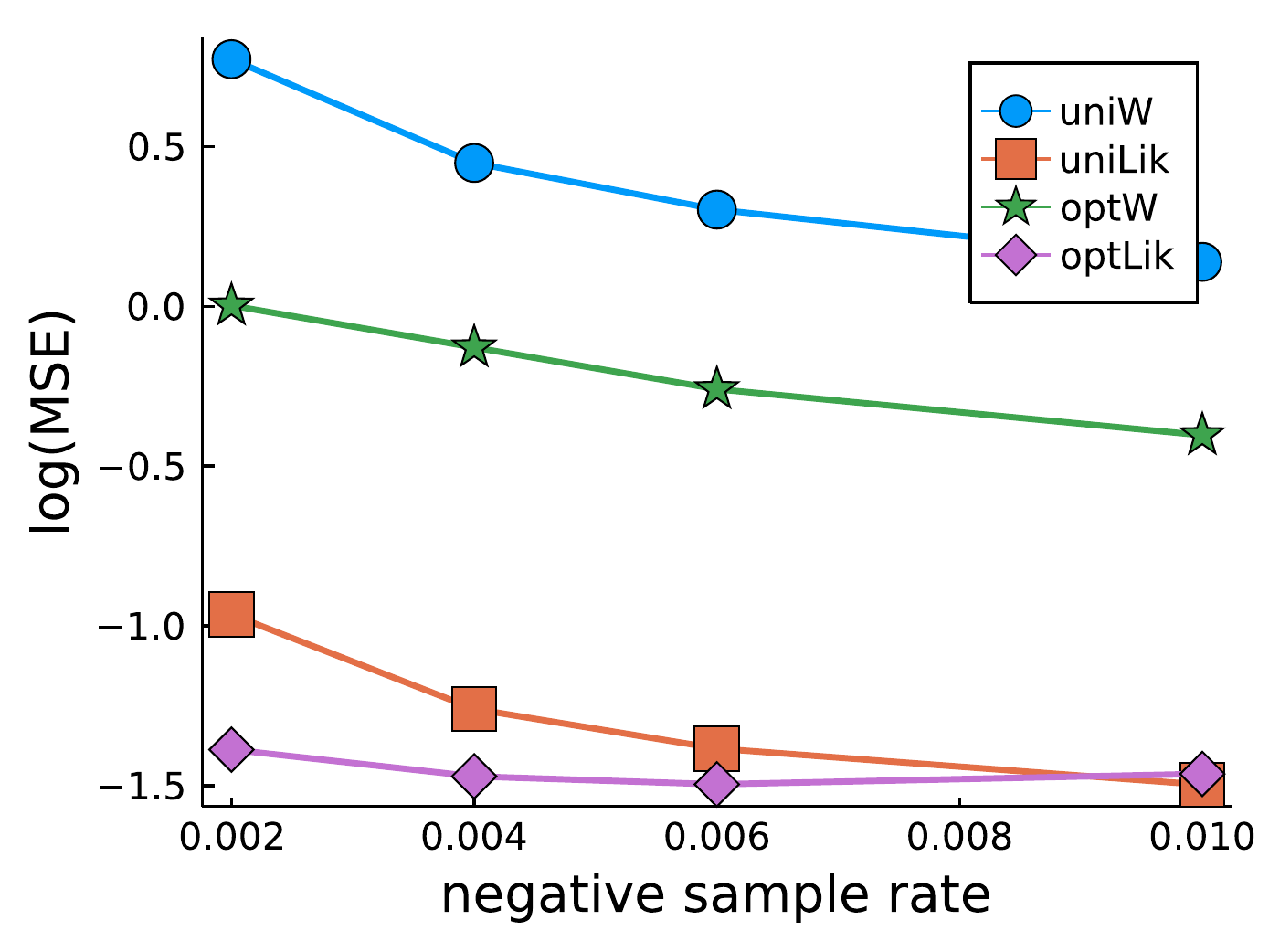}
    (c) {$\x$'s are $t_3$}
  \end{minipage}
  \begin{minipage}{0.23\textwidth}\centering
    \includegraphics[width=\textwidth]{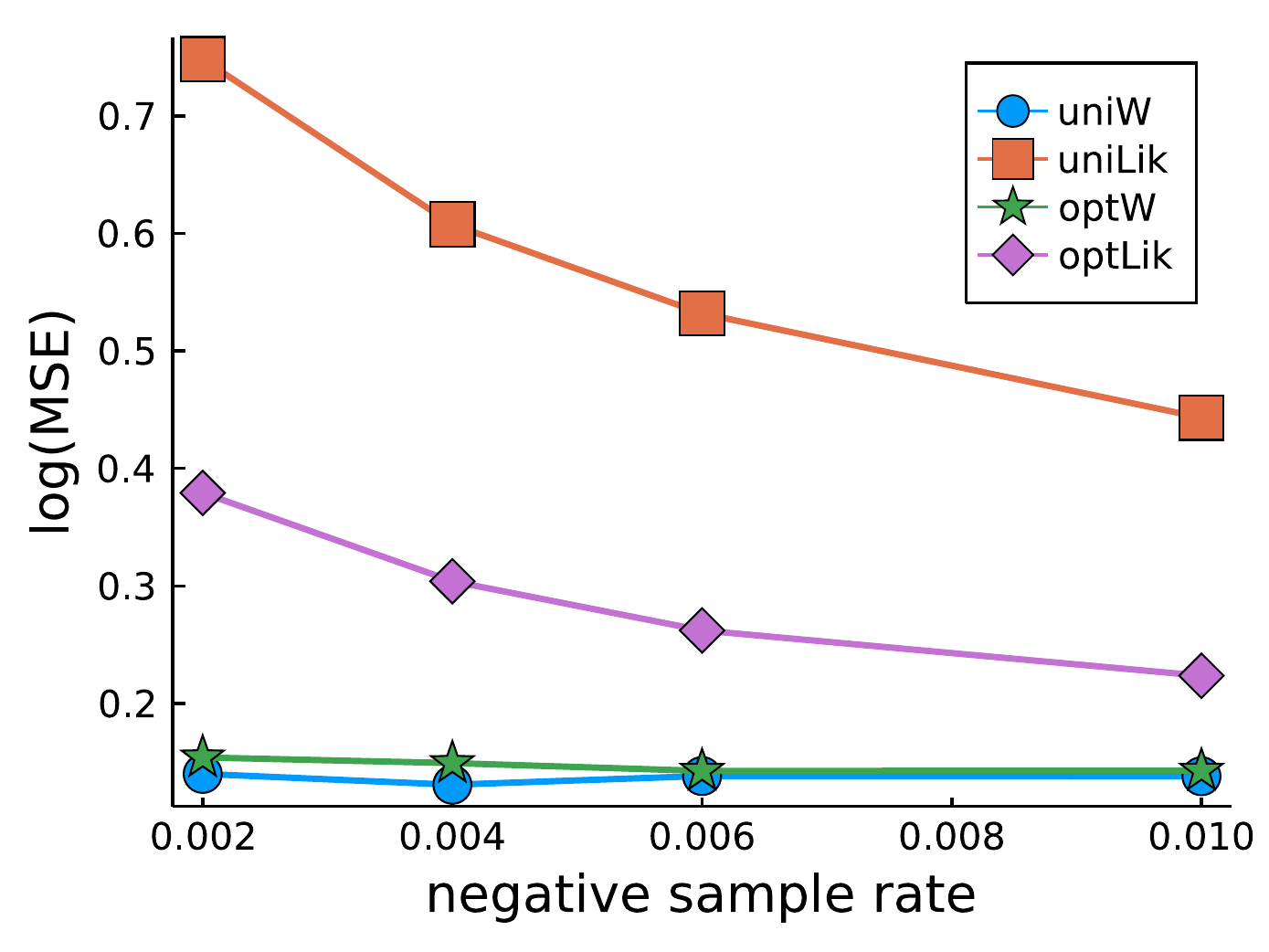}
    (d) {$\x$'s are exponential}
  \end{minipage}
  \caption{Degree of non-linearity $\xi=5\cdot 10^{-1},\xi_w=10^{-3}$. Ground-truth intercepts: (a) $\alpha=-7.4$, (b) $\alpha=-0.95$, (c) $\alpha=-6.7$, (d) $\alpha=-2.15$. Log (MSEs) of subsample estimators for different sample sizes (the smaller the better).}
  \label{fig:supp_model_mis_c_0_5}
\end{figure}

We fix $\xi_w=10^{-3}$ and tune $\xi$ to study the effect of non-linear activation. Note that when $\xi\rightarrow 0$, $\frac{1}{\xi}\tanh(\xi\x)\approx\x$ and when $\xi\rightarrow \infty$, $\frac{1}{\xi}\tanh(\xi\x)\approx 0$. As we shall see, the value $\xi$ significantly affect the non-linearity of the model. In Figure~\ref{fig:supp_model_mis_c_0_1}, we choose a relatively small $\xi=0.1$ and observe an almost indistinguishable loss in $\log(\text{MSE})$. In Figure~\ref{fig:supp_model_mis_c_0_5} we study the case when $\xi=0.5$ and find that it significantly affect the model estimation when $\x$ has long tails (Figure~\ref{fig:supp_model_mis_c_0_5}(b), (d)). When $\xi=1$, all method crash since the true (non-linear) model deviate from the tractable domain (linear). In order to avoid this happen, we suggest to re-scale and centering the data before feeding them to the model. Note that when models are not crashed, optLik still achieves best performance among all methods. 

\begin{figure}[H]\centering
  \begin{minipage}{0.23\textwidth}\centering
    \includegraphics[width=\textwidth]{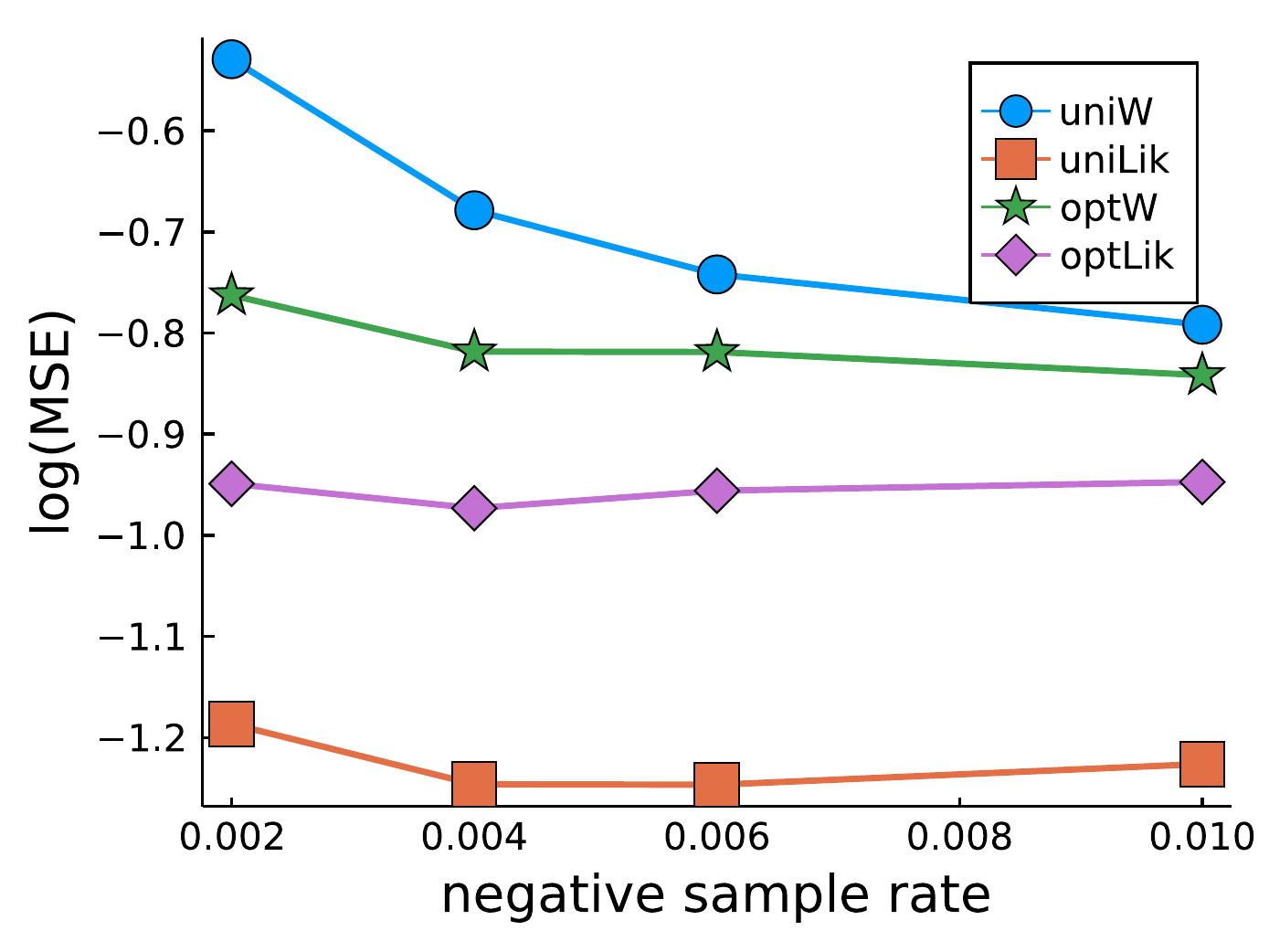}
    (a). {$\x$'s are normal}
  \end{minipage} %
  \begin{minipage}{0.23\textwidth}\centering
    \includegraphics[width=\textwidth]{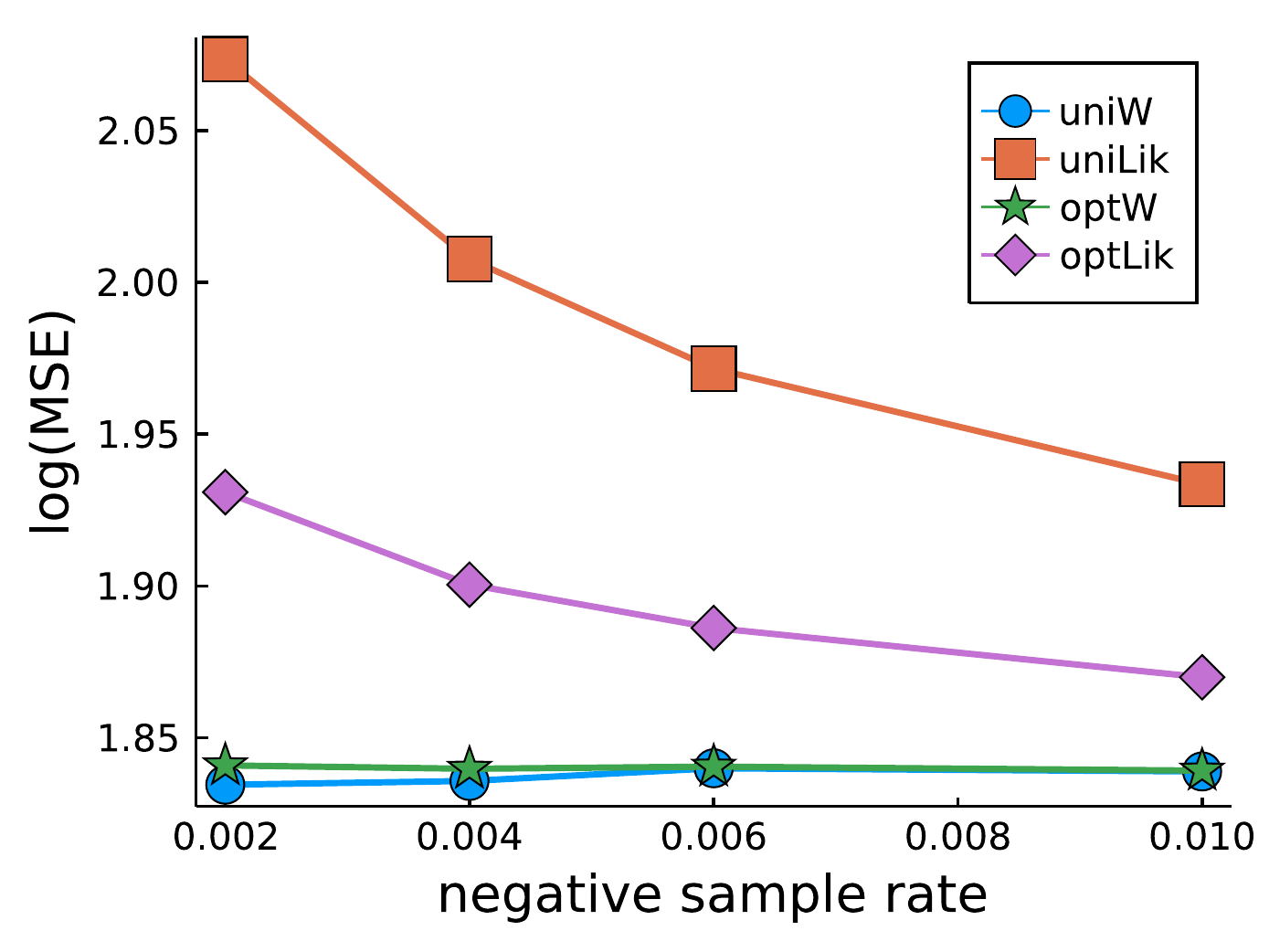}
    (b) {$\x$'s are lognormal}
  \end{minipage} %
  \begin{minipage}{0.23\textwidth}\centering
    \includegraphics[width=\textwidth]{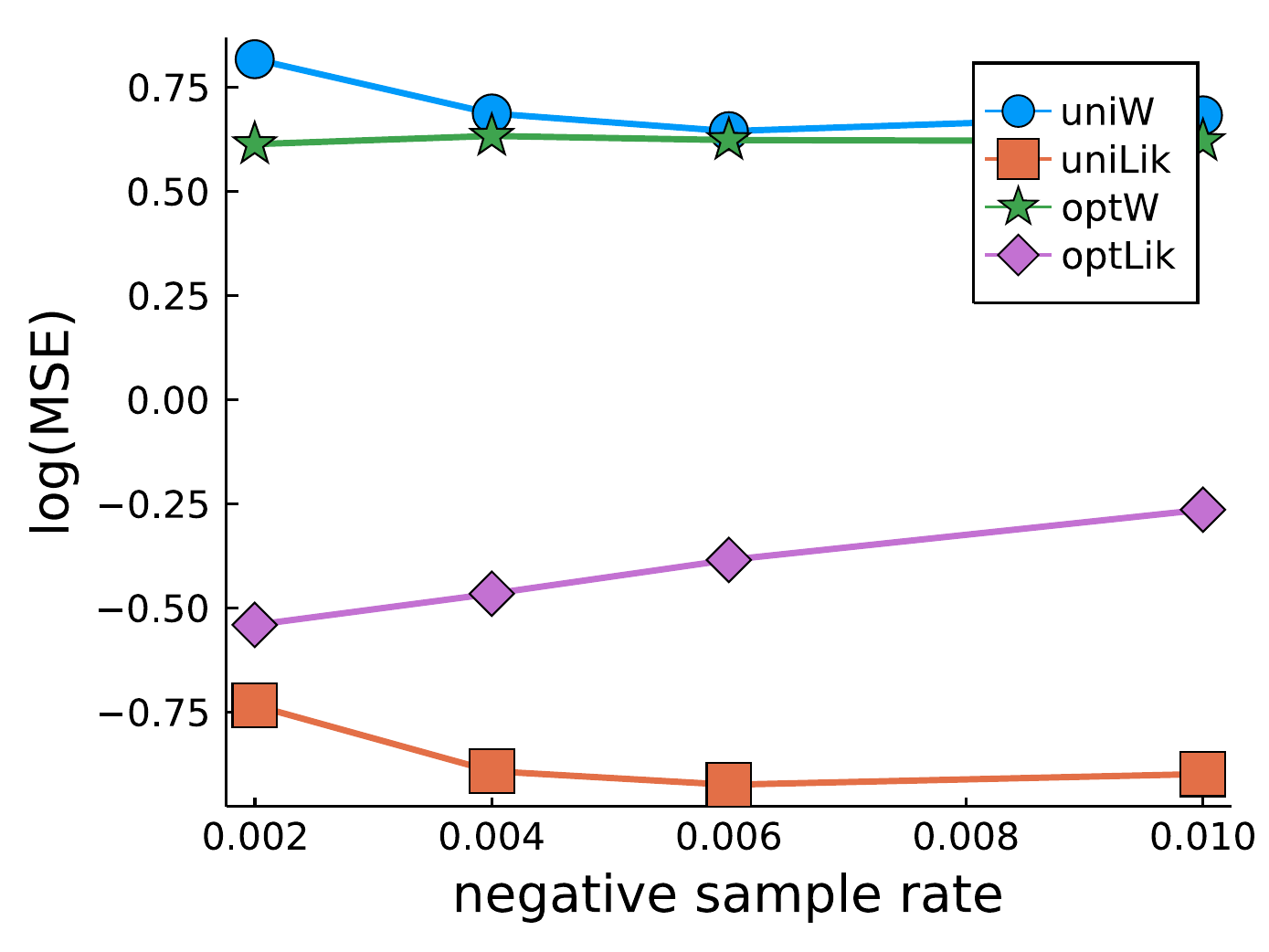}
    (c) {$\x$'s are $t_3$}
  \end{minipage}
  \begin{minipage}{0.23\textwidth}\centering
    \includegraphics[width=\textwidth]{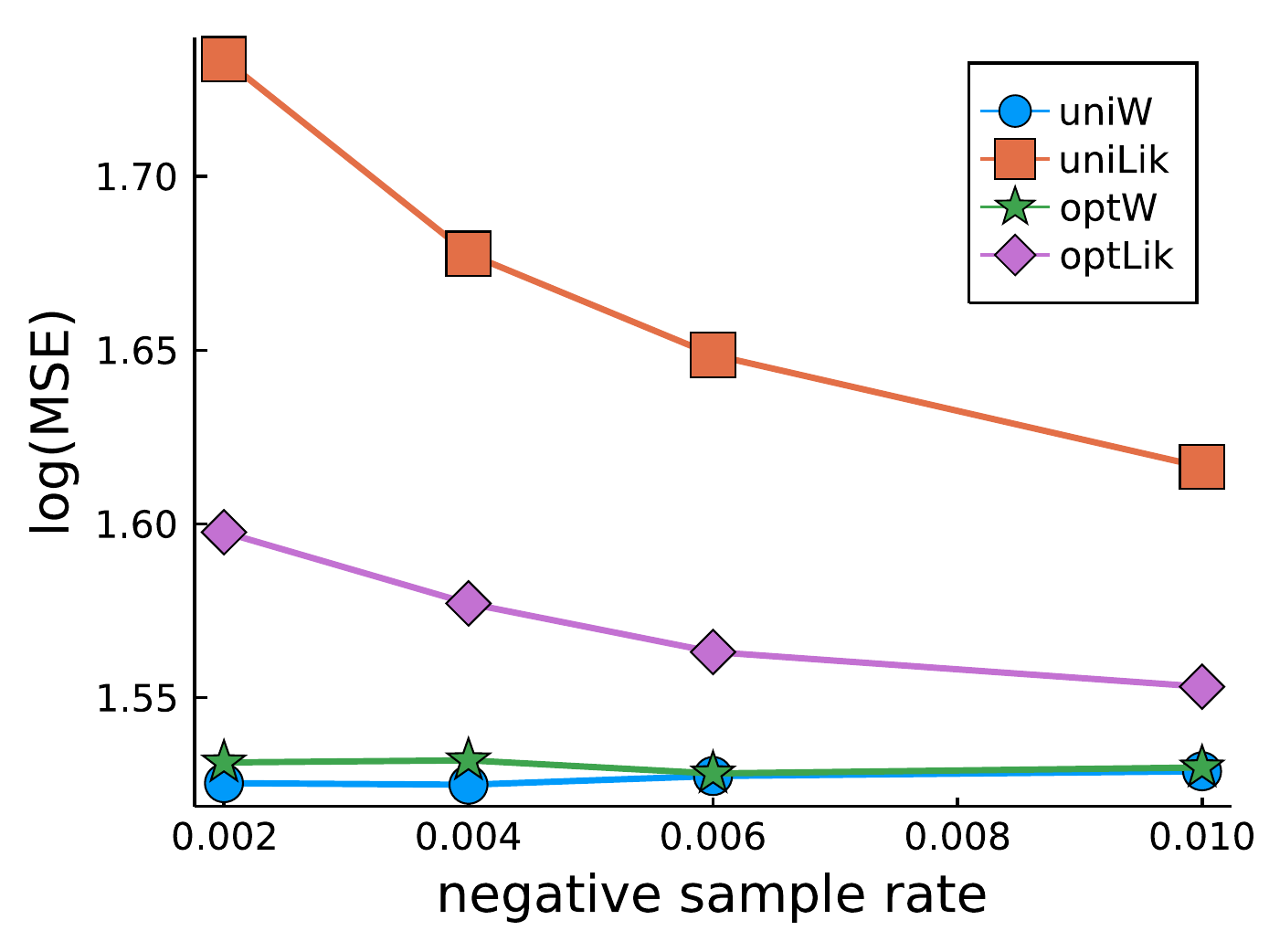}
    (d) {$\x$'s are exponential}
  \end{minipage}
  \caption{Degree of non-linearity $\xi=1.0,\xi_w=10^{-3}$. Ground-truth intercepts: (a) $\alpha=-7.05$, (b) $\alpha=-1.75$, (c) $\alpha=-6.4$, (d) $\alpha=-2.85$. Log (MSEs) of subsample estimators for different sample sizes (the smaller the better).}
  \label{fig:supp_model_mis_c_1}
\end{figure}

\subsection{Tune $\|\mathbf{W}-I\|_{2\rightarrow 2}$ by trying different $\xi_w$.}
\begin{figure}[H]\centering
  \begin{minipage}{0.23\textwidth}\centering
    \includegraphics[width=\textwidth]{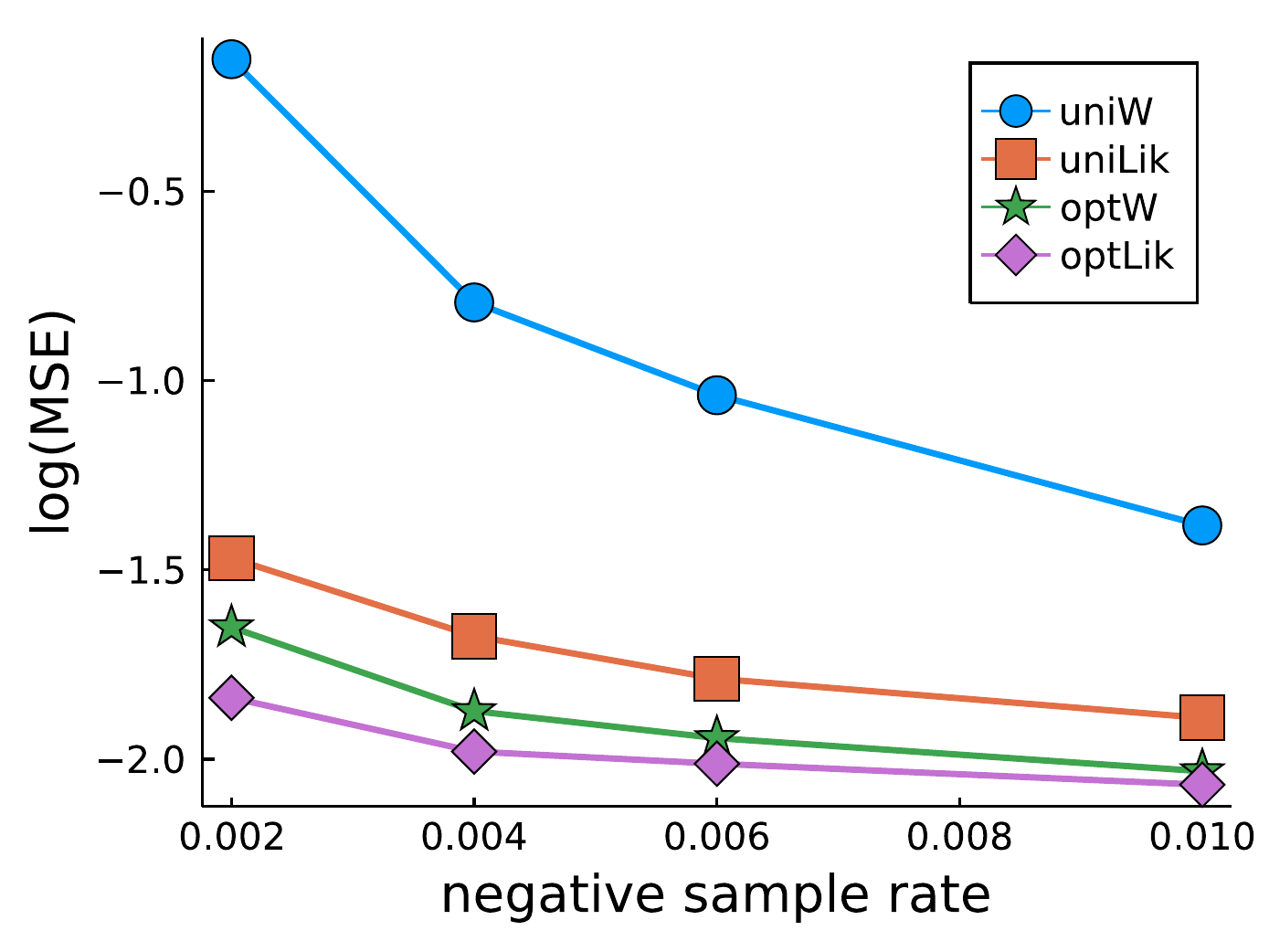}
    (a). {$\x$'s are normal}
  \end{minipage} %
  \begin{minipage}{0.23\textwidth}\centering
    \includegraphics[width=\textwidth]{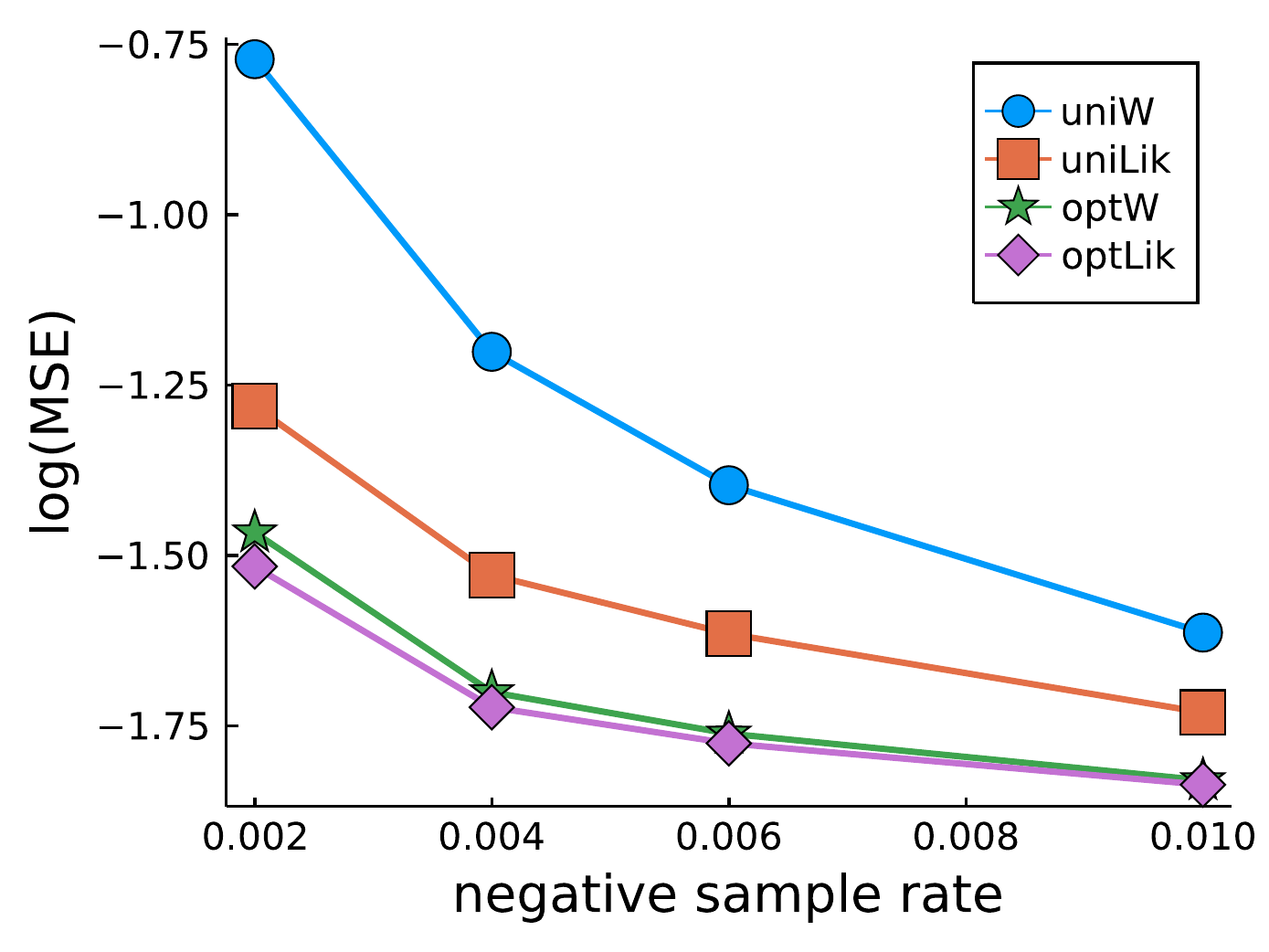}
    (b) {$\x$'s are lognormal}
  \end{minipage} %
  \begin{minipage}{0.23\textwidth}\centering
    \includegraphics[width=\textwidth]{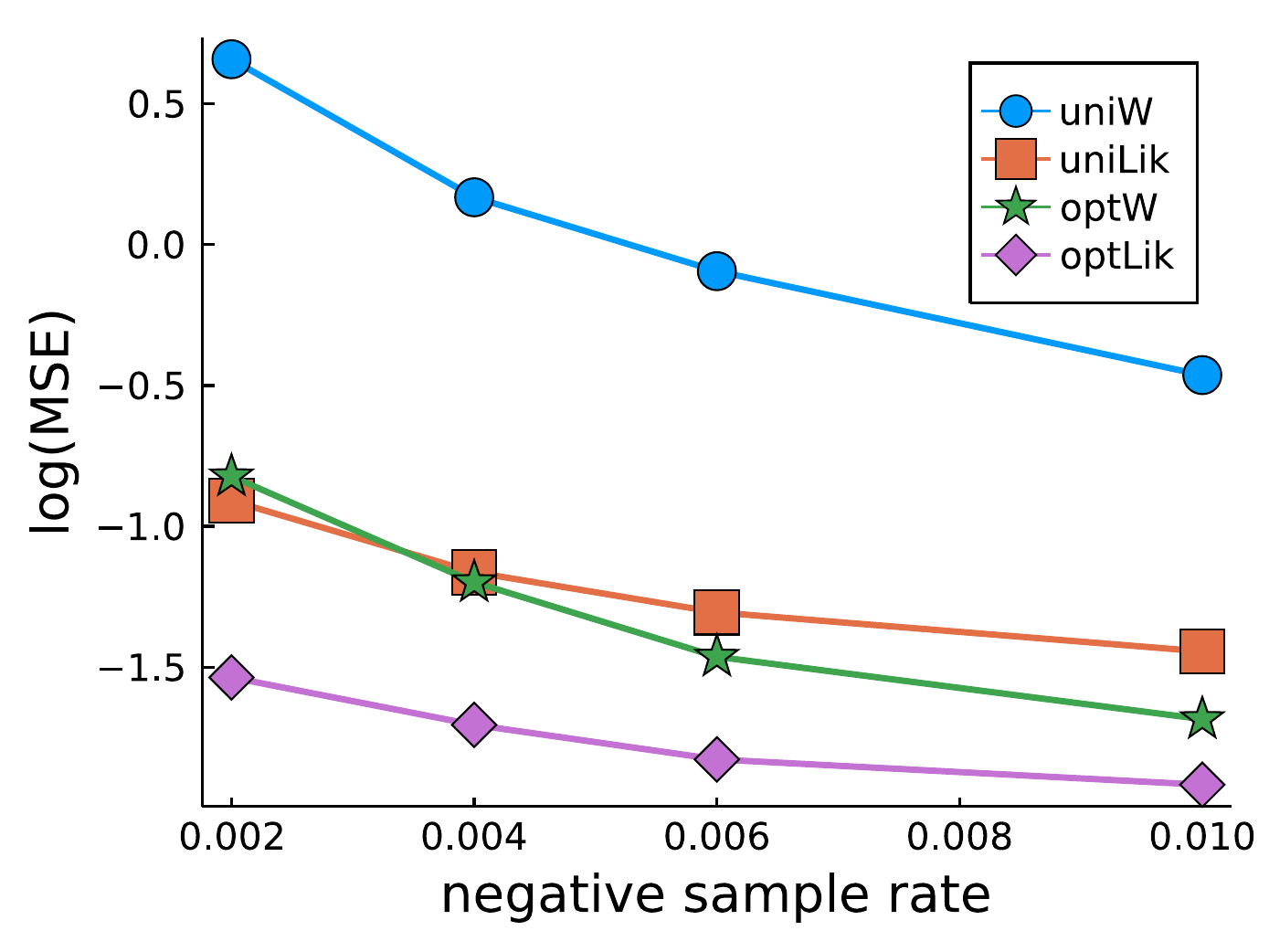}
    (c) {$\x$'s are $t_3$}
  \end{minipage}
  \begin{minipage}{0.23\textwidth}\centering
    \includegraphics[width=\textwidth]{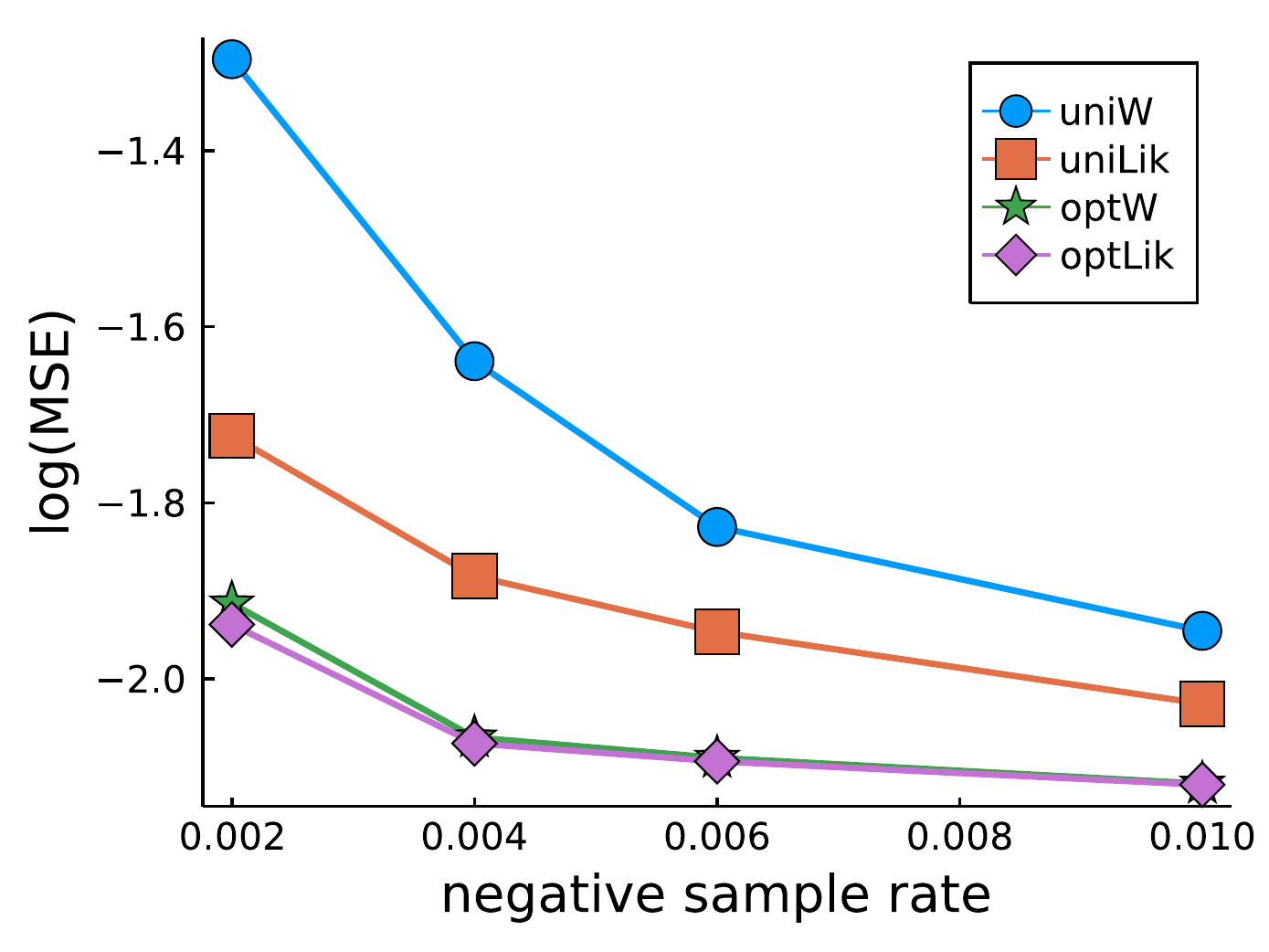}
    (d) {$\x$'s are exponential}
  \end{minipage}
  \caption{Degree of non-linearity $\xi=10^{-3},\xi_w=2\cdot 10^{-2}$. Ground-truth intercepts: (a) $\alpha=-7.7$, (b) $\alpha=-0.5$, (c) $\alpha=-6.9$, (d) $\alpha=-1.9$. Log (MSEs) of subsample estimators for different sample sizes (the smaller the better).}
  \label{fig:supp_model_mis_c_w_0_02}
\end{figure}

\begin{figure}[H]\centering
  \begin{minipage}{0.23\textwidth}\centering
    \includegraphics[width=\textwidth]{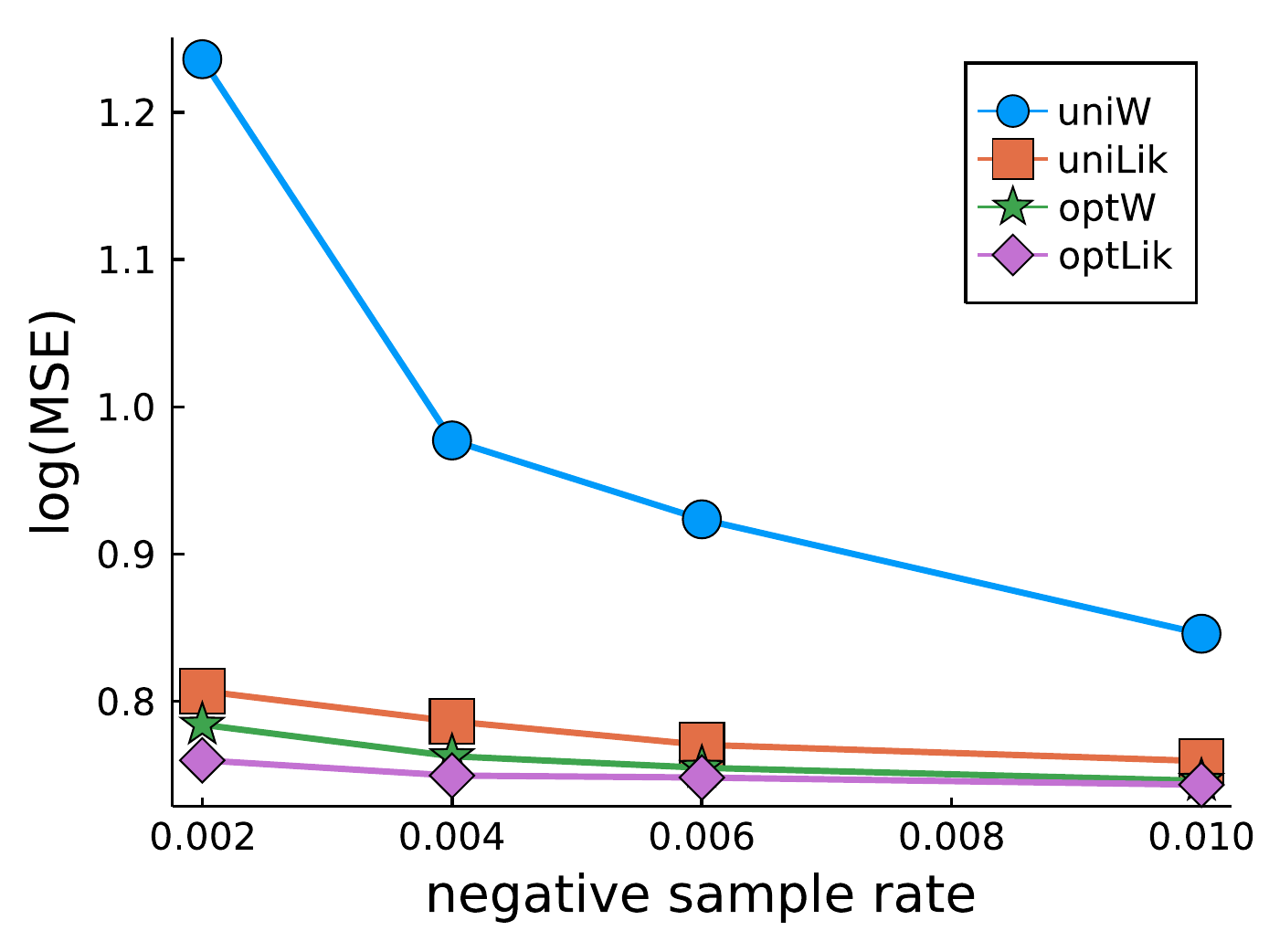}
    (a). {$\x$'s are normal}
  \end{minipage} %
  \begin{minipage}{0.23\textwidth}\centering
    \includegraphics[width=\textwidth]{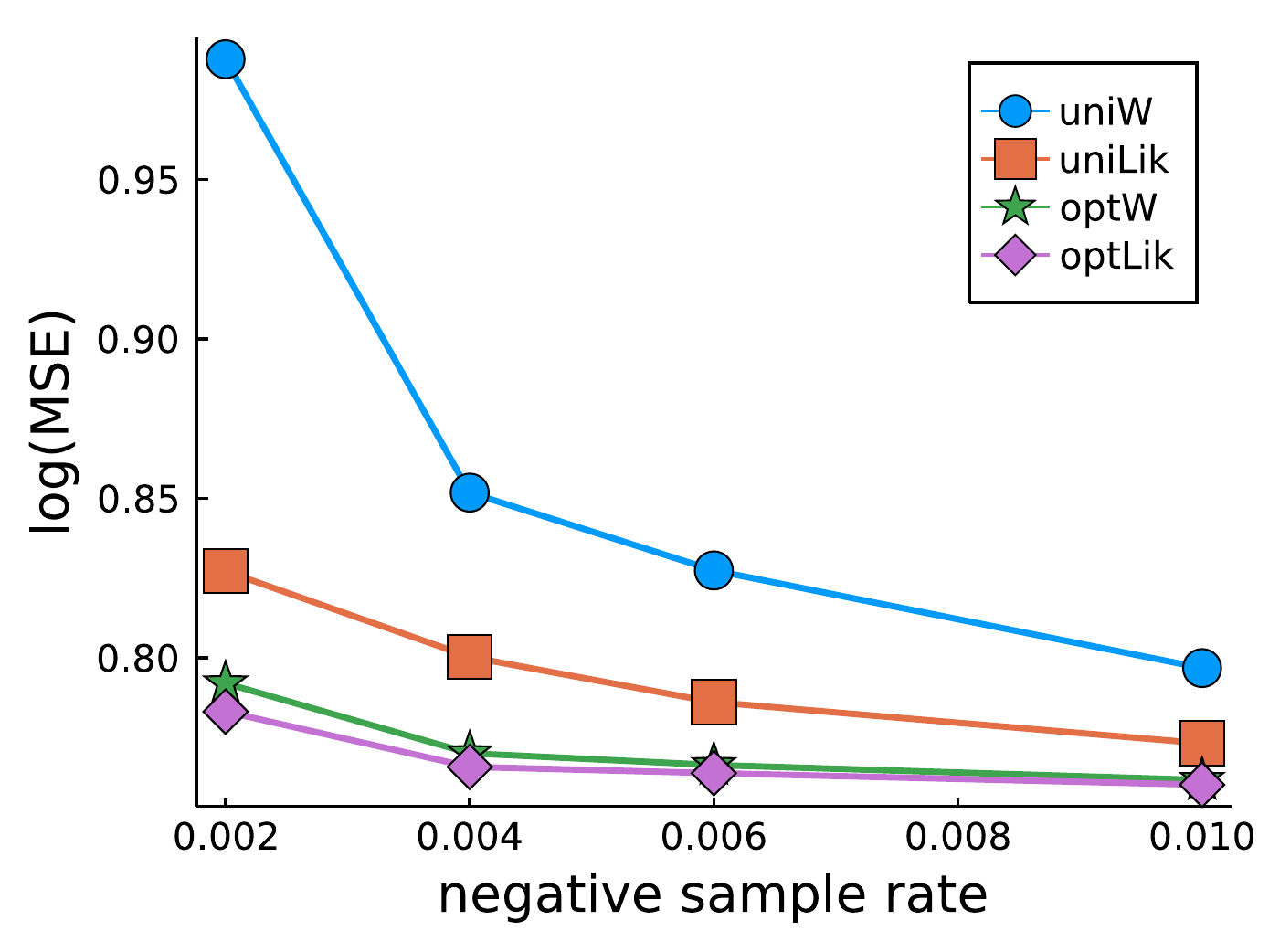}
    (b) {$\x$'s are lognormal}
  \end{minipage} %
  \begin{minipage}{0.23\textwidth}\centering
    \includegraphics[width=\textwidth]{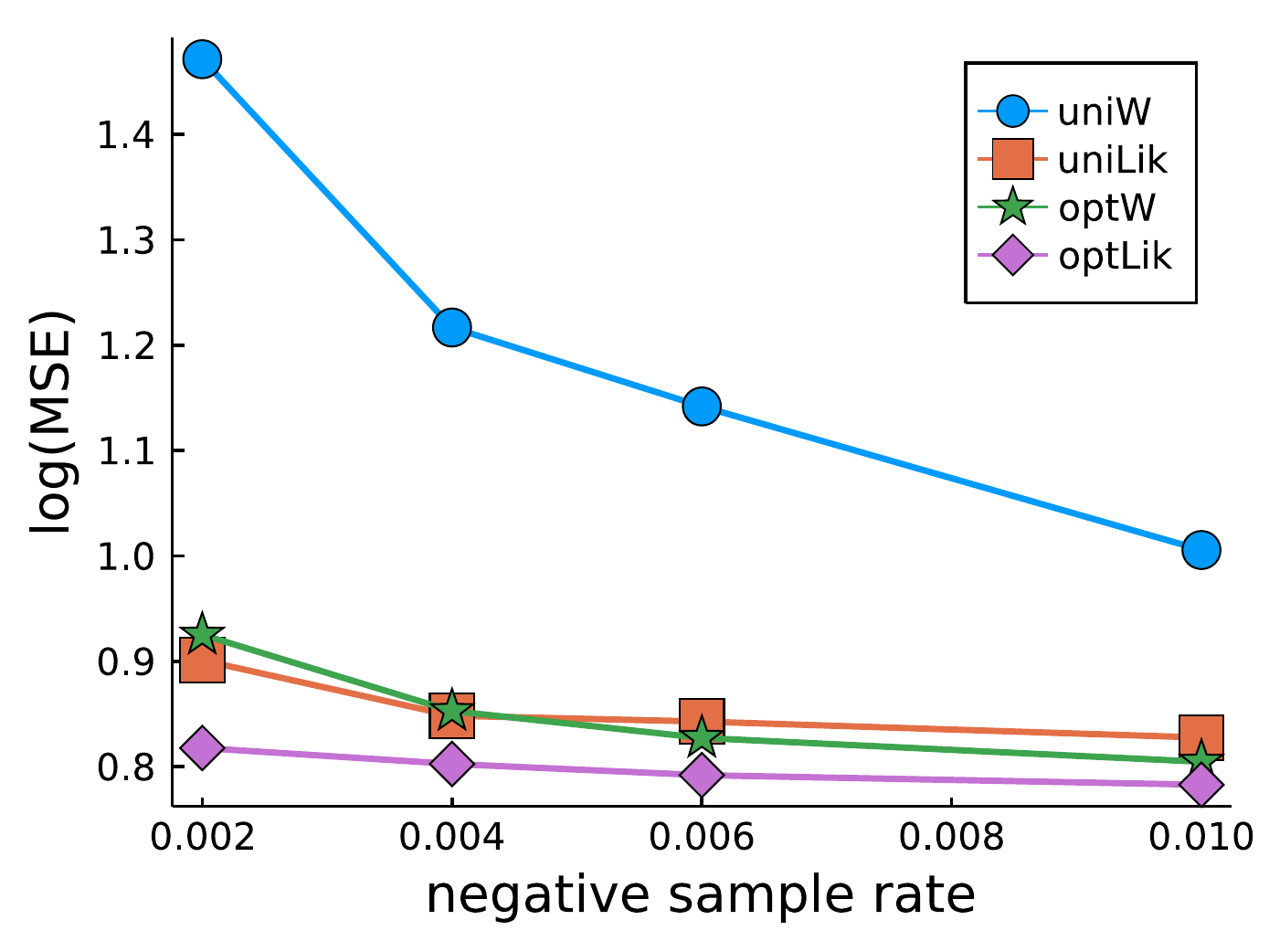}
    (c) {$\x$'s are $t_3$}
  \end{minipage}
  \begin{minipage}{0.23\textwidth}\centering
    \includegraphics[width=\textwidth]{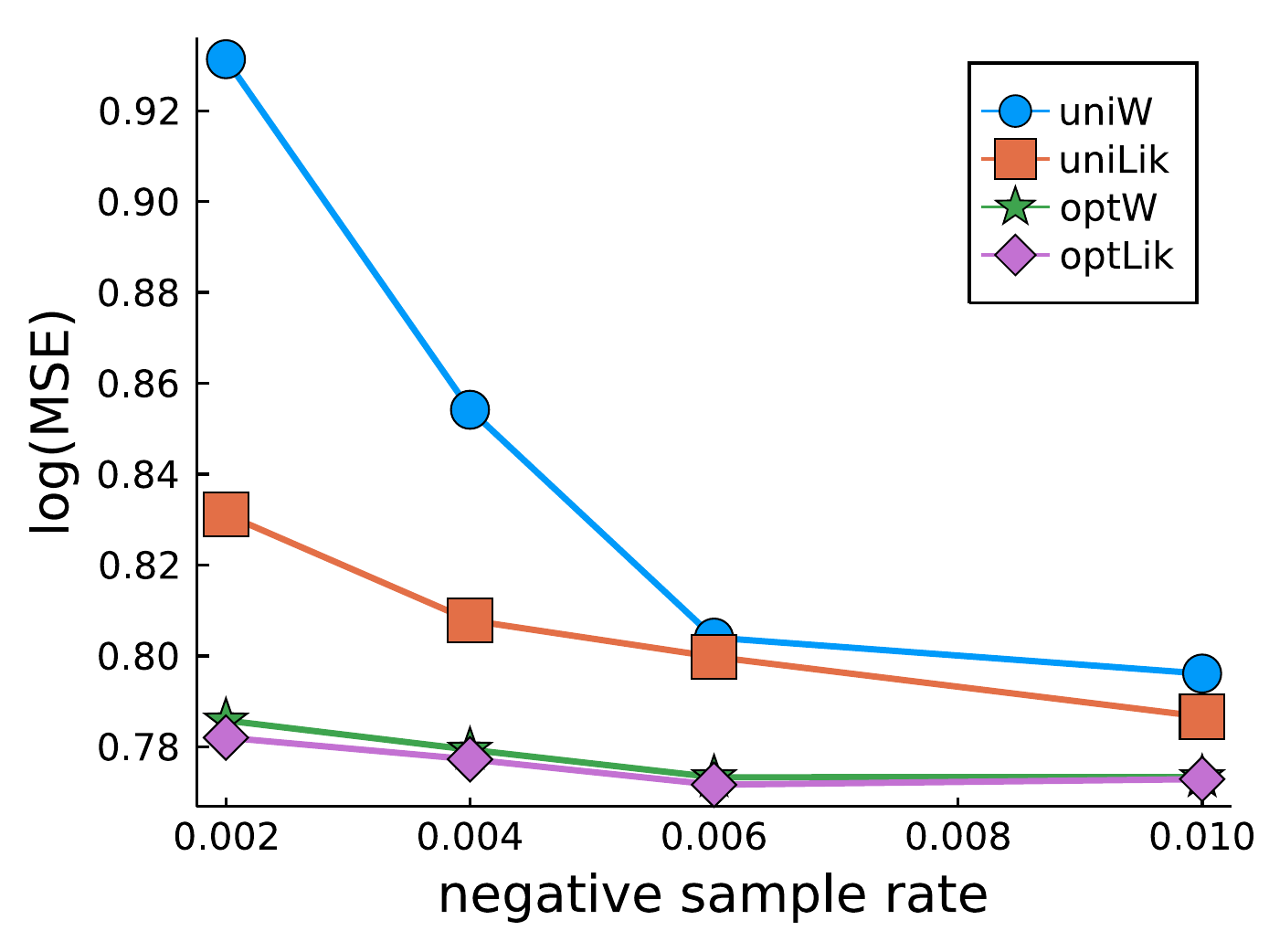}
    (d) {$\x$'s are exponential}
  \end{minipage}
  \caption{Degree of non-linearity $\xi=10^{-3},\xi_w=10^{-1}$. Ground-truth intercepts: (a) $\alpha=-8.25$, (b) $\alpha=-0.25$, (c) $\alpha=-6.6$, (d) $\alpha=-2.4$. Log (MSEs) of subsample estimators for different sample sizes (the smaller the better).}
  \label{fig:supp_model_mis_c_w_0_1}
\end{figure}

We fix $\xi=10^{-3}$ and focus on studying the effect of generative process of $\mathbf{W}$ to the estimation error. In Figure~\ref{fig:supp_model_mis_c_w_0_02}, we set $\xi_w=2\cdot 10^{-2}$ to be a relatively small number such that  $\|\mathbf{W}-I\|_{2\rightarrow 2}\approx 0.20 < 1$. In this case, each sampling method suffers from model mis-specification with higher $\log(\text{MSE})$. Still optLik is the best among all generative processes of $\x$. In Figure~\ref{fig:supp_model_mis_c_w_0_1}, we set $\xi_w=10^{-1}$ and $\|\mathbf{W}-I\|_{2\rightarrow 2}\approx 1$. In this case, the noise level is comparable to the ground-truth. All sampling models crash such that the estimated models deviate significantly from the ground-truth.

\bibliography{ref}
\bibliographystyle{natbib}

\end{document}